%% file: main.tex
\documentclass[10pt,twocolumn,letterpaper]{article}

\usepackage[pagenumbers]
{cvpr}      

\usepackage[accsupp]{axessibility} 

\definecolor{cvprblue}{rgb}{0.21,0.49,0.74}
\usepackage[pagebackref,breaklinks,colorlinks,allcolors=cvprblue]{hyperref}
\usepackage{algorithm}
\usepackage{algpseudocode}
\usepackage{multirow}
\usepackage{graphicx}
\usepackage{subcaption}
\usepackage[table]{xcolor}
\usepackage{colortbl}
\usepackage{arydshln}
\usepackage[sectionbib]{chapterbib}

\newcommand{\myName}{BALM\xspace}
\vspace{-9pt}
\def\paperID{18320} 
\def\confName{CVPR}
\def\confYear{2026}

\title{BALM: A Model-Agnostic Framework for Balanced Multimodal Learning \\under Imbalanced Missing Rates}

\author{
Phuong-Anh Nguyen,
Tien Anh Pham,
Duc-Trong Le,
Cam-Van Thi Nguyen\thanks{Corresponding author}\\
VNU University of Engineering and Technology\\
Hanoi, Vietnam\\
{\tt\small \{22028332, 23021696, trongld, vanntc\}@vnu.edu.vn}
}

\begin{document}
\maketitle
\input{sec/0_abstract}  
\input{sec/1_intro}
\input{sec/2_related}

\input{sec/3_methodology_v0.3}

\input{sec/4_experiment_setting}

\input{sec/5_results}

    \bibliographystyle{ieeenat_fullname}
    \bibliography{main}

\input{sec/X_suppl}

\end{document}

%% file: sec/0_abstract.tex
\begin{abstract}
Learning from multiple modalities often suffers from imbalance, where information-rich modalities dominate optimization while weaker or partially missing modalities contribute less. This imbalance becomes severe in realistic settings with imbalanced missing rates (IMR), where each modality is absent with different probabilities, distorting representation learning and gradient dynamics. We revisit this issue from a training-process perspective and propose \textbf{BALM}, a model-agnostic plug-in framework to achieve balanced multimodal learning under IMR. 
The framework comprises two complementary modules: the Feature Calibration Module (\texttt{FCM}), which recalibrates unimodal features using global context to establish a shared representation basis across heterogeneous missing patterns; the Gradient Rebalancing Module (\texttt{GRM}), which balances learning dynamics across modalities by modulating gradient magnitudes and directions from both distributional and spatial perspectives. BALM can be seamlessly integrated into diverse backbones, including multimodal emotion recognition (MER) models, without altering their architectures. Experimental results across multiple MER benchmarks confirm that BALM consistently enhances robustness and improves performance under diverse missing and imbalance settings.
Code available at: \url{https://github.com/np4s/BALM_CVPR2026.git}
\end{abstract}

%% file: sec/1_intro.tex
\section{Introduction}

Multimodal learning has achieved substantial progress by jointly exploiting complementary information from audio, visual, and textual modalities~\cite{baltruvsaitis2018multimodal,poria2017review, liang2024foundations, xu2023multimodal}. 
However, real-world multimodal systems rarely operate under perfect conditions, as sensor failures, recording noise, or acquisition costs often cause partial or complete modality loss~\cite{wu2024deep,zhang2024multimodal, pham2019MCTN}. 
This phenomenon leads to \textit{incomplete multimodal learning}, where missing data affect both representation quality and cross-modal interaction~\cite{zhang2024multimodal}. 
Effectively handling incomplete or missing modalities remains a central challenge for building reliable multimodal systems~\cite{lincyin}. 
Beyond incompleteness, multimodal learning inherently faces \textit{modality imbalance}~\cite{wu2022characterizing,xu2025balancebenchmark, wang2020makes}, where modalities contribute unequally due to differences in signal reliability, feature granularity, and data scale. When modalities are missing, this imbalance worsens as the dominant modality overfits while the less frequent or corrupted ones receive limited supervision, leading to a biased equilibrium that weakens robustness and generalization.

\begin{figure}[t]
    \centering
    \includegraphics[width=0.95\linewidth]{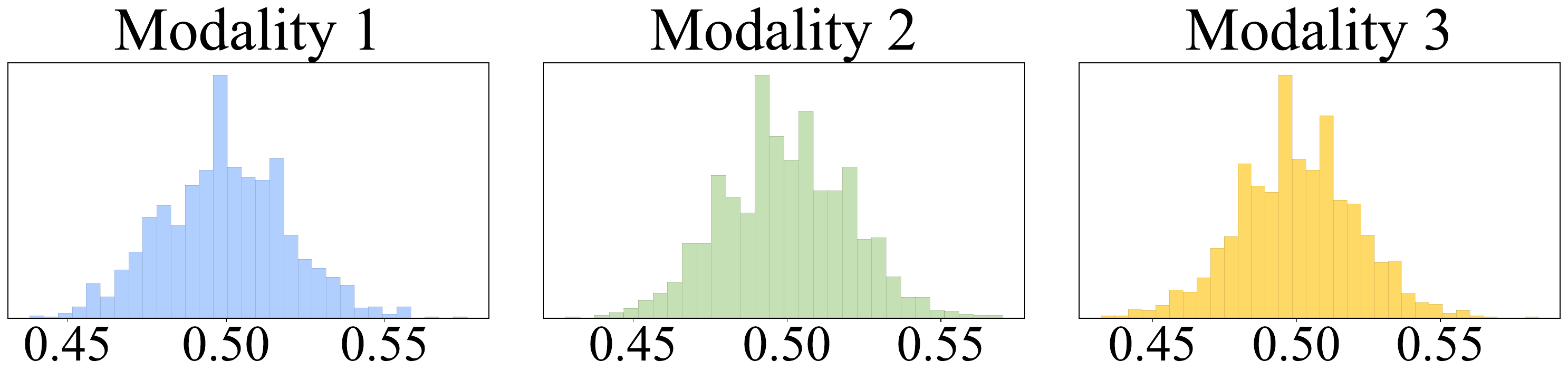}
    \caption{Modality-specific missing rates' distribution of a tri-modal dataset under Share Missing Rate (SMR=0.5).}
    \label{fig:share-mr}
\end{figure}
\begin{figure}[t]
    \centering
    \includegraphics[width=0.95\linewidth]{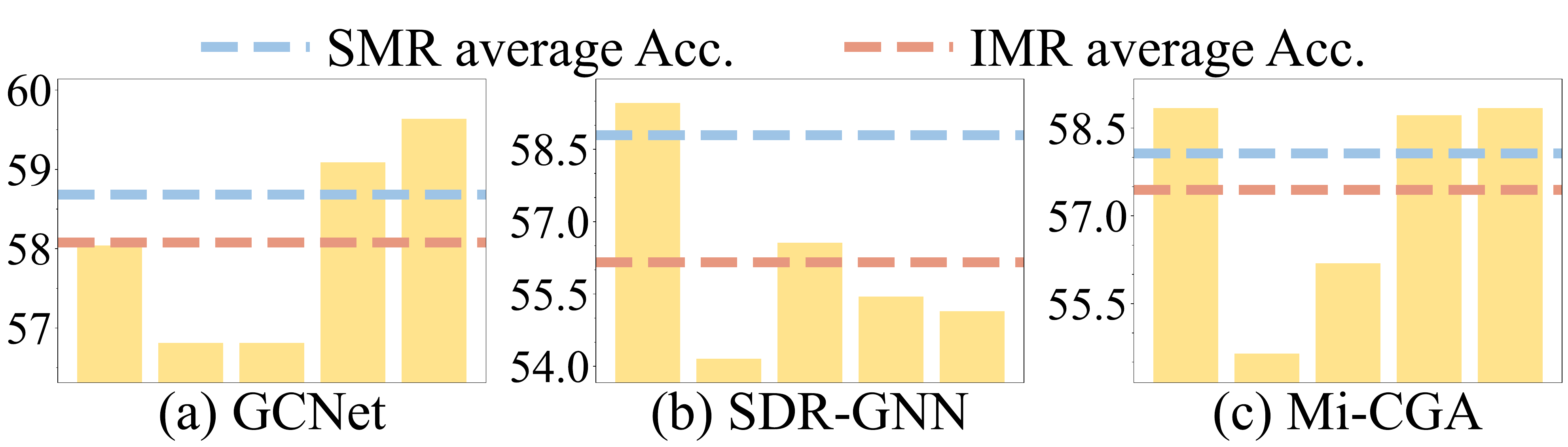}
    \caption{Model accuracy under SMR=0.5 and five equivalent IMR settings with $(r_A, r_L, r_V)\in\{0.3,0.5,0.7\}$ on IEMOCAP. 
    }
    \label{fig:smr-vs-imr}
\end{figure}

Recent advances in missing-modality learning have explored two main paradigms to enhance robustness under incomplete data:  
\textit{alignment-based methods}, which align representations between complete and incomplete inputs using contrastive, correlation, or augmentation techniques~\cite{liu2024contrastive,pham2019MCTN,lian2023gcnet,nguyen2025-micga}, and  
\textit{generation-based methods}, which reconstruct missing modalities through autoencoders, variational inference, graph reasoning, or diffusion models~\cite{ma2021smil,zhao2021MMIN,wang2023incomplete,xu2024leveraging}.  
These approaches generally overlook the imbalance that arises when missingness occurs during training. Notably, some models~\cite{lian2023gcnet,fu2024sdr} compute reconstruction losses using the full data, whereas some approaches~\cite{zhao2021MMIN,xu2024leveraging} do not employ reconstruction but rely on pretrained encoders trained on full modalities, leaving challenges of incomplete training data unaddressed.

Most prior studies further simplify the problem by adopting the \textit{Shared Missing Rate (SMR)} assumption~\cite{lian2023gcnet,fu2024sdr,zhao2021MMIN}, where all modalities are randomly dropped with the same probability.  
As illustrated in Fig.~\ref{fig:share-mr}, when $SMR\!=\!0.5$, each modality exhibits comparable missing proportions, resulting in a statistically balanced yet idealized setting.  
In contrast, realistic multimodal systems often encounter uneven data degradation across modalities due to heterogeneous acquisition or noise.  
We refer to this scenario as the \textit{Imbalanced Missing Rate (IMR)} condition~\cite{shi2024passion,sun2024redcore,zhao2025mce}, which introduces persistent exposure disparity and optimization bias across modalities.
As shown in Fig.~\ref{fig:smr-vs-imr}, traditional missing-modality models exhibit unstable behavior under \textit{IMR} settings, particularly when the missing rates across modalities differ significantly. 
This imbalance causes performance fluctuations, as models struggle to adapt when one modality becomes substantially less reliable than the others.


\textbf{Our approach.}
We study multimodal learning under the \textit{Imbalanced Missing Rate (IMR)} setting, where modality absence occurs from the beginning of training and follows distinct missing rates. 
Imbalanced Missing Rate introduces two coupled challenges: 
(1) \textit{representation imbalance}, where heterogeneous missing patterns distort unimodal feature distributions and hinder consistent cross-modal fusion~\cite{tsai2019multimodal}; and 
(2) \textit{learning imbalance}, where gradients are dominated by frequently observed modalities, leading to biased convergence~\cite{guo2024classifier}. 
To address these issues, we propose \textbf{\myName} (\textit{\textbf{B}alanced \textbf{A}gnostic \textbf{L}earning under Imbalanced \textbf{M}issing Rates}), 
a lightweight and model-agnostic plug-in framework that mitigates both representation- and optimization-level imbalance under IMR. 
Our propose \myName integrates seamlessly with existing multimodal backbones, particularly those used in MER. It comprises two complementary modules: 
a Feature Calibration Module (\texttt{FCM}) that aligns representations across varying missing patterns, 
and a Gradient Rebalancing Module (\texttt{GRM}) that harmonizes optimization dynamics by adaptively modulating gradient magnitudes and directions.
Our main contributions are summarized as follows:
\begin{itemize}[leftmargin=*]
    \item We propose \textbf{\myName}, a plug-in framework that enables robust multimodal learning under the \textit{Imbalanced Missing Rate} condition without redesigning existing architectures.
    \item We design two complementary modules: a \textit{Feature Calibration Module} \texttt{(FCM)} for representation-level alignment and a \textit{Gradient Rebalancing Module} \texttt{(GRM)} for optimization-level balancing.
    \item We conduct extensive experiments on multiple MER benchmarks, showing that \myName{} consistently enhances robustness across standard backbones, imbalance-oriented models, and missing-modality methods.

\end{itemize}

%% file: sec/2_related.tex
\section{Related Work}

\subsection{Incomplete Multimodal Learning}
Recent studies on incomplete multimodal learning can be broadly categorized into two paradigms~\cite{zhang2024multimodal}.  
\textbf{Alignment-based methods} aim to bridge the gap between complete and incomplete inputs by aligning their latent representations.  
Contrastive objectives~\cite{li2024correlation,liu2024contrastive} encourage modality-invariant embeddings, while correlation-based techniques such as canonical correlation analysis and its extensions~\cite{hotelling1992relations,wang2015deep, andrew2013deep} learn cross-modal coherence.  
Other efforts incorporate masking or noise-based augmentation~\cite{yuan2023noise, pham2019MCTN} to simulate missingness and improve robustness.  
Although effective at aligning feature spaces, these methods depend on balanced modality exposure and struggle under heterogeneous missing conditions.   
\textbf{Generation-based methods} explicitly reconstruct missing modalities to recover full-modality representations.  
Autoencoder-based frameworks~\cite{tran2017CRA,ma2021smil} learn deterministic mappings between observed and missing modalities, while variational inference models~\cite{zhao2021MMIN,wang2023distribution} impose probabilistic constraints to improve generative consistency.  
Graph-based completion networks~\cite{lian2023gcnet,nguyen2025-micga} capture structural dependencies among modalities and instances, whereas diffusion-based approaches~\cite{wang2024incomplete,dai2025gamma,dai2025unbiased} leverage denoising priors to iteratively refine reconstructed signals.  
Despite their generative flexibility, these methods still assume identical missing probabilities across modalities and rarely consider imbalance during optimization.

\subsection{Imbalanced Multimodal Learning}
Imbalance in multimodal learning occurs when some modalities dominate training because they provide stronger or more reliable information~\cite{wang2020makes,wu2022characterizing,zhang2024multimodal}.  
This causes biased gradients and uneven representation learning, where dominant modalities are updated more effectively while weaker ones lag behind.
Prior studies address this issue through data re-sampling~\cite{wei2024enhancing}, adaptive feature calibration~\cite{zhang2024multimodal,ma2023calibrating}, objective regularization~\cite{xu2023mmcosine,ma2023calibrating,wei2024mmpareto}, and gradient modulation~\cite{li2023boosting,peng2022balanced,wei2024diagnosing,hua2024reconboost, nguyen2024ada2i}. 
These approaches enhance cooperation among modalities when all are fully available.  
Under incomplete conditions, imbalance worsens as low-availability modalities provide fewer samples and gradients, slowing convergence and degrading representations.    
The interplay between modality imbalance and missing data further complicates optimization and weakens cross-modal alignment.  
Our work addresses these issues by jointly calibrating features and rebalancing gradients for stable learning under heterogeneous missing conditions.

%% file: sec/3_methodology_v0.3.tex
\section{Methodology}

\begin{figure}[t!]
    \centering
    \includegraphics[width=1\linewidth]{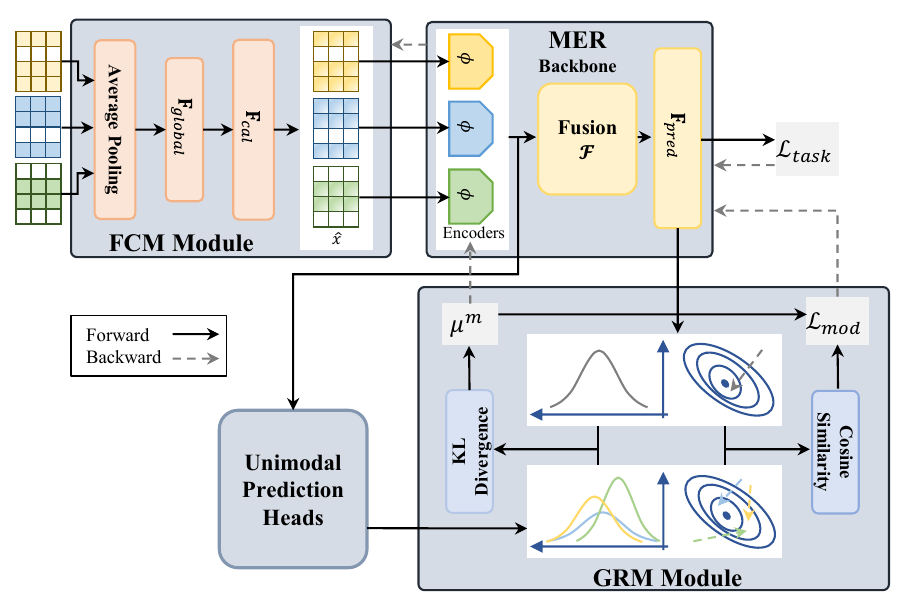}
    \caption{Overview of \myName{}, consisting of a Feature Calibration Module (\texttt{FCM}) and a Gradient Rebalancing Module (\texttt{GRM}) for balanced multimodal learning under imbalanced missing rates.}
    \label{fig:framework}
\end{figure}

Given a multimodal dataset $\mathbb{D}=\{(x_i, y_i)\}_{i=1}^{N}$ with $M$ modalities and label space $\mathcal{Y}$,
the multimodal feature of sample $i$ is denoted as 
$x_i=\{x_i^m\}_{m=1}^{M}$, where $x_i^m\in\mathbb{R}^{d_m}$ represents the feature of modality $m$.
To simulate missing-modality scenarios, we introduce a stochastic masking operator 
$\mathcal{M}(\cdot;\mathbf{r})$ parameterized by a missing-ratio vector 
$\mathbf{r}=[r_1,\!\dots\!,r_M]$, where $r_m\in[0,1)$ denotes the probability that modality $m$ is absent.
The missing mask vector of a multimodal sample is obtained via $
    e_i\sim \mathcal{M}(x_i;\mathbf{r})$, 
where each observation indicator $e^m_i\in\{0,1\}$ indicates the availability of modality $m$ for sample $i$. The indicator is generated with regards to the missing probability $r_m$ while ensuring at least one modality is available for each sample. The incomplete dataset is defined as
$\tilde{\mathbb{D}}=\{(\tilde{x}_i,y_i)\}^N_{i=1}$,
where $\tilde{x}_i$ is the incomplete multimodal sample with unimodal feature $\tilde{x}^m_i=x^m_i$ if $e^m_i=1$, and $\tilde{x}^m_i=\vec{0}$ if otherwise.

\subsection{Shared- and Imbalanced Missing Rate}
\label{sec:problem_formulation}
\textbf{Shared Missing Rate (SMR).}
In the shared setting, all modalities follow the same Bernoulli missingness pattern, where for each sample $i$ and modality $m$. The observation indicator $e_i^m\in\{0,1\}$ is drawn via
$P(e_i^m \!=\! 1)\!=\! 1 \!-\! r_{\text{shared}}$ and $P(e_i^m \!=\! 0) \!=\! r_{\text{shared}}$,
with $r_{\text{shared}}\in[0,1)$ denoting the shared missing rate. 
The joint masking distribution is:
\begin{equation}
p_{\text{SMR}}(e)
= \frac{\prod_{m=1}^{M} (1 - r_{\text{shared}})^{e^m} r_{\text{shared}}^{1 - e^m}}
       {1 - r_{\text{shared}}^{M}},
\label{eq:psmr}
\vspace{-3pt}
\end{equation}
where $e = \{e^m\}_{m=1}^{M}$ represents the binary masking pattern.
Here, the denominator normalizes by excluding the all-missing case.
For each modality $m$, the empirical missing ratio 
$\hat{r}_m = \frac{1}{N}\sum_{i=1}^{N} (1 - e_i^m)$ 
follows a Binomial distribution $B(N, r_{\text{shared}})$ 
with expectation and variance:
\begin{equation}
\mathbb{E}[\hat{r}_m] = r_{\text{shared}}, \qquad
\mathrm{Var}[\hat{r}_m] = \frac{r_{\text{shared}}(1 - r_{\text{shared}})}{N}.
\vspace{-5pt}
\end{equation}
As $N$ increases, $\hat{r}_m$ converges to 
$\mathcal{N}\!\big(r_{\text{shared}}, \tfrac{r_{\text{shared}}(1 - r_{\text{shared}})}{N}\big)$,
implying all modalities are statistically equivalent (see Fig.~\ref{fig:share-mr}), 
and any difference among $\hat{r}_m$ arises from stochastic fluctuation rather than systematic imbalance.

\textbf{Imbalanced Missing Rate (IMR).}
In realistic multimodal environments, 
different modalities exhibit heterogeneous missing probabilities 
($r_m \neq r_{m'}$) due to sensor reliability, recording noise, or acquisition cost.
The corresponding masking distribution becomes modality-specific:
\begin{equation}
p_{\text{IMR}}(e)
= \frac{\prod_{m=1}^{M} (1 - r_m)^{e^m} r_m^{1 - e^m}}
       {1 - \prod_{m=1}^{M} r_m},
\end{equation}
where $p_{\text{IMR}}(e)$ and $p_{\text{SMR}}(e)$ denote the probability of observing a particular masking pattern $e$ under IMR and SMR, respectively.
Unlike the homogeneous $p_{\text{SMR}}$, the heterogeneous $p_{\text{IMR}}$ 
assigns higher probability mass to combinations retaining low-$r_m$ modalities.
The degree of imbalance can be quantified by a divergence:
\begin{equation}
\Delta_{\text{IMR}}
= \mathcal{D}\!\big(p_{\text{IMR}}(e)\,\|\,p_{\text{SMR}}(e)\big),
\end{equation}
where $\mathcal{D}(\cdot\|\cdot)$ is a distributional divergence measure, e.g., KL or Jensen–Shannon distance. 
A larger $\Delta_{\text{IMR}}$ reflects stronger modality heterogeneity and sampling bias.

\textbf{From Distributional to Learning Imbalance.}
The heterogeneous masking distribution $p_{\text{IMR}}$ induces a persistent
\textit{distributional shift} between the training data distribution
$\mathbb{P}_{\text{train}}(\tilde{x},y)$ and the full-modality distribution
$\mathbb{P}_{\text{full}}(x,y)$, whose magnitude scales with $\Delta_{\text{IMR}}$.
This imbalance further propagates to the optimization process.
Given a backbone model $\mathcal{F}$ that produces prediction $\hat{y}_i=\mathcal{F}(\tilde{x}_i)$,
the learning objective under IMR can be formulated as:
\vspace{-5pt}
\begin{equation}
    \mathcal{L}_{\text{IMR}}
    = \mathbb{E}_{(\tilde{X},Y)\sim P_{\text{IMR}}}
      \Big[\sum_{m=1}^{M}(1-r_m)\,\ell(f_m(\tilde{X}_m),Y)\Big],
\end{equation}
where $\ell$ denotes the task loss for modality branch $m$.
The gradient contribution of each modality scales with its availability ratio $(1-r_m)$ is computed via:
\begin{equation}
    \label{eq:gradient-imbalance}\frac{\partial\mathcal{L}_{\text{IMR}}}{\partial\theta_m}
    \propto (1-r_m)\,
    \mathbb{E}\!\left[\frac{\partial\ell(f_m(\tilde{X}_m),Y)}{\partial\theta_m}\right].
\end{equation}
Consequently, modalities with higher missing ratios receive weaker supervision and converge more slowly,
resulting in \textbf{representation underfitting} and \textbf{optimization dominance} by low-missing-rate modalities. 
(See more in Suppl.B)

\textbf{Motivation.}
The above analysis reveals two coupled challenges under IMR:
(1) \textit{uneven exposure leading to inconsistent intra-modal representations}, and
(2) \textit{gradient-level inequality during optimization}.
To address these issues, we propose a model-agnostic framework that performs rebalancing at both the representation and optimization levels, as illustrated in Fig.~\ref{fig:framework}.

\subsection{Representation-level Rebalancing with Feature Calibration Module}
\label{sec:representation}

IMR introduces an imbalanced exposure pattern, where some modalities are observed far more often than others. This imbalance distorts the corresponding unimodal representations, weakening their semantic consistency and compromising cross-modal learning \cite{sun2024redcore}. Here, we introduce \textbf{Feature Calibration Module (\texttt{FCM})}
\label{sec:FCM}
to align modality-specific features that are distorted by uneven exposure under IMR.
Instead of imputing missing signals, it leverages global contextual cues aggregated from available modalities to recalibrate unimodal embeddings.

\textbf{Feature aggregation under IMR.}
Following~\cite{hwang2023self}, we obtain modality-wise descriptors 
to represent the overall statistics of available features. 
Given the masked features $\tilde{x}_i^m$ and binary indicators $e_i^m$
from the masking operator $\mathcal{M}(x;\mathbf{r})$, 
the unimodal global descriptor is computed by averaging available features:
\vspace{-5pt}
\begin{equation}
    x^m_{\textit{glob}} =
    \frac{\sum_{i=1}^{N} \tilde{x}^m_i}
         {\varepsilon + \sum_{i=1}^{N} e^m_i},
    \label{eq:avg-pool}
    \vspace{-3pt}
\end{equation}
where $x^m_{\textit{global}}\in\mathbb{R}^{d_m}$ and $\varepsilon$ is a smoothing hyper-parameter. 

\textbf{Cross-modal context extraction.}
The unimodal descriptors are concatenated and passed through a fully connected layer
to capture shared cross-modal context:
\begin{equation}
    x_{\textit{global}} =
    \mathbf{ReLU}\!\big(
    \mathbf{F}_{\textit{global}}([x^{m_1}_{\textit{glob}}, x^{m_2}_{\textit{glob}}, ..., x^{m_M}_{\textit{glob}}])
    \big),
\end{equation}
where $x_{\textit{global}}\in\mathbb{R}^{d_{\textit{global}}}$ encodes 
the global information jointly derived from all available modalities $\{m_j\}$. 
This representation is then projected to modality-specific calibration weights 
through independent fully connected layers:
\vspace{-5pt}

\begin{equation}
    w^m_{\textit{cal}} = \mathbf{F}^m_{\textit{cal}}(x_{\textit{global}}),
\end{equation}
where $w^m_{\textit{cal}}\in\mathbb{R}^{d_m}$ represents the contextual calibration weight for modality $m$.

\textbf{Feature calibration.}
Each modality feature is adaptively recalibrated through excitation-based calibration weights 
derived from global contextual information~\cite{hu2018squeeze}. These weights highlight the importance of each modality and help correct representation bias under IMR.
Formally, a gating operation with sigmoid activation is applied to rescale each unimodal feature:
\vspace{-5pt}
\begin{equation}
    \hat{x}^m_i = (1+\sigma(w^m_{\textit{cal}})) \odot \tilde{x}^m_i,
\end{equation}
where $\hat{x}^m_i\in\mathbb{R}^{d_m}$, 
$\odot$ and $\sigma(\cdot)$ denote element-wise multiplication and sigmoid function, respectively.
Features of missing modalities remain as $\vec{0}$.

Although the recalibrated features $\hat{x}^m_i$ differ in composition, 
they preserve dimensionality and seamlessly integrated into any backbone. 
This shared calibration mitigates IMR-induced representation discrepancies and provides a consistent feature basis for downstream optimization.

\subsection{Model-Agnostic Integration of Multimodal Emotion Recognition Backbone}
\label{sec:prediction}

For compatibility with existing Multimodal Emotion Recognition (MER) frameworks, we keep this stage unchanged. Each sample in MER corresponds to an utterance $u_i$ with multimodal inputs $\tilde{x}_i=\{\tilde{x}^a_i,\tilde{x}^v_i,\tilde{x}^l_i\}$ 
and an emotion label $y_i\!\in\!\mathcal{Y}$, where $a,v,l$ denote acoustic, visual, lexical modalities, respectively. 
The model predicts $\hat{y}_i$ representing the probability distribution over classes. 
Our plug-in modules work seamlessly with any backbone, treating missing modalities as $\vec{0}$ during training and inference. 

\textbf{Unimodal encoding.}
Each modality feature $\hat{x}_i^m$ from the Feature Calibration Module (FCM) 
is processed by a modality-specific encoder to obtain semantic embeddings:
\vspace{-5pt}
\begin{equation}
    z_i^m = \phi^m(\hat{x}_i^m),
    \label{eq:embedding}
\end{equation}
where $z_i^m \in \mathbb{R}^{d_{\textit{emb}}}$ denotes the embedding of modality $m$, 
and $\phi^m(\cdot)$ represents its encoder\footnote{In practice, $\phi^m(\cdot)$ is typically implemented using an LSTM, GRU, or multilayer perceptron (MLP) depending on the modality type.}.

\textbf{Multimodal fusion.}
The resulting unimodal embeddings are then combined through a fusion network:
\vspace{-2pt}
\begin{equation}
    h_i = \mathcal{F}([z_i^a, z_i^v, z_i^l]),
    \label{eq:fused}
    \vspace{-3pt}
\end{equation}
where $h_i \in \mathbb{R}^{d_h}$ is the fused multimodal representation, 
and $\mathcal{F}(\cdot)$ denotes the fusion function\footnote{$\mathcal{F}(\cdot)$ can take forms such as concatenation, attention, tensor fusion, or graph-based aggregation, depending on the backbone architecture.}.

\textbf{Prediction layer.}
The fused representation is mapped to the output space through a fully connected prediction head followed by a softmax layer:
\vspace{-3pt}
\begin{equation}
    \hat{y}_i = \delta(\mathbf{F}_{\textit{pred}}(h_i)),
    \label{eq:pred}
\end{equation}
where $\hat{y}_i \in \mathbb{R}^{|\mathcal{Y}|}$ is the predicted class probability distribution 
and $\delta(\cdot)$ denotes the softmax operator.

\textbf{Optimization objective.}
A task-specific loss $\ell(\cdot)$, typically Cross-Entropy or L1, 
is employed to measure the discrepancy between the predicted and ground-truth labels:
\vspace{-3pt}
\begin{equation}
    \mathcal{L}_{\textit{task}} 
    = \sum_{i=1}^{N} \ell(\hat{y}_i, y_i),
    \label{eq:loss}
    \vspace{-3pt}
\end{equation}
where $N$ is the number of utterances in a mini-batch.
This loss is used to optimize the parameters of the backbone via gradient-based updates:
\vspace{-5pt}
\begin{equation}
    \theta_{(t+1)} 
    = \theta_{(t)} - \alpha \frac{\partial \mathcal{L}_{\textit{task}}}{\partial \theta},
    \label{eq:upd}
    \vspace{-3pt}
\end{equation}
where $\theta$ denotes the model parameters, 
$t$ is the current training iteration, and $\alpha$ is the learning rate.

\subsection{Optimization-level Rebalancing with Gradient Rebalancing Module}
\label{sec:optimization}

Generally, popular modalities drive most gradient updates, resulting in skewed convergence ~\cite{guo2024classifier}. Here, we propose \textbf{Gradient Rebalancing Module \texttt{(GRM)}} to balance learning dynamics across modalities, in which each modality-specific component receives fair and consistent supervision under IMR.
We use the latest unimodal embeddings as optimization targets, while their encoders act as modality-specific components.

\paragraph{Unimodal Prediction Heads.} 
\label{sec:pred-m}
To quantify optimization discrepancies among modalities, 
we introduce lightweight unimodal prediction heads that operate independently.
Each head consists of two fully connected layers applied to the embeddings from Eq.~\ref{eq:embedding}. 
The first layer projects the embedding dimension to that of the multimodal representation in Eq.~\ref{eq:fused}, 
followed by a ReLU activation as follows:
\vspace{-3pt}
\begin{equation}
    h^m_i = \mathbf{ReLU}\!\big(\mathbf{F}^m_{\textit{map}}(z^m_i)\big),
    \vspace{-3pt}
\end{equation}
where $h^m_i \in \mathbb{R}^{d_h}$.
The second layer produces modality-specific predictions using the same formulation as Eq.~\ref{eq:pred}:
\vspace{-3pt}
\begin{equation}
    \hat{y}^m_i = \delta\!\big(\mathbf{F}^m_{\textit{pred}}(h^m_i)\big),
    \label{eq:pred-m}
\end{equation}
where $\hat{y}^m_i \in \mathbb{R}^{|\mathcal{Y}|}$ and $\delta(\cdot)$ denotes the softmax operator.
We deliberately design these two-layer classifiers to ensure that 
the gradient matrices of all modality-specific prediction layers $\mathbf{F}^m_{\textit{pred}}$ 
share the same shape $(d_h \times |\mathcal{Y}|)$ as the framework’s main prediction layer $\mathbf{F}_{\textit{pred}}$. 
Each unimodal head is optimized using the same loss function $\ell(\cdot)$ as in Eq.~\ref{eq:loss}, where the task-specific loss for modality $m$ is:
\vspace{-3pt}
\begin{equation}
\begin{aligned}
\mathcal{L}^m_{\textit{task}} &= \sum_{i=1}^{N} \ell(\hat{y}^m_i, y_i), &
\theta^m_{(t+1)} &= \theta^m_{(t)} - \alpha \frac{\partial \mathcal{L}^m_{\textit{task}}}{\partial \theta^m}.
\label{eq:uni-loss,upd}
\end{aligned}
\end{equation} where $\theta^m$ is updated via standard gradient descent.

\paragraph{Gradient Rebalancing Module (\texttt{GRM}).}
\label{sec:grm}

This module mitigates the optimization bias introduced by imbalanced missing modalities. 
It adjusts gradient updates of modality-specific encoders from two complementary perspectives: 
\textit{distribution-driven} and \textit{spatial-driven}. 

\textbf{Distribution-driven modulation.}
Since complete unimodal distributions are unobservable under IMR, modality contributions become skewed as dominant modalities overshadow weaker ones.
We therefore approximate the multimodal prediction distribution as a stable reference capturing intra- and inter-modal information.
To measure the discrepancy between unimodal predictions and the multimodal reference,
we use the Kullback–Leibler (KL) divergence as:
\vspace{-4pt}
\begin{equation}
    \mathcal{D}^m_{\mathrm{KL}} 
    = \sum_{i=1}^{N} 
      \mathbf{KL}\big(\hat{y}^m_i \,\|\, \hat{y}_i\big),
      \label{eq:kl-div}
      \vspace{-3pt}
\end{equation}
where $\hat{y}_i$ and $\hat{y}^m_i$ denote the multimodal and unimodal predicted class probability distribution 
from Eq.~\ref{eq:pred} and Eq.~\ref{eq:pred-m}, respectively.
A larger $\mathcal{D}^m_{\mathrm{KL}}$ indicates slower convergence of modality $m$ toward the shared multimodal distribution.

Absolute divergence values are insufficient for tracking learning dynamics, 
as dominant modalities stabilize quickly while others fluctuate. 
We thus compute the \textit{relative learning progress} of each modality 
from the change in divergence between consecutive iterations:
\vspace{-3pt}
\begin{equation}
    \Delta^m_{\mathrm{KL}} =
    \begin{cases}
        \mathcal{D}^{m^{(t)}}_{\mathrm{KL}}, & t = 0,\\
        \mathcal{D}^{m^{(t-1)}}_{\mathrm{KL}} - \mathcal{D}^{m^{(t)}}_{\mathrm{KL}}, & \text{otherwise},
    \end{cases}
    \label{eq:kl-diff}
\end{equation}
A larger $\Delta^m_{\mathrm{KL}}$ indicates faster alignment with the multimodal reference.
Based on these progress signals, we define a \textit{modulation coefficient} for each modality:
\begin{equation}
    \mu^m = \rho 
    \frac{\sum_{m' \in \{a,v,l\}, m' \neq m} \Delta^{m'}_{\mathrm{KL}}}
         {\sum_{m' \in \{a,v,l\}} \Delta^{m'}_{\mathrm{KL}}},
    \label{eq:coef}
    \vspace{-5pt}
\end{equation}
where $\rho$ is a saturation hyperparameter controlling the modulation intensity.
A smaller $\mu^m$ indicates that modality $m$ is learning faster (larger $\Delta^m_{\mathrm{KL}}$), 
and thus its gradient update should be attenuated; 
conversely, slower modalities receive stronger updates.
Accordingly, the parameters of unimodal encoders are updated as:
\vspace{-3pt}
\begin{equation}
    \theta^{\phi^m}_{(t+1)} 
    = \theta^{\phi^m}_{(t)} 
    - \alpha\, \mu^m 
      \frac{\partial \mathcal{L}_{\mathrm{task}}}{\partial \theta^{\phi^m}_{(t)}},
    \label{eq:enc-mod}
    \vspace{-3pt}
\end{equation}
where $\theta^{\phi^m}$ denotes the parameters of encoder $\phi^m$.

\textbf{Spatial-driven modulation.}
\label{sec:spatial}
Previous work~\cite{wu2022characterizing, zhang2024multimodal, wang2020makes} 
shows that multimodal optimization follows the dominant modality. 
This imbalance worsens under missing modalities, 
so we reorient gradients toward a balanced multimodal direction to stabilize training.
Directly computing full unimodal gradients is computationally expensive,
and comparing them against the entire model’s gradients is often infeasible. 
Following~\cite{guo2024classifier}, 
we approximate the overall model and modality-specific gradients 
using those of the multimodal and unimodal prediction heads, respectively:
\vspace{-4pt}
\begin{equation}
\nabla_{\textit{pred}} 
= \nabla_{\theta^{\mathbf{F}_{\textit{pred}}}} \mathcal{L}_{\textit{task}}
= \frac{\partial \mathcal{L}_{\textit{task}}}{\partial \theta^{\mathbf{F}_{\textit{pred}}}},
\label{eq:grad-fuse}
\end{equation}
\begin{equation}
\nabla^m_{\textit{pred}} 
= \nabla_{\theta^{\mathbf{F}^m_{\textit{pred}}}} \mathcal{L}^m_{\textit{task}}
= \frac{\partial \mathcal{L}^m_{\textit{task}}}{\partial \theta^{\mathbf{F}^m_{\textit{pred}}}},
\label{eq:grad-m}
\end{equation}
where $\nabla_{\textit{pred}}$ represents the gradient of the main multimodal classifier 
and $\nabla^m_{\textit{pred}}$ corresponds to the gradient of modality $m$'s prediction head.
\begin{table*}[t!]
\centering
\caption{Performance on IEMOCAP under different IMR configurations.}
\label{tab:res-iemocap-mean}
\input{material/res-iemocap-mean}
\end{table*}
We measure the spatial discrepancy between these gradients via cosine similarity:
\begin{equation}
    \mathcal{D}^m_{\mathrm{cos}} 
    = \text{cos}_{sim}\!\big(\nabla^m_{\textit{pred}},\, \nabla_{\textit{pred}}\big),
\end{equation}
A lower cosine similarity indicates that modality $m$ 
is optimized in a direction less aligned with the multimodal objective. 
To avoid over-constraint, we adopt a soft spatial alignment strategy, which minimizes the weighted absolute deviation of cosine similarities:

\vspace{-4pt}
\begin{equation}
    \mathcal{L}_{\textit{mod}}
    = \sum_{m\in\{a,v,l\}} 
      \big|\,\mathcal{D}^m_{\mathrm{cos}} \times \mu^m\,\big|,
    \label{eq:loss-mod}
\end{equation}
where $\mu^m$ are dynamic modulation coefficients from Eq.~\ref{eq:coef},
adaptively weighting each modality’s contribution during spatial alignment under IMR.

\paragraph{Overall Objective.}
\label{sec:overall}
The overall training objective integrates the task loss and the modulation loss as:
\vspace{-5pt}

\begin{equation}
    \mathcal{L}
    = \mathcal{L}_{\textit{task}} 
    + \tau\,\mathcal{L}_{\textit{mod}},
    \label{eq:ovr-loss}
\end{equation}
where $\tau$ is a trade-off hyperparameter. 

The modality-specific classifiers introduced in Sec.~\ref{sec:pred-m} 
are optimized independently using $\mathcal{L}^m_{\textit{task}}$ as auxiliary objectives 
and are therefore excluded from Eq.~\ref{eq:ovr-loss}.
The complete training procedure is summarized in Suppl.~\ref{supl:training}.

%% file: material/res-iemocap-mean.tex
\vspace{-3pt}
\resizebox{\linewidth}{!}{
\begin{tabular}{lcccccccccccc}
\toprule
\multirow{2}{*}{\textbf{Model}} &
   \multicolumn{2}{c}{(0.3,0.5,0.7)} &
   \multicolumn{2}{c}{(0.3,0.7,0.5)} &
   \multicolumn{2}{c}{(0.5,0.3,0.7)} &
   \multicolumn{2}{c}{(0.5,0.7,0.3)} &
   \multicolumn{2}{c}{(0.7,0.3,0.5)} &
   \multicolumn{2}{c}{(0.7,0.5,0.3)} \\ 
\cmidrule(lr){2-13}
 & Acc. & w-F1 & Acc. & w-F1 & Acc. & w-F1 & Acc. & w-F1 & Acc. & w-F1 & Acc. & w-F1 \\ 
\midrule
\rowcolor{gray!10}\multicolumn{13}{c}{\textbf{\textit{Group 1 (Models addressing missing or imbalanced modalities)}}}\\
MMIN \cite{zhao2021MMIN}    & 55.97 & 49.92 & 55.87 & 49.83 & 56.40 & 50.28 & 55.87 & 49.75 & 55.87 & 49.85 & 55.83 & 49.73 \\
SDR-GNN \cite{fu2024sdr} & 59.46 & 59.34 & 54.16 & 53.59 & 58.53 & 58.42 & 56.56 & 56.77 & 55.14 & 55.22 & 55.45 & 55.17 \\
Mi-CGA \cite{nguyen2025-micga}  & 58.84 & 58.80 & 54.65 & 54.61 & 60.50 & 60.25 & 56.19 & 56.11 & 58.84 & 58.63 & 58.72 & 58.22 \\

MoMKE \cite{xu2024leveraging}   & 55.39 & 54.19 & 50.83 & 49.70 & 60.18 & 59.94 & 50.77 & 50.15 & 58.26 & 58.45 & 54.78 & 54.92 \\
GCNet \cite{lian2023gcnet}   & 58.04 & 58.36 & 56.81 & 56.83 & 61.61 & 62.06 & 56.81 & 56.99 & 59.64 & 59.78 & 59.09 & 59.19 \\\hdashline[1pt/2pt]
Ada2I \cite{nguyen2024ada2i} & 61.98 & 60.68 & 61.98 & 61.46 & 61.37 & 60.45 & 58.10 & 58.22 & 61.74 & 60.90 & 57.73 & 54.76 \\
RedCore \cite{sun2024redcore} & 46.27 & 46.32 & 43.38 & 43.40 & 50.15 & 47.92 & 43.69 & 43.60 & 50.59 & 50.53 & 46.09 & 46.21 \\
MCE \cite{zhao2025mce} & 46.52 & 45.54 & 44.36 & 43.48 & 50.65 & 49.22 & 42.64 & 41.70 & 50.34 & 49.45 & 44.73 & 44.40 \\
\midrule
\rowcolor{blue!10}\textbf{GCNet\textsubscript{+BALM}} & {61.31}\textsubscript{$\pm0.75$} & {61.33}\textsubscript{$\pm0.70$} & {57.36}\textsubscript{$\pm0.76$} & {57.00}\textsubscript{$\pm0.83$} & {61.31}\textsubscript{$\pm0.63$} & {61.62}\textsubscript{$\pm0.59$} & {57.64}\textsubscript{$\pm1.10$} & {57.64}\textsubscript{$\pm1.22$} & {59.33}\textsubscript{$\pm0.46$} & {59.39}\textsubscript{$\pm0.50$} & {60.26}\textsubscript{$\pm0.83$} & {60.58}\textsubscript{$\pm0.96$} \\
\footnotesize\textit{$\Delta$ (vs origin GCNet)} & \footnotesize\textit{+3.27} & \footnotesize\textit{+2.97} & \footnotesize\textit{+0.55} & \footnotesize\textit{+0.17} & \footnotesize\textit{-0.30} & \footnotesize\textit{-0.44} & \footnotesize\textit{+0.83} & \footnotesize\textit{+0.65} & \footnotesize\textit{-0.31} & \footnotesize\textit{-0.39} & \footnotesize\textit{+1.17} & \footnotesize\textit{+1.39} \\
\footnotesize\textit{$\Delta_1$ (vs Grp1 mean)} & \footnotesize\textit{+6.00} & \footnotesize\textit{+7.19} & \footnotesize\textit{+4.61} & \footnotesize\textit{+5.39} & \footnotesize\textit{+4.21} & \footnotesize\textit{+5.55} & \footnotesize\textit{+5.06} & \footnotesize\textit{+5.98} & \footnotesize\textit{+3.03} & \footnotesize\textit{+4.04} & \footnotesize\textit{+6.21} & \footnotesize\textit{+7.76} \\\hline

\rowcolor{gray!10}\multicolumn{13}{c}{\textbf{\textit{Group 2 (Typical MER Models)}}}\\
MMGCN \cite{hu-etal-2021-mmgcn} & 63.59 & 63.62 & 59.09 & 58.99 & 64.63 & 64.95 & 57.79 & 57.99 & 64.14 & 64.27 & 60.63 & 60.85 \\
\rowcolor{blue!10}\textbf{MMGCN\textsubscript{+BALM}} & {64.39}\textsubscript{$\pm0.88$} & {63.82}\textsubscript{$\pm0.89$} & {60.07}\textsubscript{$\pm0.70$} & {60.04}\textsubscript{$\pm0.65$} & {65.37}\textsubscript{$\pm0.61$} & {65.03}\textsubscript{$\pm0.63$} & {62.29}\textsubscript{$\pm0.75$} & {62.35}\textsubscript{$\pm0.70$} & {65.00}\textsubscript{$\pm1.54$} & {64.98}\textsubscript{$\pm1.77$} & {61.74}\textsubscript{$\pm0.96$} & {61.20}\textsubscript{$\pm0.82$} \\
\footnotesize\textit{$\Delta$ (vs origin MMGCN)} & \footnotesize\textit{+0.80} & \footnotesize\textit{+0.20} & \footnotesize\textit{+0.98} & \footnotesize\textit{+1.05} & \footnotesize\textit{+0.74} & \footnotesize\textit{+0.08} & \footnotesize\textit{+4.50} & \footnotesize\textit{+4.36} & \footnotesize\textit{+0.86} & \footnotesize\textit{+0.71} & \footnotesize\textit{+1.11} & \footnotesize\textit{+0.35} \\
\footnotesize\textit{$\Delta_1$ (vs Group1 mean)}  & \footnotesize\textit{+9.08} & \footnotesize\textit{+9.68} & \footnotesize\textit{+7.32} & \footnotesize\textit{+8.43} & \footnotesize\textit{+7.95} & \footnotesize\textit{+8.96} & \footnotesize\textit{+9.71} & \footnotesize\textit{+10.69} & \footnotesize\textit{+8.70} & \footnotesize\textit{+9.63} & \footnotesize\textit{+7.69}
& \footnotesize\textit{+8.38}\\
\hline
MMDFN \cite{hu2022mm} & 63.34 & 63.00 & 60.38 & 60.01 & 67.34 & 67.40 & 57.92 & 56.25 & 66.48 & 66.67 & 63.65 & 64.00 \\


\rowcolor{blue!10}\textbf{MMDFN\textsubscript{+BALM}} & {67.34}\textsubscript{$\pm0.84$} & {66.64}\textsubscript{$\pm0.94$} & {63.40}\textsubscript{$\pm0.45$} & {63.11}\textsubscript{$\pm0.47$} & {68.76}\textsubscript{$\pm0.22$} & {68.25}\textsubscript{$\pm0.18$} & {64.02}\textsubscript{$\pm1.11$} & {63.76}\textsubscript{$\pm1.05$} & {67.22}\textsubscript{$\pm1.03$} & {67.02}\textsubscript{$\pm1.06$} & {65.50}\textsubscript{$\pm0.47$} & {64.98}\textsubscript{$\pm0.64$} \\
\footnotesize\textit{$\Delta$ (vs origin MMDFN)} & \footnotesize\textit{+4.00} & \footnotesize\textit{+3.64} & \footnotesize\textit{+3.02} & \footnotesize\textit{+3.10} & \footnotesize\textit{+1.42} & \footnotesize\textit{+0.85} & \footnotesize\textit{+6.10} & \footnotesize\textit{+7.51} & \footnotesize\textit{+0.74} & \footnotesize\textit{+0.35} & \footnotesize\textit{+1.85} & \footnotesize\textit{+0.98} \\
\footnotesize\textit{$\Delta_1$ (vs Group1 mean)} & \footnotesize\textit{+12.03} & \footnotesize\textit{+12.50} & \footnotesize\textit{+10.65} & \footnotesize\textit{+11.50} & \footnotesize\textit{+11.34} & \footnotesize\textit{+12.18} & \footnotesize\textit{+11.44} & \footnotesize\textit{+12.10} & \footnotesize\textit{+10.92} & \footnotesize\textit{+11.67} & \footnotesize\textit{+11.45} & \footnotesize\textit{+12.16} \\
\bottomrule
\end{tabular}}
\vspace{-7pt}
\parbox{\linewidth}{
  \raggedleft
\scriptsize \textit{Note.} $(r_A, r_L, r_V)$ denotes the missing rates for Audio, Language, and Visual modalities.\\
$\Delta$ and $\Delta_1$ denote performance differences relative to the original backbone and mean of Group~1 models, respectively.
}

%% file: sec/4_experiment_setting.tex
\section{Experiments}
\subsection{Experimental Settings}

\paragraph{Datasets and Evaluation Metrics.}  
We conduct experiments on two  multimodal emotion recognition datasets: \textbf{IEMOCAP}~\cite{busso2008iemocap} (\textit{6-way}) and \textbf{CMU-MOSEI}~\cite{zadeh2018mosei}, each containing audio (A), visual (V), and lexical (L) modalities.   
We adopt Accuracy (\textit{Acc.}) and Weighted F1-score (\textit{w-F1}) as evaluation metrics to  assess model performance.

\vspace{-10pt}
\paragraph{Baselines.}
We evaluate \myName{} with two main groups of baselines:
(1)~\textbf{Models addressing missing or imbalanced modalities}, including
\textit{MMIN}~\cite{zhao2021MMIN},
\textit{SDR-GNN}~\cite{fu2024sdr},
\textit{Mi-CGA}~\cite{nguyen2025-micga},
\textit{MoMKE}~\cite{xu2024leveraging}, and
\textit{GCNet}~\cite{lian2023gcnet} for missing-modality learning;
\textit{Ada2I}~\cite{nguyen2024ada2i} for modality imbalance;
\textit{RedCore}~\cite{sun2024redcore} and \textit{MCE}~\cite{zhao2025mce} which jointly handles both aspects;
and (2) \textbf{Typical MER backbones},
\textit{MMGCN}~\cite{hu-etal-2021-mmgcn} and
\textit{MMDFN}~\cite{hu2022mm}, representing standard multimodal architectures.
All models are trained from scratch, with missing modalities simulated under shared (\textit{SMR}) or imbalanced (\textit{IMR}) rates defined in Sec.~\ref{sec:problem_formulation}.

Our plug-in (\textbf{+BALM}) is integrated without altering backbone architectures and supports both early- and late-fusion setups.
This setup enables evaluating \myName{} across incomplete, imbalanced, and standard MER backbones.
Further dataset statistics, baseline descriptions, and implementation details are provided in the Suppl.~\ref{supl:implement}.

%% file: sec/5_results.tex

\begin{table*}[t!]
\centering
\caption{Performance on CMU-MOSEI under different IMR configurations.}
\label{tab:res-mosei-mean}
\input{material/res-mosei-mean}
\end{table*}
\subsection{Main Results}
\paragraph{Performance Comparison.}
Tables~\ref{tab:res-iemocap-mean} and~\ref{tab:res-mosei-mean} present results under different IMR settings with $(r_A, r_L, r_V)\!\in\!\{0.3,0.5,0.7\}$, where integrating \textbf{\myName } consistently improves all metrics across all backbones.  

As shown in Table~\ref{tab:res-iemocap-mean}, integrating BALM consistently boosts both missing-modality and emotion-recognition baselines.  
For example, GCNet\textsubscript{+\myName} yields a gain of up to $+3.27$\% Acc over the original GCNet on the $(0.3,0.5,0.7)$ setting. It consistently outperforms other missing-modality baselines. Under the $(0.7,0.5,0.3)$ configuration, it improves w-F1 by roughly $+7.0$\% compared with the \textit{Group~1} mean. These results confirm its effectiveness in reducing bias caused by uneven modality availability.
Compared with recent IMR-oriented works of \textit{Group~1}, which explicitly model missing patterns, stronger fusion backbones integrated with BALM, notably MMDFN\textsubscript{+\myName} and MMGCN\textsubscript{+\myName}, achieve even higher performance across different missing scenarios. In addition, \myName elevates both MER models in all missing configurations (up to $7.51\%$ w-F1 gap over the original backbone of MMDFN\textsubscript{+\myName}), and raises their performance lower-bound to above $60\%$ for both Acc and w-F1.
These results suggest that rebalancing, representation- and optimization-wise, provides complementary benefits for high-capacity fusion architectures.  
Ada2I, which addresses imbalance learning for MER, performs better than conventional missing-modality models but still degrades under IMR, showing that gradient balancing alone is insufficient when missingness occurs.  
\begin{figure}[t!]
\centering
    \includegraphics[width=\linewidth]{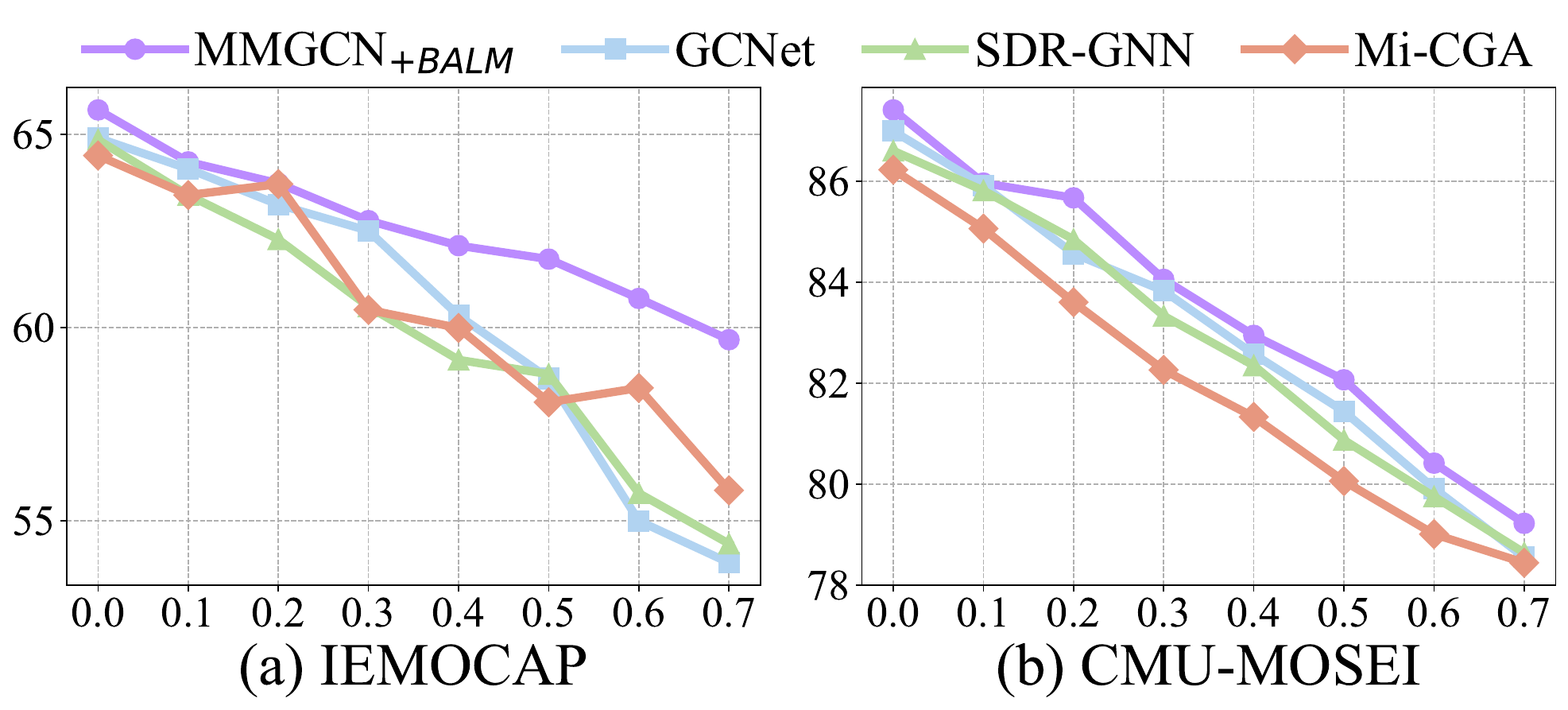}
    \caption{Average accuracy under different SMR settings}
    \label{fig:smr-imr}
\end{figure}

Likewise, on CMU-MOSEI (Table~\ref{tab:res-mosei-mean}), \myName consistently improves performance across IMR settings with smaller but steady gains, reflecting the dataset’s lower modality noise and weaker sensitivity to missing-rate variations. Persistent with the observation in the previous dataset, high language-missing rates in CMU-MOSEI lead to higher performance degradation, and in return, such configurations highlight the effectiveness of \myName in elevating backbones' performance. Specifically, MMGCN\textsubscript{+\myName} achieves a $+1.71\%$ Acc gain over the original MMGCN, widen the gap with \textit{Group~1} mean to $+4.12\%$ Acc under $(0.5,0.7,0.3)$ setting, a similar boost can also be found on GCNet\textsubscript{+\myName} under the same setting. 

For both dataset, moderate to high loss of visual (e.g., $(0.7,0.3,0.5), (0.5,0.3,0.7)$) usually leads to lower degradation comparing to other modalities. Whereas, high missing rates in language (e.g., $(0.3,0.7,0.5)$) lead to profound performance drops, confirming that lexical information is the most discriminative yet vulnerable modality. When audio missingness is also high alongside language (e.g., $(0.5,0.7,0.3)$), the degradation becomes more pronounced, underscoring the critical role of robust auditory cues. These divergent behaviors when modalities take turn suffering low/high missing rate, along with the consistent gains of the backbones after integrating with \myName, further validate the need to address modalities' discrepancy stem from IMR.
\paragraph{Evaluating Robustness under SMR Settings.}
To ensure a fair cross-evaluation, we adopt the same settings and model configurations as in the IMR experiments, while uniformly applying identical missing rates to all modalities (SMR).  
As shown in Fig.~\ref{fig:smr-imr}, integrating \textbf{BALM} consistently outperforms all baselines under SMR settings,\footnote{Each result is averaged over five independent runs with different random seeds for each shared missing rate.} 
indicating that our approach not only handles the challenging IMR scenario but also maintains superior robustness and generalization under balanced missing conditions.

\begin{table}[t!]
    \centering
    \caption{Average performance on different modality combinations.
    }
    \label{tab:FCM-analysis}
    \input{material/abl-FCM}
\end{table}
\begin{figure}[t!]
    \centering
    \includegraphics[width=\linewidth]{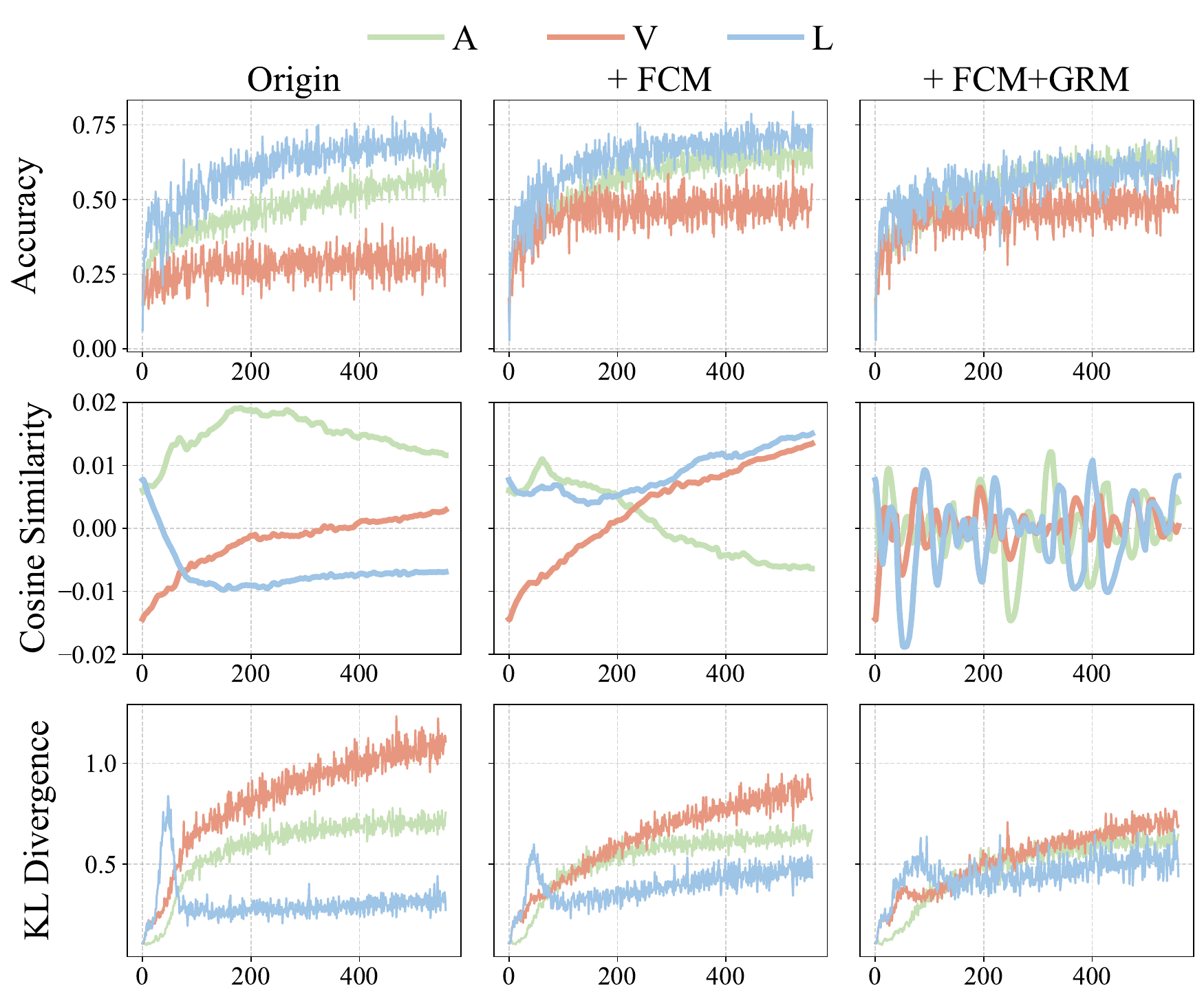}
    \caption{Modality discrepancies on IEMOCAP with MMGCN.
}
    \label{fig:GRM-analysis}
    
\vspace{-7pt}
\end{figure}
\subsection{Ablation Study}
\paragraph{Effect of Rebalancing Modules.}
Table~\ref{tab:FCM-analysis} summarizes the average performance over modality combinations under the IMR setting $(r_A,r_L,r_V)\!=\!(0.5,0.7,0.3)$. 
\texttt{FCM} consistently improves results on both datasets, boosting w-F1 by $+2.81\%$ on IEMOCAP and $+0.92\%$ on MOSEI.

As illustrated in Fig.~\ref{fig:GRM-analysis}, \texttt{FCM} effectively enhances feature-level consistency by reducing representation discrepancies among modalities; however, it does not fully resolve the imbalance in gradient magnitudes across branches. 
Building on this, \texttt{GRM} further rebalances the gradient flow.
In particular, the lexical branch tends to dominate optimization, resulting in large cross-modal divergence. 
With \texttt{GRM}, the inter-modal cosine similarity increases and the KL divergence stabilizes, indicating that gradient rebalancing harmonizes multimodal optimization and alleviates dominance from high-availability modalities.


\paragraph{Modality Combination.}

\begin{figure}[t!]
    \centering
    \includegraphics[width=\linewidth]{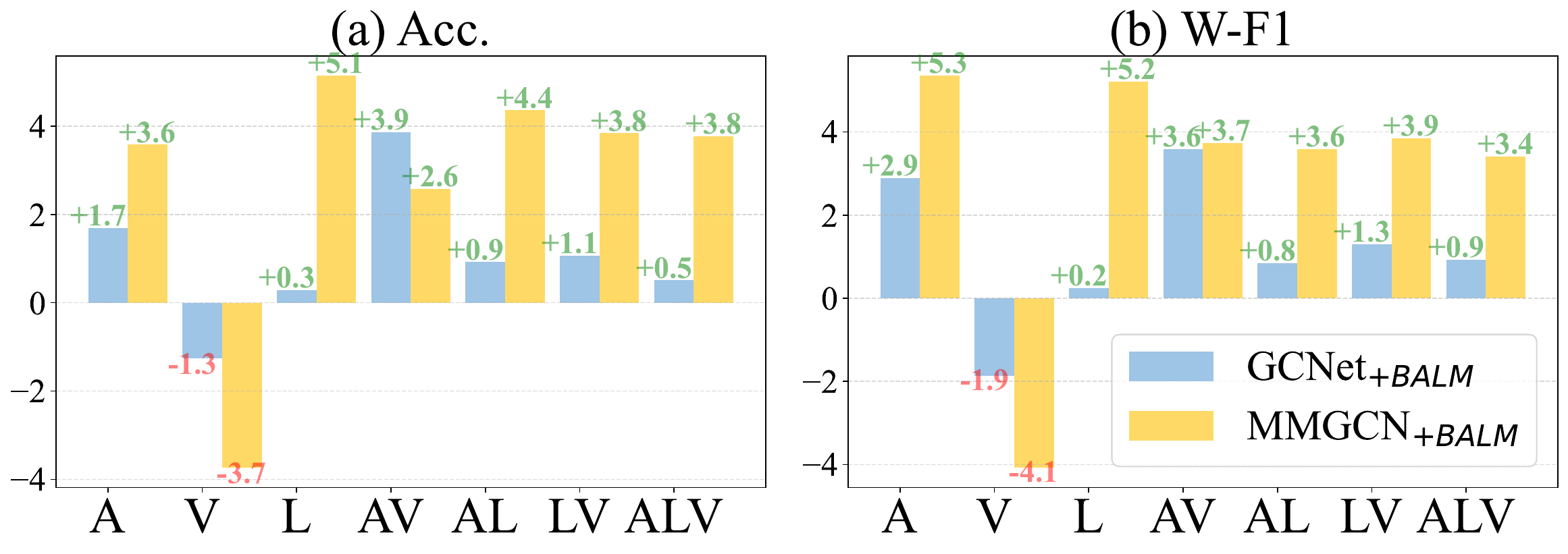}
    \caption{Improvement of GCNet$_{+BALM}$ and MMGCN$_{+BALM}$ comparing to GCNet on IEMOCAP}\vspace{-9pt}
    \label{fig:modal}
\end{figure}

Fig.~\ref{fig:modal} shows that BALM consistently improves both GCNet and MMGCN across most modality combinations on IEMOCAP, with the largest gains on audio and lexical branches (up to $+5.3\%$ w-F1).  
Slight drops in settings where only the visual modality is available stem partly from the low quality of visual cues in IEMOCAP and partly from the fusion architecture of each baseline.  
Fusion-based methods like MMGCN were originally developed for MER and use modality-specific encoders and fusion mechanisms, which makes them more sensitive to variations in modality quality. As a result, they experience larger performance drops.  
In contrast, GCNet incorporates mechanisms to handle missing modalities and is more alignment-aware, enabling it to capture modality-specific cues and suffer less degradation under the same conditions.

\begin{figure}[t!]
    \centering
    \begin{subfigure}{0.48\linewidth}
        \centering
        \includegraphics[width=\linewidth]{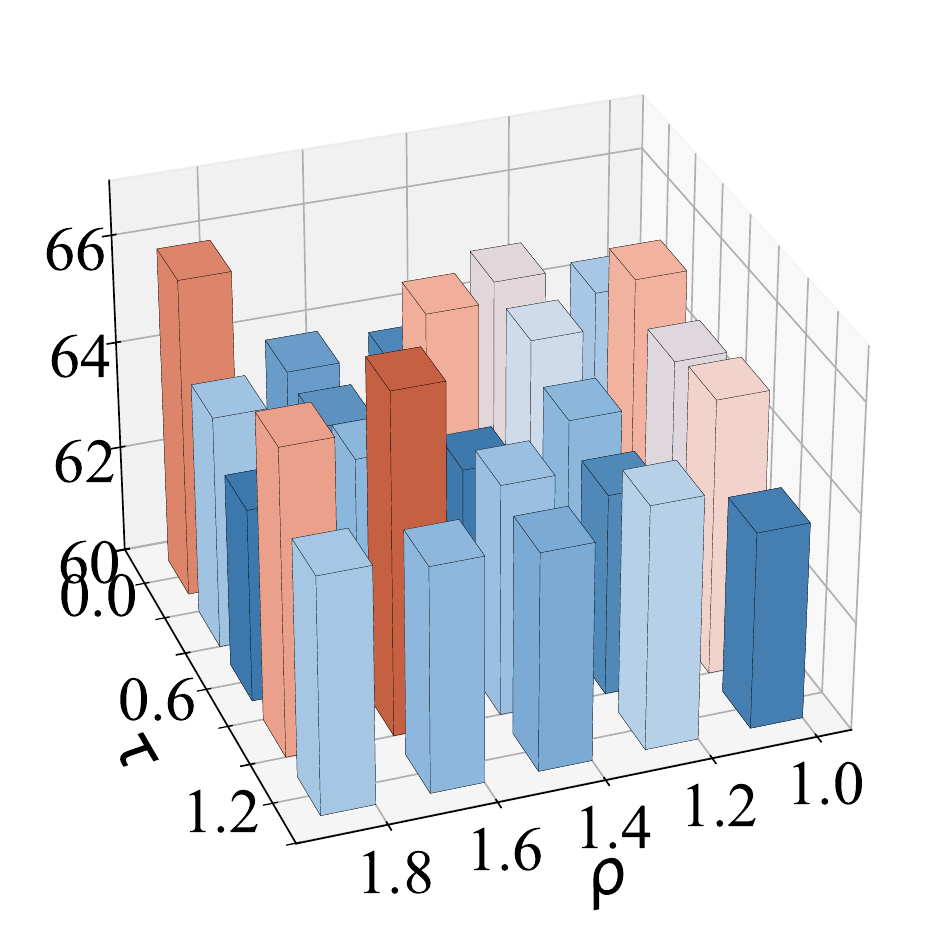}
        \caption{ \textit{Config-A}: $(0.5, 0.3, 0.7)$}
        \label{fig:tune1}
    \end{subfigure}
    \hfill
    \begin{subfigure}{0.48\linewidth}
        \centering
        \includegraphics[width=\linewidth]{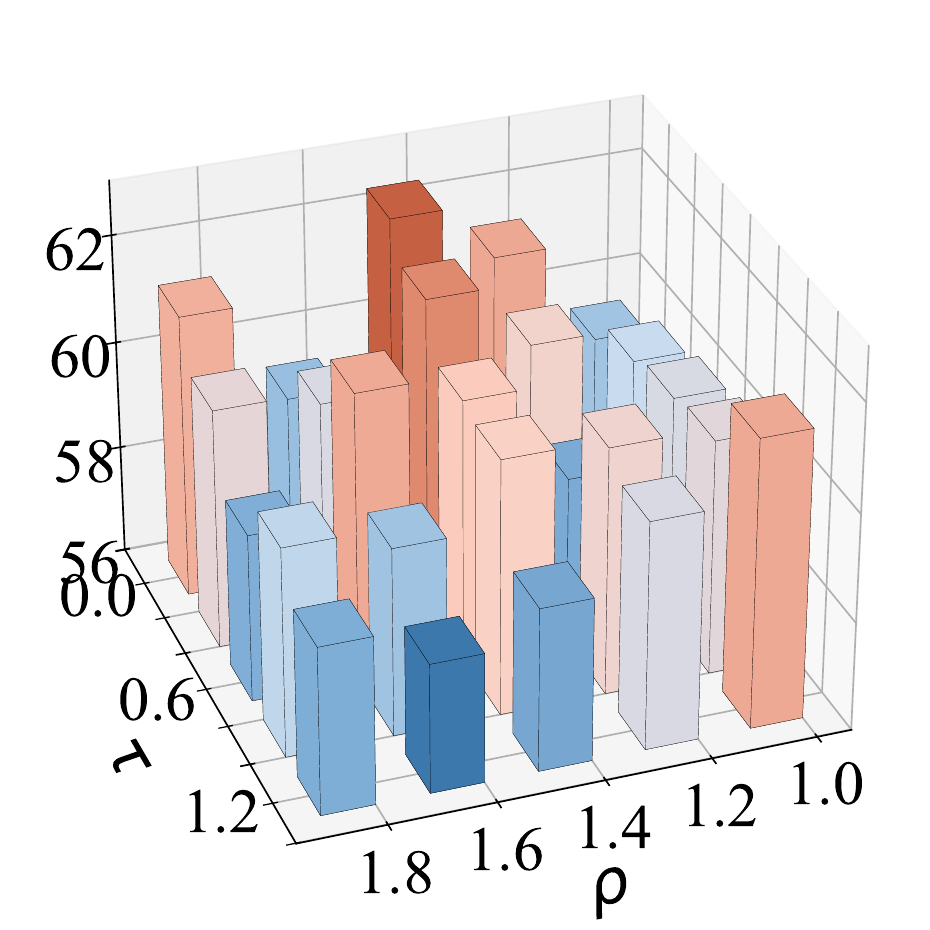}
        \caption{ \textit{Config-B}: $(0.5, 0.7, 0.3)$}
        \label{fig:tune2}
    \end{subfigure}
    \caption{Weighted-F1 sensitivity to hyperparameters under two missing-rate configurations of audio, language and visual.
    }

    \label{fig:tune_all}
\end{figure}

\vspace{-8pt}
\paragraph{Hyperparameter Sensitivity under Contrasting Missing Rates.}
Fig.~\ref{fig:tune_all} compares w-F1 of MMGCN\textsubscript{+\myName} under two contrasting missing-rate settings on IEMOCAP: 
\textit{Config--A:} $(0.5,0.3,0.7)$ and \textit{Config--B:} $(0.5,0.7,0.3)$. 
In \textit{Config--A}, where the least missing-sensitive modality, i.e. visual, is highly missing, w-F1 is stable around $63-64\%$ despite the change in $\tau,\rho$. 
Conversely, in \textit{Config--B}, where the more vulnerable modality, i.e. lexical, suffers highest missing rate, the imbalance-aware regularization yields more distinct effectiveness, as higher performance can be achieved with proper selection of $\rho$ or $\tau$. 
These results underscore the importance of \textbf{adaptive regulation} for imbalanced missing-modality scenarios.

\paragraph{Severe Missing Rate Evaluation.}
Table~\ref{tab:stress-test} presents the robustness evaluation under extreme missing-modality conditions.
As audio and language are informative modalities in MER, their absence typically leads to notable degradation.
Nevertheless, BALM consistently stabilizes performance across settings.
For instance, when 90\% of the language modality is missing, MMGCN\textsubscript{+BALM} achieves a $+4.78\%$ accuracy gain on IEMOCAP compared to the mean of baselines (eg. GCNet, MMGCN).
Under heavy audio missing 
$(0.9,0.6,0.6)$, it still improves by $+2.72\%$ on accuracy.
Although a few settings exhibit minor performance drops, the overall trend indicates BALM effectively mitigates imbalance effects and maintains robustness even when informative modalities are severely incomplete.

\begin{table}[t!]
    \centering
    \caption{Performance of MMGCN with BALM under severe missing-modality settings, where the arrows indicate the relative change compared to the mean performance of the two baselines. 
}

    \label{tab:stress-test}
    \input{material/abl-stress-test}
\end{table}

\vspace{-3pt}
\section{Conclusion}
\vspace{-3pt}
This paper presents \myName{}, a lightweight and model-agnostic framework for multimodal learning under imbalanced missing conditions. 
By combining the Feature Calibration Module (\texttt{FCM}) and Gradient Rebalancing Module (\texttt{GRM}), \myName{} effectively mitigates both representation and optimization imbalance.  
Extensive experiments on multiple MER backbones demonstrate that \myName{} consistently improves performance across standard, incomplete, and imbalanced-modality settings.  
The results highlight its robustness to heterogeneous missing patterns and its ability to maintain stable optimization even when modality availability varies significantly.  
Future work will explore broader multimodal tasks and large-scale pretrained backbones.

%% file: material/res-mosei-mean.tex
\resizebox{\linewidth}{!}{%
\begin{tabular}{lcccccccccccc}
\toprule
\multirow{2}{*}{\textbf{Model}} &
   \multicolumn{2}{c}{$(0.3,0.5,0.7)$} &
   \multicolumn{2}{c}{$(0.3,0.7,0.5)$} &
   \multicolumn{2}{c}{$(0.5,0.3,0.7)$} &
   \multicolumn{2}{c}{$(0.5,0.7,0.3)$} &
   \multicolumn{2}{c}{$(0.7,0.3,0.5)$} &
   \multicolumn{2}{c}{(0.7,0.5,0.3)} \\ 
\cmidrule(lr){2-13}
 & Acc. & w-F1 & Acc. & w-F1 & Acc. & w-F1 & Acc. & w-F1 & Acc. & w-F1 & Acc. & w-F1 \\ 
\midrule
\rowcolor{gray!10}\multicolumn{13}{c}{\textbf{\textit{Group 1 (Models addressing missing or imbalanced modalities)}}}\\
MMIN~\cite{zhao2021MMIN}    & 75.87 & 75.52 & 71.93 & 71.60 & 79.72 & 79.69 & 71.46 & 71.30 & 78.92 & 78.45 & 75.65 & 75.81 \\
SDR-GNN~\cite{fu2024sdr} & 80.13 & 80.14 & 77.38 & 77.38 & 83.35 & 83.03 & 76.69 & 76.75 & 83.54 & 83.44 & 81.67 & 81.21 \\
Mi-CGA~\cite{nguyen2025-micga}  & 81.04 & 80.97 & 77.38 & 77.11 & 82.55 & 82.37 & 76.42 & 76.45 & 82.86 & 82.64 & 80.65 & 80.33 \\
MoMKE~\cite{xu2024leveraging}   & 78.43 & 78.39 & 74.16 & 74.19 & 82.17 & 81.86 & 74.38 & 73.93 & 82.11 & 81.79 & 77.27 & 76.77 \\ 
GCNet~\cite{lian2023gcnet}  & 81.43 & 81.07 & 78.23 & 78.14 & 82.64 & 82.78 & 78.34 & 78.51 & 83.76 & 83.57 & 81.45 & 81.24 \\\hdashline[1pt/2pt]
Ada2I \cite{nguyen2024ada2i} & 82.42 & 82.28 & 78.07 & 77.53 & 83.82 & 83.57 & 77.68 & 76.97 & 83.38 & 83.49 & 81.87 & 81.61 \\
RedCore~\cite{sun2024redcore} & 74.88 & 72.89 & 70.51 & 64.25 & 75.05 & 72.82 & 75.01 & 72.79 & 74.37 & 72.71 & 74.94 & 73.59 \\
\hline
\rowcolor{blue!10}\textbf{GCNet\textsubscript{+\myName}} &
{81.54}\textsubscript{$\pm0.58$} & {81.40}\textsubscript{$\pm0.59$} & {78.51}\textsubscript{$\pm0.61$} & {78.33}\textsubscript{$\pm0.49$} &
  {83.27}\textsubscript{$\pm0.26$} & {83.24}\textsubscript{$\pm0.33$} & {79.50}\textsubscript{$\pm0.99$} & {79.22}\textsubscript{$\pm0.84$} &
  {83.63}\textsubscript{$\pm0.35$} & {83.57}\textsubscript{$\pm0.48$} & {82.09}\textsubscript{$\pm0.75$} & {81.84}\textsubscript{$\pm0.72$}
 \\
  \footnotesize\textit{$\Delta$ (vs origin GCNet)} &
  \footnotesize\textit{+0.11} & \footnotesize\textit{+0.33} &
  \footnotesize\textit{+0.28} & \footnotesize\textit{+0.19} &
  \footnotesize\textit{+0.63} & \footnotesize\textit{+0.46} &
  \footnotesize\textit{+1.16} & \footnotesize\textit{+0.71} &
  \footnotesize\textit{-0.13} & \footnotesize\textit{+0.00} &
  \footnotesize\textit{+0.64} & \footnotesize\textit{+0.60} \\
\footnotesize\textit{$\Delta_1$ (vs Group1 mean)} &
  \footnotesize\textit{+2.37} & \footnotesize\textit{+2.65} &
  \footnotesize\textit{+3.13} & \footnotesize\textit{+4.02} &
  \footnotesize\textit{+1.94} & \footnotesize\textit{+2.37} &
  \footnotesize\textit{+3.79} & \footnotesize\textit{+3.98} &
  \footnotesize\textit{+2.35} & \footnotesize\textit{+2.70} &
  \footnotesize\textit{+3.02} & \footnotesize\textit{+3.19} \\
\midrule
\rowcolor{gray!10}\multicolumn{13}{c}{\textbf{\textit{Group 2 (Typical MER Models)}}}\\
MMGCN~\cite{hu-etal-2021-mmgcn} & 80.82 & 80.20 & 78.32 & 77.35 & 84.76 & 84.60 & 78.12 & 78.17 & 84.04 & 84.04 & 81.54 & 81.62 \\
\rowcolor{blue!10}\textbf{MMGCN}\textbf{\textsubscript{+BALM}} &
  {82.03}\textsubscript{$\pm0.31$} & {81.74}\textsubscript{$\pm0.28$} & {78.73}\textsubscript{$\pm0.27$} & {78.33}\textsubscript{$\pm0.31$} &
  {84.54}\textsubscript{$\pm0.32$} & {84.34}\textsubscript{$\pm0.29$} & {79.83}\textsubscript{$\pm0.49$} & {79.44}\textsubscript{$\pm0.37$} &
  {84.84}\textsubscript{$\pm0.29$} & {84.70}\textsubscript{$\pm0.32$} & {82.50}\textsubscript{$\pm0.49$} & {82.36}\textsubscript{$\pm0.54$}
 \\
  \footnotesize\textit{$\Delta$ (vs origin MMGCN)} &
  \footnotesize\textit{+1.21} & \footnotesize\textit{+1.54} &
  \footnotesize\textit{+0.41} & \footnotesize\textit{+0.98} &
  \footnotesize\textit{-0.22} & \footnotesize\textit{-0.26} &
  \footnotesize\textit{+1.71} & \footnotesize\textit{+1.27} &
  \footnotesize\textit{+0.80} & \footnotesize\textit{+0.66} &
  \footnotesize\textit{+0.96} & \footnotesize\textit{+0.74} \\
\footnotesize\textit{$\Delta_1$ (vs Group1 mean)} &
  \footnotesize\textit{+2.86} & \footnotesize\textit{+2.99} &
  \footnotesize\textit{+3.35} & \footnotesize\textit{+4.02} &
  \footnotesize\textit{+3.21} & \footnotesize\textit{+3.47} &
  \footnotesize\textit{+4.12} & \footnotesize\textit{+4.20} &
  \footnotesize\textit{+3.56} & \footnotesize\textit{+3.83} &
  \footnotesize\textit{+3.43} & \footnotesize\textit{+3.71} \\\hline
MMDFN~\cite{hu2022mm} & 82.17 & 81.76 & 78.37 & 78.14 & 84.18 & 83.92 & 78.87 & 79.03 & 84.37 & 84.30 & 81.89 & 81.69 \\
\rowcolor{blue!10}\textbf{MMDFN}\textbf{\textsubscript{+BALM}} &
  {82.06}\textsubscript{$\pm0.43$} & {81.73}\textsubscript{$\pm0.45$} & {78.81}\textsubscript{$\pm0.58$} & {78.50}\textsubscript{$\pm0.63$} &
  {84.34}\textsubscript{$\pm0.44$} & {84.14}\textsubscript{$\pm0.39$} & {79.36}\textsubscript{$\pm0.39$} & {79.19}\textsubscript{$\pm0.39$} &
  {84.78}\textsubscript{$\pm0.21$} & {84.68}\textsubscript{$\pm0.26$} & {82.53}\textsubscript{$\pm0.35$} & {82.29}\textsubscript{$\pm0.28$}
 \\
  \footnotesize\textit{$\Delta$ (vs origin MMDFN)} &
  \footnotesize\textit{-0.11} & \footnotesize\textit{-0.03} &
  \footnotesize\textit{+0.44} & \footnotesize\textit{+0.36} &
  \footnotesize\textit{+0.16} & \footnotesize\textit{+0.22} &
  \footnotesize\textit{+0.49} & \footnotesize\textit{+0.16} &
  \footnotesize\textit{+0.41} & \footnotesize\textit{+0.38} &
  \footnotesize\textit{+0.64} & \footnotesize\textit{+0.60} \\
\footnotesize\textit{$\Delta_1$ (vs Group1 mean)} &
  \footnotesize\textit{+2.89} & \footnotesize\textit{+2.98} &
  \footnotesize\textit{+3.43} & \footnotesize\textit{+4.19} &
  \footnotesize\textit{+2.63} & \footnotesize\textit{+3.03} &
  \footnotesize\textit{+3.65} & \footnotesize\textit{+3.95} &
  \footnotesize\textit{+3.50} & \footnotesize\textit{+3.81} &
  \footnotesize\textit{+3.46} & \footnotesize\textit{+3.64} \\
\bottomrule
\end{tabular}%
}
\vspace{-7pt}
\parbox{\linewidth}{
  \raggedleft
\scriptsize \textit{Note.} $(r_A, r_L, r_V)$ denotes the missing rates for Audio, Language, and Visual modalities.\\
$\Delta$ and $\Delta_1$ denote performance differences relative to the original model and the mean of Group~1 models, respectively.
}

%% file: material/abl-FCM.tex
\vspace{-3pt}
\resizebox{0.7\columnwidth}{!}{%
\begin{tabular}{lcccc}
\toprule
\textbf{Setting} & \multicolumn{2}{c}{\textbf{MMGCN}} & \multicolumn{2}{c}{\textbf{+FCM}} \\ 
\cmidrule(lr){2-3} \cmidrule(lr){4-5}
 & Acc & w-F1 & Acc & w-F1 \\ 
\midrule
\textbf{IEMOCAP} 
& 48.09 & 44.81 
& \textbf{50.99}\textsubscript{($\uparrow$2.90)} 
& \textbf{47.62}\textsubscript{($\uparrow$2.81)} \\ 
\textbf{CMU-MOSEI} 
& 75.23 & 69.17 
& \textbf{75.97}\textsubscript{($\uparrow$0.74)} 
& \textbf{70.09}\textsubscript{($\uparrow$0.92)} \\
\bottomrule
\end{tabular}%
}

%% file: material/abl-stress-test.tex
\resizebox{0.83\linewidth}{!}{%
    \begin{tabular}{clcccc}
    \hline
    \multirow{2}{*}{$(r_A,r_L,r_V)$} & \multirow{2}{*}{\textbf{Model}} & \multicolumn{2}{c}{\textbf{IEMOCAP}} & \multicolumn{2}{c}{\textbf{CMU-MOSEI}} \\ 
    \cmidrule(lr){3-4} \cmidrule(lr){5-6}
     &  & Acc & w-F1 & Acc & w-F1 \\ 
    \midrule

    \multirow{3}{*}{$(0.1,0.1,0.9)$} 
    & GCNet & 63.71 & 63.43 & 84.62 & 84.64 \\ 
    & MMGCN & 68.02 & 67.69 & 85.91 & 85.52 \\
    \rowcolor{blue!10}
    & MMGCN\textbf{\textsubscript{+BALM}}            
    & \textbf{65.87}\textsubscript{\textit{$\uparrow$0.01}} 
    & \textbf{65.57}\textsubscript{\textit{$\uparrow$0.01}}  
    & \textbf{85.91}\textsubscript{\textit{$\uparrow$0.64}} 
    & \textbf{85.70}\textsubscript{\textit{$\uparrow$0.62}} \\ 
    \midrule

    \multirow{3}{*}{$(0.1,0.9,0.1)$} 
    & GCNet & 55.14 & 55.09 & 72.15 & 71.69 \\ 
    & MMGCN & 52.74 & 52.69 & 71.55 & 71.70 \\ 
    \rowcolor{blue!10}
    & MMGCN\textbf{\textsubscript{+BALM}}            
    & \textbf{58.72}\textsubscript{\textit{$\uparrow$4.78}} 
    & \textbf{58.11}\textsubscript{\textit{$\uparrow$4.22}}  
    & \textbf{72.62}\textsubscript{\textit{$\uparrow$0.77}} 
    & \textbf{72.15}\textsubscript{\textit{$\uparrow$0.46}} \\ 
    \midrule

    \multirow{3}{*}{$(0.9,0.1,0.1)$} 
    & GCNet & 59.40 & 58.83 & 86.24 & 86.10 \\ 
    & MMGCN & 64.82 & 64.93 & 86.43 & 86.42 \\ 
    \rowcolor{blue!10}
    & MMGCN\textbf{\textsubscript{+BALM}}            
    & \textbf{66.24}\textsubscript{\textit{$\uparrow$4.13}} 
    & \textbf{66.27}\textsubscript{\textit{$\uparrow$4.39}}  
    & \textbf{86.35}\textsubscript{\textit{$\uparrow$0.01}} 
    & \textbf{86.38}\textsubscript{\textit{$\uparrow$0.12}} \\ 
    \midrule

    \multirow{3}{*}{$(0.6,0.6,0.9)$} 
    & GCNet & 56.93 & 56.54 & 81.81 & 81.86 \\
    & MMGCN & 60.63 & 60.75 & 81.32 & 81.16 \\ 
    \rowcolor{blue!10}
    & MMGCN\textbf{\textsubscript{+BALM}}            
    & \textbf{61.74}\textsubscript{\textit{$\uparrow$2.96}} 
    & \textbf{61.92}\textsubscript{\textit{$\uparrow$3.28}}  
    & \textbf{81.65}\textsubscript{\textit{$\uparrow$0.09}} 
    & \textbf{81.62}\textsubscript{\textit{$\uparrow$0.11}} \\ 
    \midrule

    \multirow{3}{*}{$(0.6,0.9,0.6)$} 
    & GCNet & 54.84 & 54.85 & 76.00 & 76.02 \\ 
    & MMGCN & 55.64 & 55.61 & 76.36 & 75.97 \\ 
    \rowcolor{blue!10}
    & MMGCN\textbf{\textsubscript{+BALM}}            
    & \textbf{57.30}\textsubscript{\textit{$\uparrow$2.06}} 
    & \textbf{57.48}\textsubscript{\textit{$\uparrow$2.25}}  
    & \textbf{76.55}\textsubscript{\textit{$\uparrow$0.37}} 
    & 75.30\textsubscript{\textit{$\downarrow$0.70}} \\ 
    \midrule

    \multirow{3}{*}{$(0.9,0.6,0.6)$} 
    & GCNet & 57.98 & 58.16 & 81.45 & 81.06 \\ 
    & MMGCN & 59.95 & 59.89 & 81.59 & 81.50 \\ 
    \rowcolor{blue!10}
    & MMGCN\textbf{\textsubscript{+BALM}}            
    & \textbf{61.68}\textsubscript{\textit{$\uparrow$2.72}} 
    & \textbf{61.63}\textsubscript{\textit{$\uparrow$2.61}}  
    & \textbf{82.00}\textsubscript{\textit{$\uparrow$0.48}} 
    & \textbf{81.55}\textsubscript{\textit{$\uparrow$0.27}} \\ 

    \hline
    \end{tabular}%
    }

%% file: sec/X_suppl.tex
\clearpage
\setcounter{page}{1}
\setcounter{section}{0}
\maketitlesupplementary
\renewcommand{\thesection}{\Alph{section}}
\renewcommand{\thesubsection}{\thesection.\arabic{subsection}}


\section{Training procedure of \myName}
\label{supl:training}
The overall training procedure of \myName, including both the feature calibration and gradient rebalancing stages, is summarized in Algorithm~\ref{alg:overall}.

\begin{algorithm}[ht]
\caption{Training procedure of the proposed plug-in framework \myName}
\label{alg:overall}
\begin{algorithmic}[1]
\Require Multimodal dataset $\mathbb{D}$; model $(\phi^m,\mathcal{F},\mathbf{F}_{\textit{pred}})$; 
missing rates $\mathbf{r}=(r_a,r_v,r_l)$; total epochs $E$; hyperparameters $\tau,\rho,\alpha$.
\State Initialize modulation loss $\mathcal{L}_{\textit{mod}}\!\gets\!0$ and iteration $t\!\gets\!0$
\State Generate missing mask $e\!\sim\!\mathcal{M}(x;\mathbf{r})$
\State Generate incomplete dataset 
       $\tilde{\mathbb{D}}\!\gets\!\{(\tilde{x}_i,y_i)\}$
\For{$\textit{epoch}=1$ \textbf{to} $E$}
  \For{each $(\tilde{x}_i,y_i)\!\in\!\tilde{\mathbb{D}}$}
    \State \textbf{Calibrate} unimodal features: $\hat{x}^m_i\!\leftarrow\!\texttt{FCM}(\tilde{x}_i)$
    \State \textbf{Encode} modalities: $z^m_i\leftarrow \text{Eq.~\ref{eq:embedding}}$
    \State \textbf{Fuse} embeddings: $h_i\!\leftarrow\!\text{Eq.~\ref{eq:fused}}$  
    \State \textbf{Compute} multimodal prediction: 
           $\hat{y}_i\!\leftarrow\! \text{Eq.~\ref{eq:pred}}$
    \State \textbf{Compute} task loss: 
           $\mathcal{L}_{\textit{task}}\!\leftarrow\! \text{Eq.~\ref{eq:loss}}$
    \State \textbf{Aggregate} total loss: 
           $\mathcal{L}\!\leftarrow\!\mathcal{L}_{\textit{task}}+\tau\mathcal{L}_{\textit{mod}}$
    \State \textbf{Update} parameters: 
           $\theta\!\leftarrow\!\theta-\alpha\nabla_\theta\mathcal{L}$
    \State \textbf{Compute} unimodal predictions: 
           $\hat{y}^m_i\!\leftarrow\! \text{Eq.~\ref{eq:pred-m}}$
    \State \textbf{Compute} unimodal loss: 
           $\mathcal{L}^m_{\textit{task}}\!\leftarrow\! \text{Eq.~\ref{eq:uni-loss,upd}}$
    \State \textbf{Estimate} coefficients: $\mu^m\!\leftarrow\!\text{Eq.~\ref{eq:coef}}$
    \State \textbf{Modulate} encoder gradients: 
           $\theta^{\phi^m}\!\leftarrow\!\text{Eq.~\ref{eq:enc-mod}}$
    \State \textbf{Update} spatial modulation loss: 
           $\mathcal{L}_{\textit{mod}}\!\leftarrow\!\text{Eq.~\ref{eq:loss-mod}}$
    \State $t\!\leftarrow\!t+1$
  \EndFor
\EndFor
\end{algorithmic}
\end{algorithm}

\section{Theoretical Analysis}
\label{appendix:theo-analys}
\subsection{Gradient Imbalance under IMR}
\label{appl.Gradient-Im-Analysis}

This section provides a formal justification for Eq.~\ref{eq:gradient-imbalance} in the main paper,
showing how the imbalance in gradient magnitudes arises under the Imbalanced Missing Rate (IMR) condition.

\noindent\textbf{Assumption 1 (Independent Missingness).}
Each modality $m$ is independently missing according to a Bernoulli variable 
$e_i^m \in \{0,1\}$ with
$
P(e_i^m\!=\!1) = (1-r_m)$ and 
$
P(e_i^m=0)=r_m,
$,
where $r_m \in [0,1)$ denotes the missing probability of modality $m$.
The missingness indicators $\{e_i^m\}$ are assumed independent of the target label $Y$
(i.e., Missing Completely At Random-MCAR).

\noindent\textbf{Lemma 1 (Expected Gradient Scaling under IMR).}
Under above Assumption, for a differentiable per-sample loss 
$\ell_m = \ell(f_m(X_m), Y)$ with finite variance, 
the expected gradient of modality $m$ with respect to its parameters 
$\theta_m$ satisfies:
\setcounter{equation}{27}
\begin{equation}
\mathbb{E}\!\left[\nabla_{\theta_m}\mathcal{L}_{\mathrm{IMR}}\right]
= (1 - r_m)\,
\mathbb{E}\!\left[\nabla_{\theta_m}\ell(f_m(X_m), Y)\right].
\label{eq:supp-lemma}
\end{equation}

\noindent\textit{Proof.}
Let $\mathcal{L}_{\mathrm{IMR}}$ denote the expected loss under the IMR distribution:
\begin{equation}
\mathcal{L}_{\mathrm{IMR}}
= \mathbb{E}_{X,Y,e}\!
\left[\sum_{m=1}^{M} e^m \,\ell(f_m(X_m; \theta_m),Y)\right],
\label{eq:supp-loss}
\end{equation}
where $f_m(\cdot;\theta_m)$ represents the modality-specific encoder or prediction
branch parameterized by $\theta_m$, and $e^m$ indicates its availability.
Taking the gradient with respect to $\theta_m$ yields:
\begin{equation}
\nabla_{\theta_m}\mathcal{L}_{\mathrm{IMR}}
= \mathbb{E}_{X,Y,e}\!\left[e^m\,\nabla_{\theta_m}\ell(f_m(X_m;\theta_m),Y)\right].
\label{eq:supp-grad}
\end{equation}
By the linearity of expectation, we can separate the stochastic variable $e^m$:
\begin{align}
\mathbb{E}\!\left[\nabla_{\theta_m}\mathcal{L}_{\mathrm{IMR}}\right]
&= \mathbb{E}_{e}\!\left[e^m\right]\,
   \mathbb{E}_{X,Y}\!\left[\nabla_{\theta_m}\ell(f_m(X_m;\theta_m),Y)\right].
\label{eq:supp-step1}
\end{align}
Since $e^m \sim \mathrm{Bernoulli}(1 - r_m)$, we have $\mathbb{E}[e^m] = 1 - r_m$.
Substituting this into Eq.~\ref{eq:supp-step1} gives:
\begin{equation}
\mathbb{E}\!\left[\nabla_{\theta_m}\mathcal{L}_{\mathrm{IMR}}\right]
= (1 - r_m)\,
\mathbb{E}\!\left[\nabla_{\theta_m}\ell(f_m(X_m;\theta_m),Y)\right].
\label{eq:supp-result}
\end{equation}
Eq.~\ref{eq:supp-result} shows that, in expectation, the gradient propagated 
to the parameters $\theta_m$ of each modality-specific encoder is linearly 
scaled by its availability ratio $(1 - r_m)$.

\noindent\textbf{Corollary 1 (Gradient Imbalance Effect).}
Taking the $L_2$ norm of Eq.~\ref{eq:supp-result} yields
\begin{equation}
\mathbb{E}\!\left[\|\nabla_{\theta_m}\mathcal{L}_{\mathrm{IMR}}\|\right]
= (1 - r_m)\,
\mathbb{E}\!\left[\|\nabla_{\theta_m}\ell(f_m(X_m;\theta_m),Y)\|\right].
\label{eq:supp-corollary}
\end{equation}
Therefore, modalities with higher missing rates ($r_m$ large)
receive proportionally weaker gradient updates on their encoder parameters,
causing slower convergence and optimization dominance 
by low-missing-rate modalities.

The above analysis reveals that, under IMR, each modality receives a gradient
update scaled by its availability ratio $(1 - r_m)$.
This induces biased optimization dynamics where dominant modalities
converge faster while rare ones lag behind.
To counteract this imbalance, the Gradient Rebalancing Module (GRM)
introduced later (Eq.~\ref{eq:coef} and Eq.~\ref{eq:enc-mod}) adaptively rescales the gradient flow
through modulation coefficients $\mu_m$ that act as an inverse correction
to the implicit scaling factor $(1 - r_m)$.
In essence, GRM restores equilibrium among modalities by dynamically
adjusting both the magnitude and the direction of their gradients,
thereby stabilizing multimodal optimization under heterogeneous missing conditions.

\subsection{Gradient Rebalancing Rationale}

\noindent\textbf{Proposition 1.}
Building on \textbf{Lemma~1}, which shows that the expected gradient magnitude of each modality
is scaled by its availability ratio $(1 - r_m)$, we explain the rationale behind
the proposed Gradient Rebalancing Module (GRM).

\noindent\textbf{Rationale.}
GRM compensates for the imbalance in gradient magnitudes through adaptive modulation
of each modality’s update:
\begin{equation}
\begin{aligned}
\mu^m &= \rho\,
\frac{\sum_{m' \in \{a,v,l\},\, m'\neq m}\!\!\Delta^{m'}_{\mathrm{KL}}}
     {\sum_{m' \in \{a,v,l\}}\!\!\Delta^{m'}_{\mathrm{KL}}}, \\[4pt]
\theta^{\phi^m}_{(t+1)} 
&= \theta^{\phi^m}_{(t)} 
 - \alpha\,\mu^m
   \frac{\partial \mathcal{L}_{\mathrm{task}}}
        {\partial \theta^{\phi^m}_{(t)}}.
\end{aligned}
\label{eq:grm-update}
\end{equation}
Here, $\Delta^{m}_{\mathrm{KL}}$ quantifies the learning discrepancy of modality $m$
between its unimodal and fused distributions, and $\rho$ is a hyperparameter
controlling the modulation intensity.
A smaller $\mu^m$ indicates that modality $m$ is learning faster 
(\textit{i.e.}, larger $\Delta^{m}_{\mathrm{KL}}$) and thus its gradient update is attenuated,
whereas slower modalities are amplified.
This mechanism counteracts the imbalance identified in Lemma~1
and encourages all modality-specific gradients to approach an equilibrium:
\begin{equation}
\big\|\mu^m\nabla_{\theta^{\phi^m}}\mathcal{L}_{\mathrm{task}}\big\|
\approx
\big\|\mu^{m'}\nabla_{\theta^{\phi^{m'}}}\mathcal{L}_{\mathrm{task}}\big\|,
\quad \forall m,m'.
\end{equation}
Here, $\nabla_{\theta^{\phi^m}}\mathcal{L}_{\mathrm{task}}$ denotes the gradient of
the task loss with respect to the parameters of the $m$-th encoder, and $\|\cdot\|$
denotes the Euclidean norm measuring its magnitude.

Such modulation is consistent with previous analyses~\cite{peng2022balanced,guo2024classifier}
showing that gradient reweighting based on learning discrepancy stabilizes multimodal training.
GRM extends this idea to imbalanced missing-rate conditions, using KL-divergence as a
continuous signal of learning disparity rather than a fixed prior ratio $(1-r_m)$.

\section{Benchmark Datasets}
\label{supl:data}
\subsection{Dataset Description}
To evaluate the effectiveness of \myName, we conduct experiments on two widely used multimodal emotion and sentiment benchmarks: IEMOCAP~\cite{busso2008iemocap} and CMU-MOSEI~\cite{zadeh2018mosei}.  
Their key statistics are summarized in Table~\ref{tab:dataset-statistics}.

\setcounter{table}{4}
\begin{table}[h!]
    \centering
    \caption{Statistical overview of the IEMOCAP and CMU-MOSEI datasets.}
    \label{tab:dataset-statistics}
    \input{material/dataset_statistics}
\end{table}

\textbf{IEMOCAP} contains dyadic interactions between actors performing both scripted and improvised dialogues designed to elicit diverse emotions.  
The corpus comprises five sessions, each segmented into multiple utterances annotated with categorical emotion labels.  
Following the label processing in~\cite{ma2024sdt}, we adopt the common six-class setting.  
Since the original dataset only provides training and test splits, we further divide the training set into training and validation subsets using a ratio of $r$ (default 0.1).

\textbf{CMU-MOSEI} consists of 22,856 video-based utterances from over 1,000 YouTube speakers, each annotated with a sentiment score in the range $[-3, 3]$. Following~\cite{lian2023gcnet}, this dataset is trained as regression task, and evaluated as negative/positive classification task. Positive and negative classes are assigned for $<0$ and $> 0$ scores, respectively.
The official partitioning protocol is adopted to ensure consistency with previous studies. 

Similar to prior studies~\cite{hu-etal-2021-mmgcn,hu2022mm,nguyen2024ada2i}, we use Accuracy (Acc) and Weighted F1 Score (W-F1) as our main evaluation metrics.

\subsection{Multimodal Feature Extraction}
\label{supl:feature-ext}
For each utterance, multimodal features are extracted from acoustic, lexical, and visual modalities.  
The details of the extraction process for the two datasets are described as follows.

For the \textbf{IEMOCAP} dataset, we follow the feature extraction procedures outlined in~\cite{ma2024sdt} to obtain feature vectors for each modality.  
Specifically, we employ the RoBERTa-Large \cite{liu2019roberta} model to extract 1024-dimensional textual features.  
RoBERTa is fine-tuned for emotion recognition on conversation transcripts, and the embeddings of the [CLS] tokens from the last layer are used as textual representations.  
Acoustic features are extracted using openSMILE \cite{eyben2013recent} and then reduced to 1,582 dimensions via a fully connected layer, while visual features are obtained from a pre-trained DenseNet \cite{huang2017densely}, resulting in 342-dimensional representations for each utterance.

Similarly, we adopt the feature extraction methods described in~\cite{lian2023gcnet} for \textbf{CMU-MOSEI}.  
Pre-trained wav2vec\footnote{https://github.com/pytorch/fairseq/tree/main/examples/wav2vec} \cite{schneider2019wav2vec} is leveraged to extract 512-dimensional acoustic features for each utterance.  
For the textual modality, the pre-trained DeBERTa-Large model\footnote{https://huggingface.co/microsoft/deberta-large} \cite{he2020deberta} is exploited to encode word sequences into 1024-dimensional representations.  
Visual features are obtained through a two-step process: faces are first detected and aligned using the MTCNN~\cite{zhang2016joint} face detection algorithm, and the aligned frames are subsequently processed with MA-Net\footnote{https://github.com/zengqunzhao/MA-Net} \cite{zhao2021learning} to produce frame-level features.  
Finally, we aggregate these frame-level facial features into 1024-dimensional utterance-level representations using average pooling.

\section{Baseline Models} 
To evaluate the performance of \myName, we compare it with state-of-the-art methods for incomplete or imbalanced multimodal learning, as well as typical MER backbones.

\subsection{Incomplete or Imbalanced Multimodal Models}

The following baselines focus on addressing the challenges of missing modalities and uneven multimodal contribution.

\textbf{MMIN} \cite{zhao2021MMIN} learns robust joint representations by imagining the features of absent modalities from the available ones via cycle-consistent autoencoders, thereby handling uncertain missing conditions effectively.

\textbf{SDR-GNN} \cite{fu2024sdr} integrates spectral analysis into a hypergraph framework to impute missing data and explicitly retains high-frequency signals typically lost in conventional GNNs.

\textbf{Mi-CGA} \cite{nguyen2025-micga} utilizes a reconstruction module to approximate missing inputs and leverages cross-modal graph attention to capture comprehensive inter-modal dependencies.

\textbf{MoMKE} \cite{xu2024leveraging} adopts a dual-phase learning scheme in which a learnable router dynamically fuses outputs from pretrained unimodal encoders to derive a more comprehensive representation for incomplete data.

\textbf{GCNet} \cite{lian2023gcnet} captures speaker and temporal dependencies via graph neural networks to handle incomplete conversations and employs a dual-task framework to simultaneously predict target labels and restore missing features.

\textbf{Ada2I} \cite{nguyen2024ada2i} rectifies learning imbalances by dynamically re-weighting feature and modality contributions under the supervision of a learning discrepancy metric.

\textbf{RedCore} \cite{sun2024redcore} employs variational encoders to construct robust cross-modal representations and dynamically regulates auxiliary supervision based on reconstruction difficulty.

\textbf{MCE} \cite{zhao2025mce} optimizes training dynamics via game-theoretic evaluations and promotes semantic robustness through subset prediction to facilitate balanced feature capability despite imbalanced missing rates.

\subsection{MER Backbones}

To assess overall effectiveness, we further compare against mainstream backbones renowned for their multimodal context modeling and fusion mechanisms.

\textbf{MMGCN} \cite{hu-etal-2021-mmgcn} leverages a deep spectral graph network to fuse multimodal features. The framework defines utterances as interconnected nodes and captures long-range dependencies along with speaker context for emotion classification.

\textbf{MMDFN} \cite{hu2022mm} employs a gated graph architecture to selectively regulate cross-modal information flow. By dynamically filtering feature propagation across layers, it mitigates the accumulation of redundant data while strengthening inter-modal synergy.

\section{Implementation Details}
\label{supl:implement}





\begin{table}[t!]
    \centering
    \caption{Hyper-parameters Setting}
    \label{tab:exp-conf}
    \input{material/train-config}
\end{table}

We conduct experiments under varying missing configurations to evaluate the performance of different baselines on multimodal emotion recognition datasets. For each dataset $\mathbb{D}$ and missing setting $\mathbf{r}=(r_A,r_L,r_V)$, we generate modality-missing masks $e$ for the train/validation/test sets independently using the masking operator $\mathcal{M}(\cdot, \mathbf{r})$. Masking is applied prior to any processing or training, ensuring that complete data remain hidden from all models during both training and inference, while the missing masks remain fixed. The best model selected on the incomplete validation set is then evaluated on the incomplete test set.

For all baselines, we adopt their official implementations and model-specific hyperparameter settings (including learning rates) provided in their documentation. For \textbf{GCNet}, \textbf{SDR-GNN}, and \textbf{Mi-CGA}, specifically, we employ variants that omit reconstruction losses to ensure no baseline has access to any complete data.

\begin{table*}[t!]
    \centering
    \caption{Weighted-F1 sensitivity to hyperparameters under two contrastive missing-rate configurations of audio, language and visual.}
    \label{tab:spl.configAB}
    \input{material/spl.tune-configAB}
\end{table*}

For integrating \textbf{BALM} to \textbf{GCNet}, \textbf{MMGCN}, and \textbf{MMDFN}, \texttt{FCM} is added as a additional module to their architect while \texttt{GCM} is used as a separated module monitoring the training phase, the unimodal prediction heads is trained in parallel with the backbone using the same optimizer and learning rate, main architect and hyper-parameters of the backbone remain unchanged. Since \textbf{CMU-MOSEI} is treated as a regression task, we omit the softmax function ($\delta(\cdot)$ in Eq.~\ref{eq:pred}, Eq.~\ref{eq:pred-m}) and use sigmoid function to map predicted scalars to probabilities of negative/positive class for  Distribution-driven Modulation in \texttt{GRM}.

All experiments are implemented in PyTorch\footnote{https://pytorch.org/}, and tracked with Comet.ml\footnote{https://comet.ml}. Additional training configurations and hyper-parameters of \textbf{BALM} are summarized in Table~\ref{tab:exp-conf}.

\paragraph{Code Availability and Reproducibility.}
We release our full implementation and configurations at: \href{https://github.com/np4s/BALM_CVPR2026.git}{\url{https://github.com/np4s/BALM_CVPR2026.git}}.
The repository includes source code, configuration files, and brief guidelines with scripts to run experiments or adapt the framework to other datasets.
\paragraph{Masking Operator.}
Given a dataset of $N$ samples with $M$ modalities and a missing-ratio vector $\mathbf{r}=[r_1,\dots,r_M]$, our masking operator $\mathcal{M}(\cdot;\mathbf{r})$ generates missing masks based on the hypothetical ratios of missing patterns. For a missing pattern $\hat{e}=[\hat{e}^1,\dots,\hat{e}^M]$ among the $2^M$ possible patterns, its ratio is computed as:
\begin{equation}
    \hat{r}=\prod_{m=1}^M \hat{e}^m(1-r_m) \times (1-\hat{e}^m)r_m.
\end{equation}
Thus, $n=\lfloor N\hat{r}\rfloor$ samples are randomly assigned to the missing pattern $\hat{e}$, generating $n$ missing-mask vectors with $e_i=\hat{e}$. Since each sample must retain at least one modality, the $N\prod_{m=1}^M r_m$ masks corresponding to the pattern where all modalities are missing are redistributed uniformly among the $M$ patterns where exactly $M-1$ modalities are missing. Consequently, the error $\varepsilon_m$ of $\mathcal{M}(\cdot;\mathbf{r})$, defined as the discrepancy between the intended missing ratio $r_m$ and the realized missing ratio $\frac{1} {N}\sum_{i=1}^N (1-e_i^m)$ for modality $m$, satisfies $\varepsilon_m \leq \frac{1} {M}\prod_{m'=1}^M r_{m'} $.

\section{Additional Experiment Results}
\label{supl:results}

\subsection{Hyperparameter Sensitivity under Contrasting Missing Rates (Detailed Results)}
Table \ref{tab:spl.configAB} provides the complete numerical results of MMGCN\textsubscript{+\myName} on IEMOCAP for the two contrasting IMR configurations. 

Specifically, under \textit{Config--A} $(0.5,0.3,0.7)$, w-F1 remains stable across the $(\tau,\rho)$ grid, mostly lying within $63$--$64\%$, with limited sensitivity to either hyperparameter. 
The best score $66.30\%$ occurs at $(\tau{=}1.0,\rho{=}1.6)$, while all other settings fluctuate within a narrow range of about $2\%$.

In contrast, \textit{Config--B} $(0.5,0.7,0.3)$ shows noticeably larger spread, ranging from $58.39\%$ to $62.41\%$. 
Performance improves around small $\tau$ with moderate $\rho$ (best at $(\tau{=}0.1,\rho{=}1.4){=}62.41\%$), whereas higher $\rho$ consistently leads to degradation across all $\tau$. 
This pattern indicates stronger $\tau$--$\rho$ interaction when the lexical modality has the highest missing rate.

\subsection{Average performance on different modality combinations (Detailed Results)}

Table \ref{tab:spl.ablateModal-acc} and Table \ref{tab:spl.ablateModal-F1} report the complete results for Accuracy  (Acc) and weighted-F1 (w-F1), respectively, across all modality combinations-\textit{unimodal} (A, V, L), \textit{bimodal} (AV, AL, LV), and \textit{trimodal} (ALV)-under different missing-rate settings on IEMOCAP. For each combination, unspecified modalities are ablated from training and evaluating. We compare GCNet and MMGCN with their variants enhanced by \myName. Across all unimodal, bimodal, and trimodal configurations, the \myName-enhanced models consistently achieve higher Acc and W-F1, demonstrating improved robustness under varying missing-modality conditions.
The averaged performance across all configurations is presented in Fig.~\ref{fig:modal} of the main paper.

\begin{table*}[t!]
    \centering
    \caption{Performance across unimodal, bimodal, and trimodal configurations under varying missing rates (Accuracy)}
    \label{tab:spl.ablateModal-acc}
    \input{material/abl-modal-acc}

\end{table*}

\begin{table*}[ht!]
    \centering
    \caption{Performance across unimodal, bimodal, and trimodal configurations under varying missing rates (w-F1)}
    \label{tab:spl.ablateModal-F1}

\input{material/abl-modal-f1}

\end{table*}




Table~\ref{tab:spl.abl-FCM} reports the full MMGCN results across all modality combinations before and after integrating the \texttt{FCM} module. In this experiment, MMGCN and MMGCN\textsubscript{+\myName} are trained under the specified condition and tested on the complete test-set of modality combinations. The averaged performance is summarized in Table~\ref{tab:FCM-analysis} of the main paper. Overall, \texttt{FCM} provides consistent improvements across settings, highlighting the effectiveness of \myName in addressing the inconsistent representations arising from IMR.

\begin{table}[t!]
    \centering
    \caption{Performance of MMGCN on different modality combinations after training under IMR setting $(r_A,r_L,r_V)=(0.5,0.7,0.3)$, before and after plugged with \texttt{FCM} module.}
    \label{tab:spl.abl-FCM}
    \input{material/spl.abl-FCM-both}
\end{table}




\subsection{\myName under SMR settings (Numerical Results)}
Table~\ref{tab:smr-iemocap} and Table~\ref{tab:smr-mosei} report the detail results from Fig.~\ref{fig:smr-imr} in the main paper. By addressing challenges of missing modalities with \myName, MMGCN\textsubscript{+\myName} and MMDFN\textsubscript{+\myName} are able to achieve a more robust performance as the shared missing rate increases. Notably, for IEMOCAP, other methods addressing missing modalities can suffer up to over $3\%$ performance reduction, e.g. SDR-GNN and GCNet when SMR increase from $0.5$ to $0.6$, Mi-CGA when SMR increase from $0.6$ to $0.7$; while both MMGCN\textsubscript{+\myName} and MMDFN\textsubscript{+\myName} only see a $1\%-2\%$ drop across missing rates. Although more subtle, such trend in the performance's consistency can also be seen for CMU-MOSEI.
\begin{table*}[ht!]
    \centering
    \caption{Average performance of five runs under different SMR settings on IEMOCAP.}
    \label{tab:smr-iemocap}
    \input{material/smr-res-iemocap}
\end{table*}

\begin{table*}[ht!]
    \centering
    \caption{Average performance of five runs under different SMR settings on CMU-MOSEI.}
    \label{tab:smr-mosei}
    \input{material/smr-res-mosei}
\end{table*}

\subsection{Gradient Rebalancing Module Analysis}
To further investigate the two sub-modules of Gradient Rebalancing Module (GRM), we introduce two variants: \textbf{\myName-D} consisting \texttt{FCM} and \textit{distribution modulation} only, and its counterpart \textbf{\myName-S} consisting \texttt{FCM} and \textit{spatial modulation} only.

\begin{figure}[t!]
    \centering
    \includegraphics[width=\linewidth]{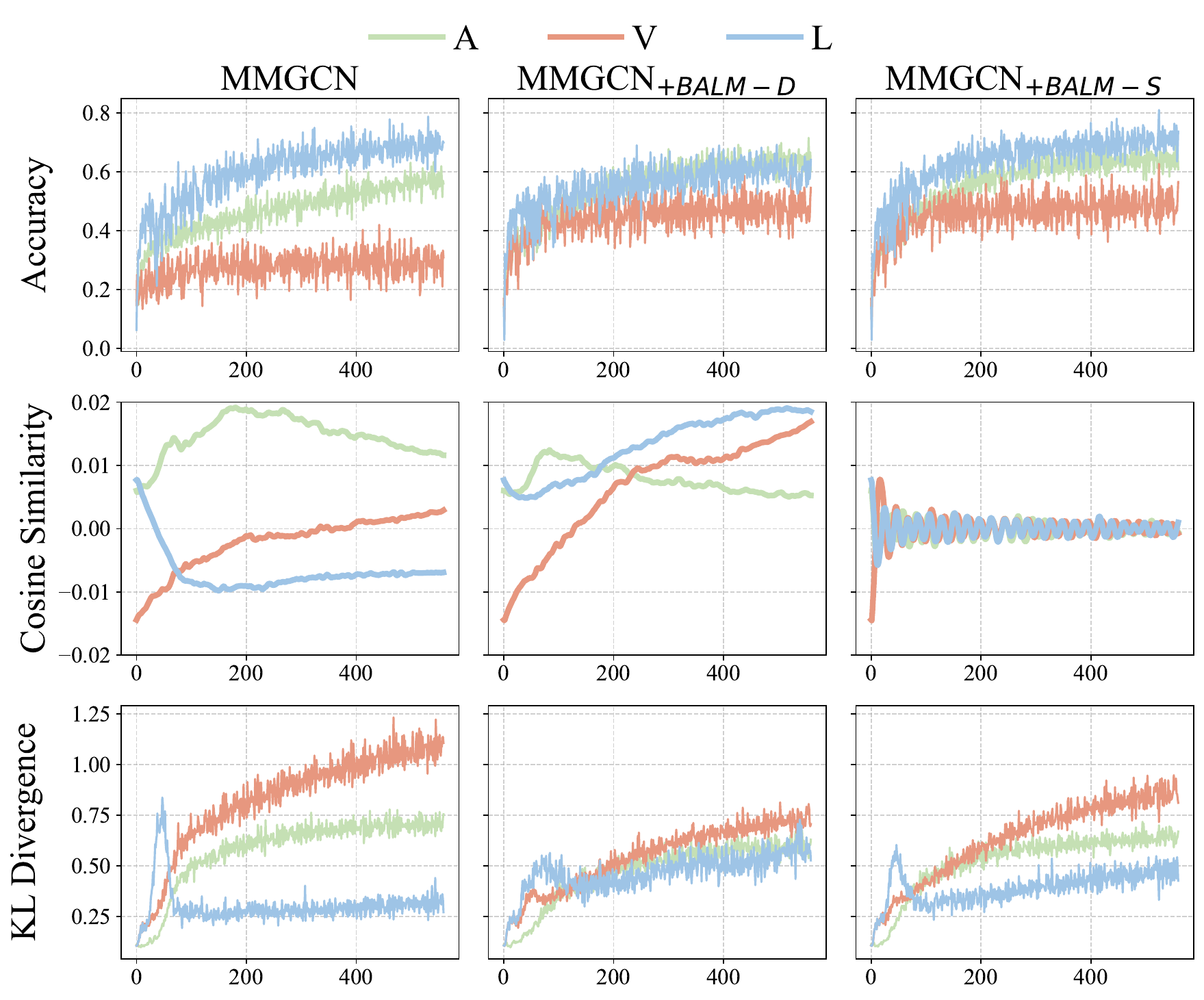}\vspace{-3pt}
    \caption{Modality discrepancies on IEMOCAP with different variants of \myName.}
    \label{fig:abl-subgrm}
\end{figure}

Fig.~\ref{fig:abl-subgrm} shows the learning progress of the modalities during training under IMR setting $(r_A,r_L,r_V)=(0.5,0.7,0.3)$, while Table~\ref{tab:abl-subgrm} displays quantitative performance of the two variants. The more robust and overall better performance of MMGCN\textsubscript{\textbf{+\myName}} (from Table~\ref{tab:res-iemocap-mean},\ref{tab:res-mosei-mean} in main paper) when compared to MMGCN\textsubscript{\textbf{+\myName-D}} and MMGCN\textsubscript{\textbf{+\myName-S}} further highlight the complementary nature of \textit{distribution-} and \textit{spatial-driven} Modulation. Fig.~\ref{fig:abl-subgrm} suggests that these two-sided modulations act as each other's regulator, keeping the backbone from being over-balanced toward either perspective (e.g., the cosine similarity of MMGCN\textsubscript{\textbf{+\myName-S}}), thus, giving a more general model.

\begin{table}[t!]
    \centering
    \caption{Performance of MMGCN on IEMOCAP under different IMR settings when integrated with two variants of \myName.}
    \label{tab:abl-subgrm}
    \input{material/tab-abl-subgrm}
\end{table}

\subsection{Distribution-driven Modulation Analysis}
We introduce the \textbf{\myName-M} variant, in which the KL divergence (i.e., $\mathbf{KL}(\cdot)$ in Eq.~\ref{eq:kl-div}) is replaced with an Mean Squared  Error (MSE) loss, while all other computations in \texttt{GRM} remain unchanged. Detailed results for GCNet\textsubscript{+\textbf{\myName-M}}, MMGCN\textsubscript{\textbf{+\myName-M}}, and MMDFN\textsubscript{\textbf{+\myName-M}}, along with their performance gaps relative to the corresponding baseline enhanced by \myName{}, are reported in Table~\ref{tab:abl-mse-iemo} and Table~\ref{tab:abl-mse-mosei}.

The improvements observed in MMDFN\textsubscript{\textbf{+\myName-M}} over MMDFN\textsubscript{\textbf{+\myName}} under most settings on IEMOCAP, up to $1.67\%$ under $(0.3,0.5,0.7)$, suggest that MSE can also serve as an effective measure of the distribution gap between multimodal and unimodal predictions for certain backbones. Nonetheless, KL divergence consistently yields better performance across different backbones and across both datasets, likely due to its softer characterization of distribution differences, verifies its advantage as a more suitable choice for distribution quantification as we aim to generalize \myName across diverse models.

\begin{table}[ht!]
    \centering
    \caption{Performance of \textbf{\myName-M} variant on IEMOCAP under different IMR settings.}
    \label{tab:abl-mse-iemo}
    \input{material/abl-mse-iemocap}
\end{table}

\begin{table}[ht!]
    \centering
    \caption{Performance of \textbf{\myName-M} variant on CMU-MOSEI under different IMR settings.}
    \label{tab:abl-mse-mosei}
    \input{material/abl-mse-mosei}
\end{table}

\subsection{Multimodal Representation}
Fig.~\ref{fig:tsne-input},~\ref{fig:tsne-emb},~\ref{fig:tsne-output} visualize the features comparison at different stages between MMGCN and MMGCN\textsubscript{+\myName} using t-SNE. As Fig.~\ref{fig:tsne-input} depicts, after being calibrated with \texttt{FCM}, the outliers of audio and lexical modalities become more visible, which is a better reflection of the dataset's properties in this specific scenario where these two modalities suffer loss at high rates, while visual features are also preprocessed into more defined clusters. Consequently, the embeddings in Fig.\ref{fig:tsne-emb} also show a better learning of modal-specific encoders, resulting in multimodal fused feature with more distinct border between \textit{hap-exc} (happy - excited) and \textit{ang-fru} (angry - frustrated)  clusters in Fig.~\ref{fig:tsne-output}.

\setcounter{figure}{7}
\begin{figure}[t!]
    \centering
    \begin{subfigure}{0.49\linewidth}
        \centering
        \includegraphics[width=\linewidth]{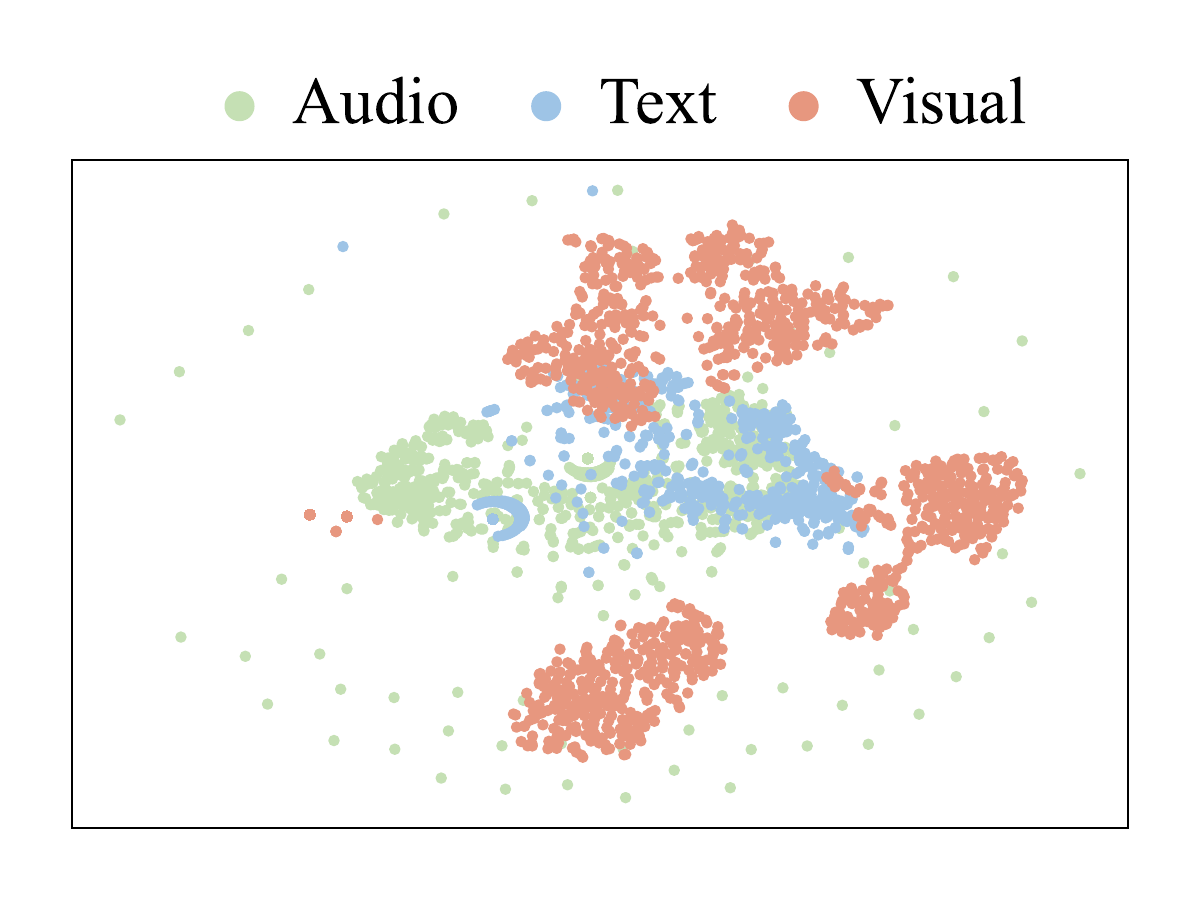}
        \caption{Original}
    \end{subfigure}
    \hfill
    \begin{subfigure}{0.49\linewidth}
        \centering
        \includegraphics[width=\linewidth]{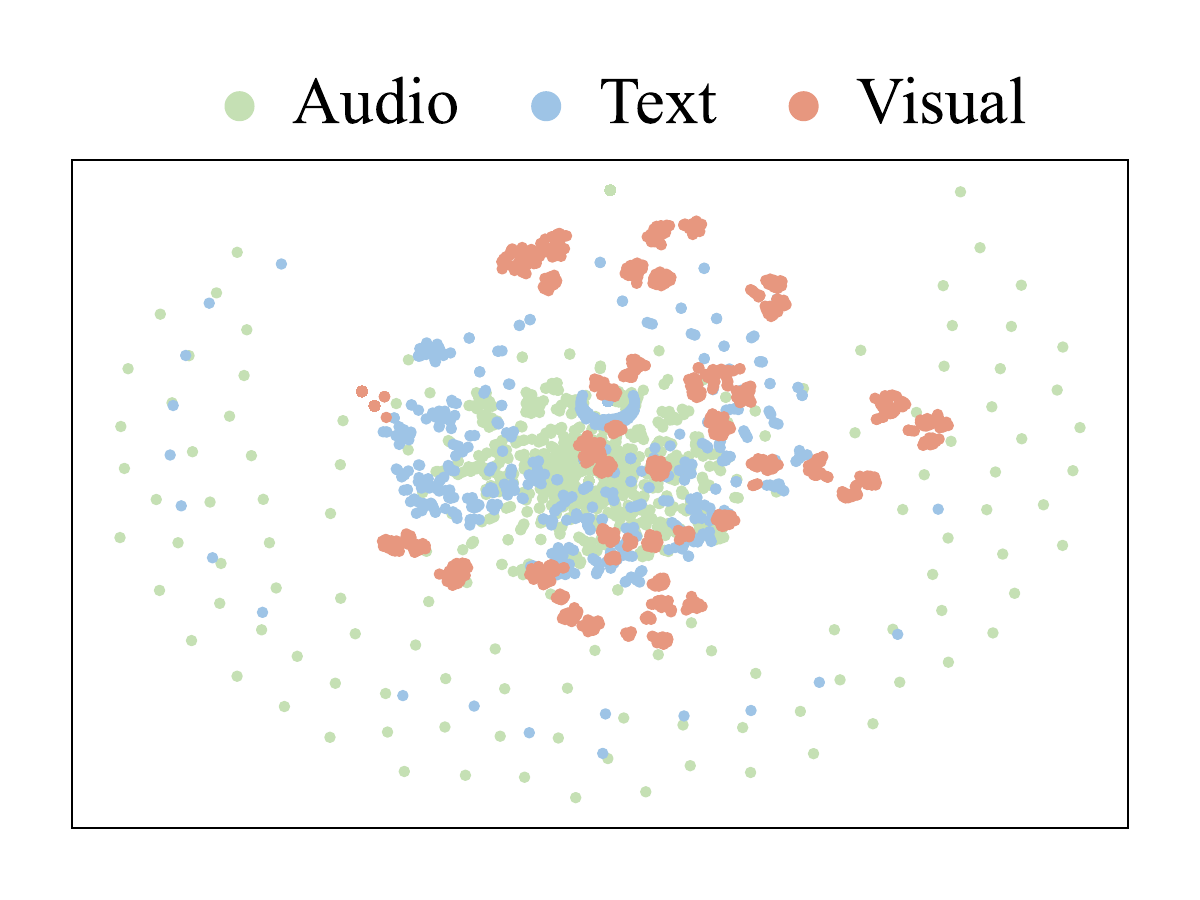}
        \caption{Calibrated}
    \end{subfigure}
    \caption{Original input of MMGCN (i.e., $\tilde{x}$) and calibrated input of MMGCN\textsubscript{+\myName} (i.e., $\hat{x}$), under IMR setting $(r_A,r_L,r_V)=(0.5,0.7,0.3)$ on IEMOCAP.}
    \label{fig:tsne-input}
\end{figure}

\begin{figure}[t!]
    \centering
    \begin{subfigure}{0.49\linewidth}
        \centering
        \includegraphics[width=\linewidth]{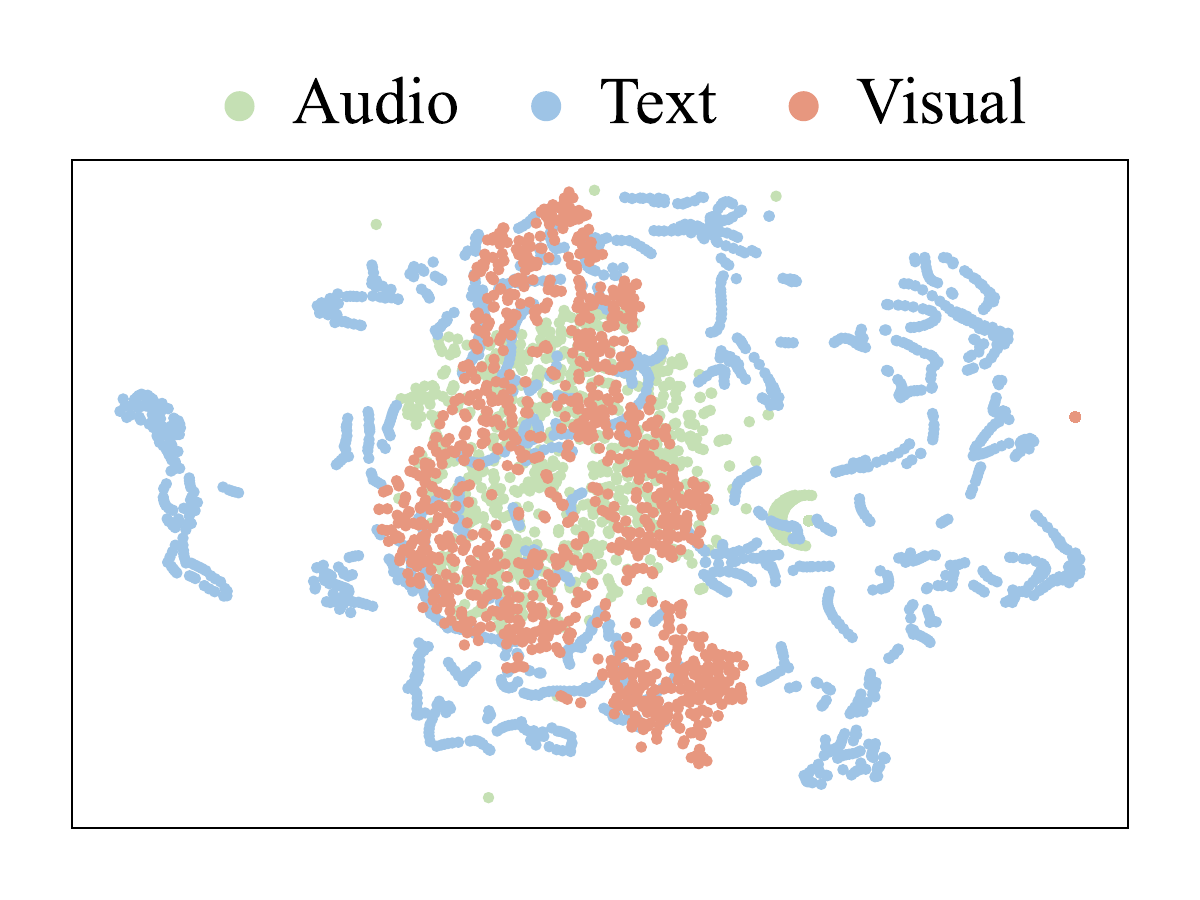}
        \caption{Embeddings of MMGCN}
    \end{subfigure}
    \hfill
    \begin{subfigure}{0.49\linewidth}
        \centering
        \includegraphics[width=\linewidth]{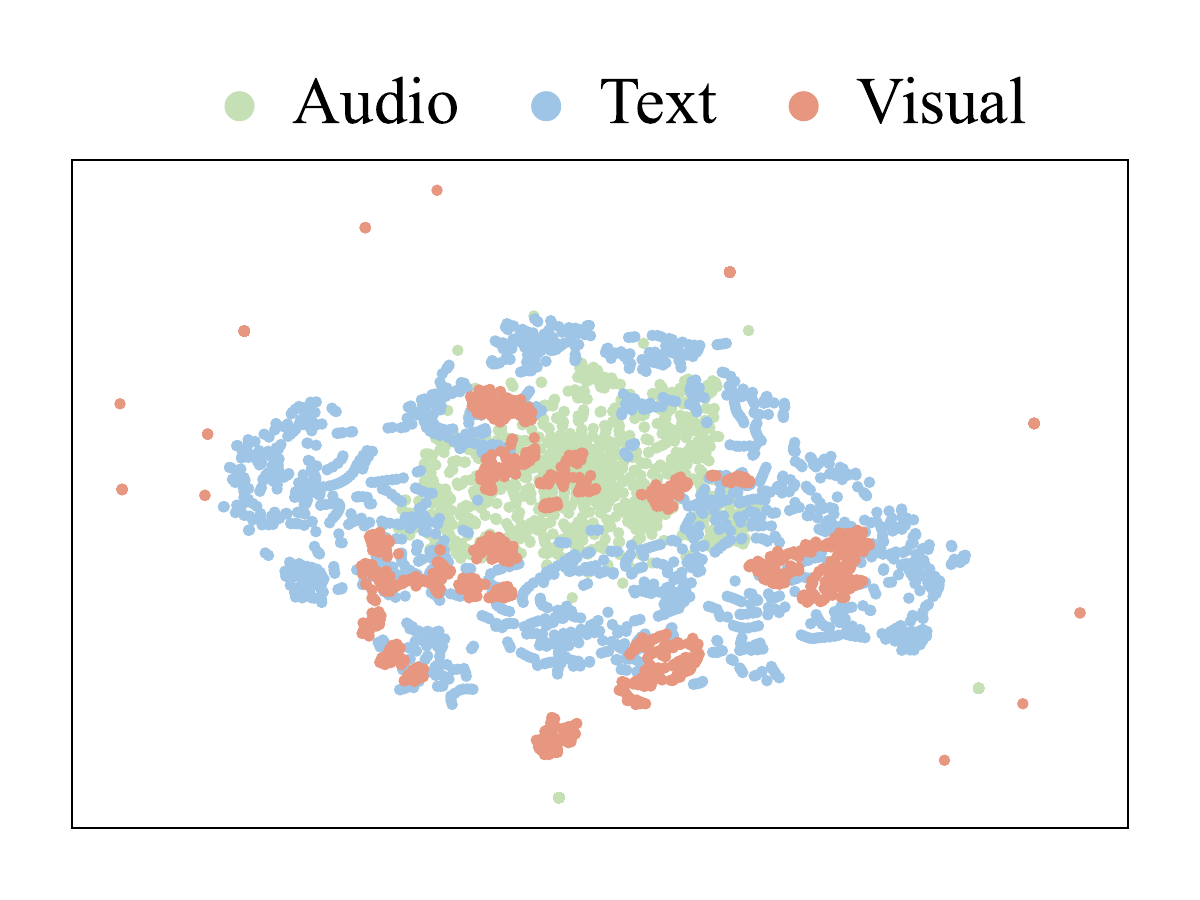}
        \caption{Embeddings of MMGCN\textsubscript{+\myName}}
    \end{subfigure}
    \caption{The embeddings (i.e., $z$ in Eq.~\ref{eq:embedding}) corresponding to the inputs from Fig.~\ref{fig:tsne-input}.}
    \label{fig:tsne-emb}
\end{figure}

\begin{figure}[t!]
    \centering
    \begin{subfigure}{0.49\linewidth}
        \centering
        \includegraphics[width=\linewidth]{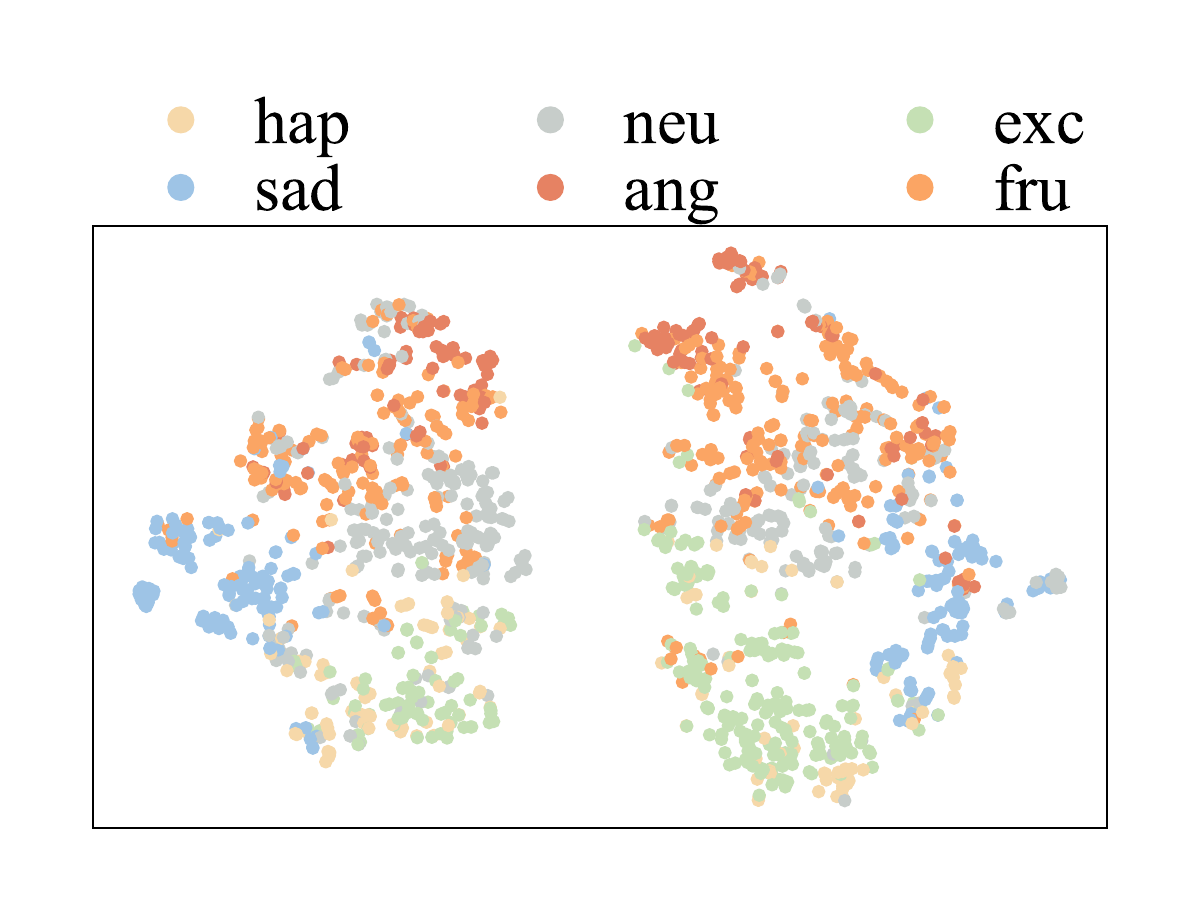}
        \caption{MMGCN}
    \end{subfigure}
    \hfill
    \begin{subfigure}{0.49\linewidth}
        \centering
        \includegraphics[width=\linewidth]{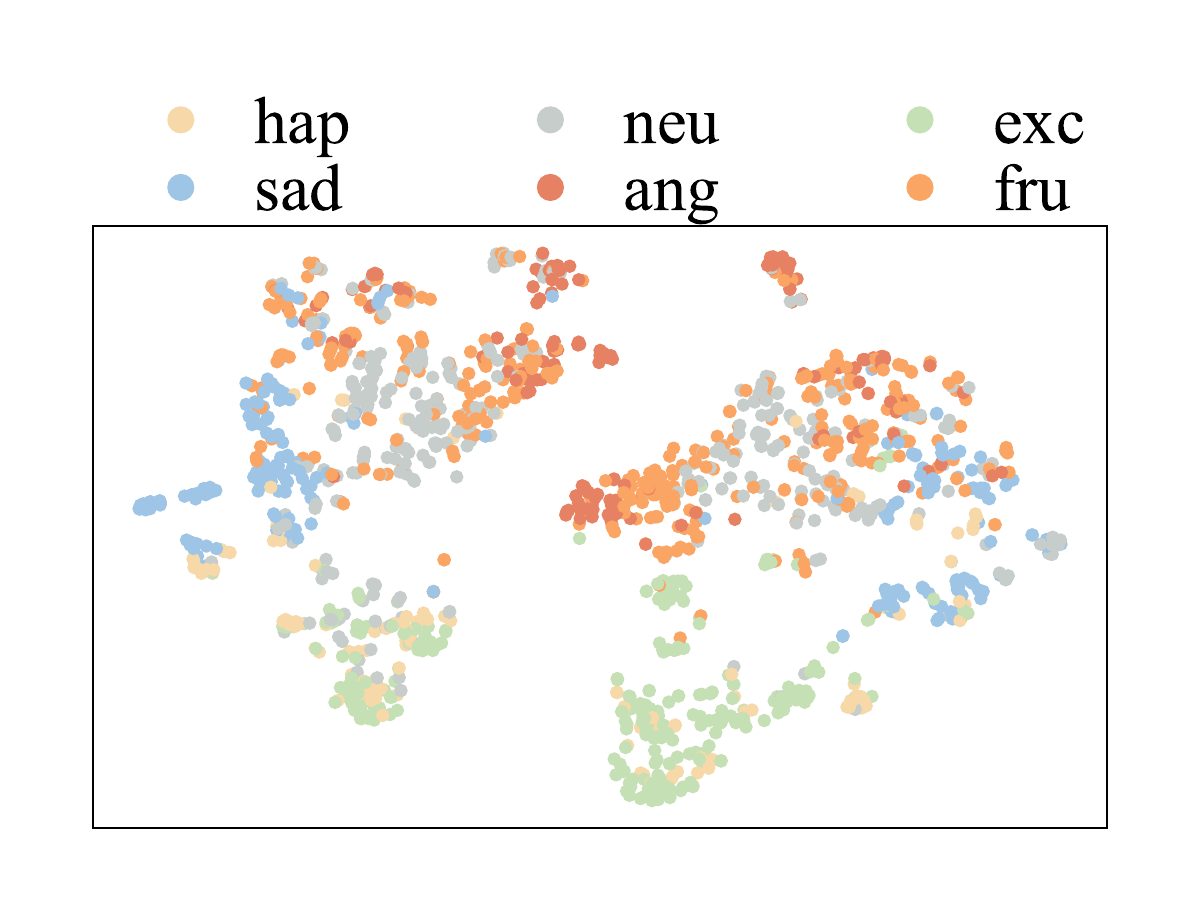}
        \caption{MMGCN\textsubscript{+\myName}}
    \end{subfigure}
    \caption{The multimodal fusion (i.e., $h$ in Eq.~\ref{eq:fused}) corresponding to the inputs from Fig.~\ref{fig:tsne-input}, colored by ground truth.}
    \label{fig:tsne-output}
\end{figure}

\subsection{Shared- and Imbalanced Missing Rates Relation}
Given a dataset of $N$ samples with $M$ modalities, and the binary observation indicator $e^m_i$ of modality $m$ for sample $i$ with $e^m_i=0$ denotes missing. Following~\cite{fu2024sdr,lian2023gcnet}, the shared missing rate is defined as:
\begin{equation}
    r_\text{shared}=\frac{\sum_{m=1}^M\sum_{i=1}^N (1-e^m_i)}{MN},
\end{equation}
while the modality-specific missing rate under IMR is given by:
\begin{equation}
    r_m=\frac{\sum_{i=1}^N (1-e^m_i)}{N}.
\end{equation}
Thus, we can assume the IMR settings equivalent to a given SMR are those that satisfy
\begin{equation}
    \sum_{m=1}^M r_m = M r_\text{shared}.
\end{equation}

Accordingly, in Fig.~\ref{fig:smr-vs-imr} of the main paper, we compare the performance of models addressing missing modalities under $SMR = 0.5$ with IMR configurations satisfying $r_A + r_L + r_V = 1.5$. The detailed results are provided in Table~\ref{tab:res-iemocap-mean} for the sampled IMR settings and in Table~\ref{tab:smr-iemocap} for $SMR = 0.5$. These experiments show that a shared missing rate alone cannot fully capture a model’s behavior, particularly under highly imbalanced missing-rate conditions, underscoring the necessity of investigating IMR scenarios.

\subsection{Modality Discrepancies}
\begin{figure}[ht!]
    \centering
    \includegraphics[width=\linewidth]{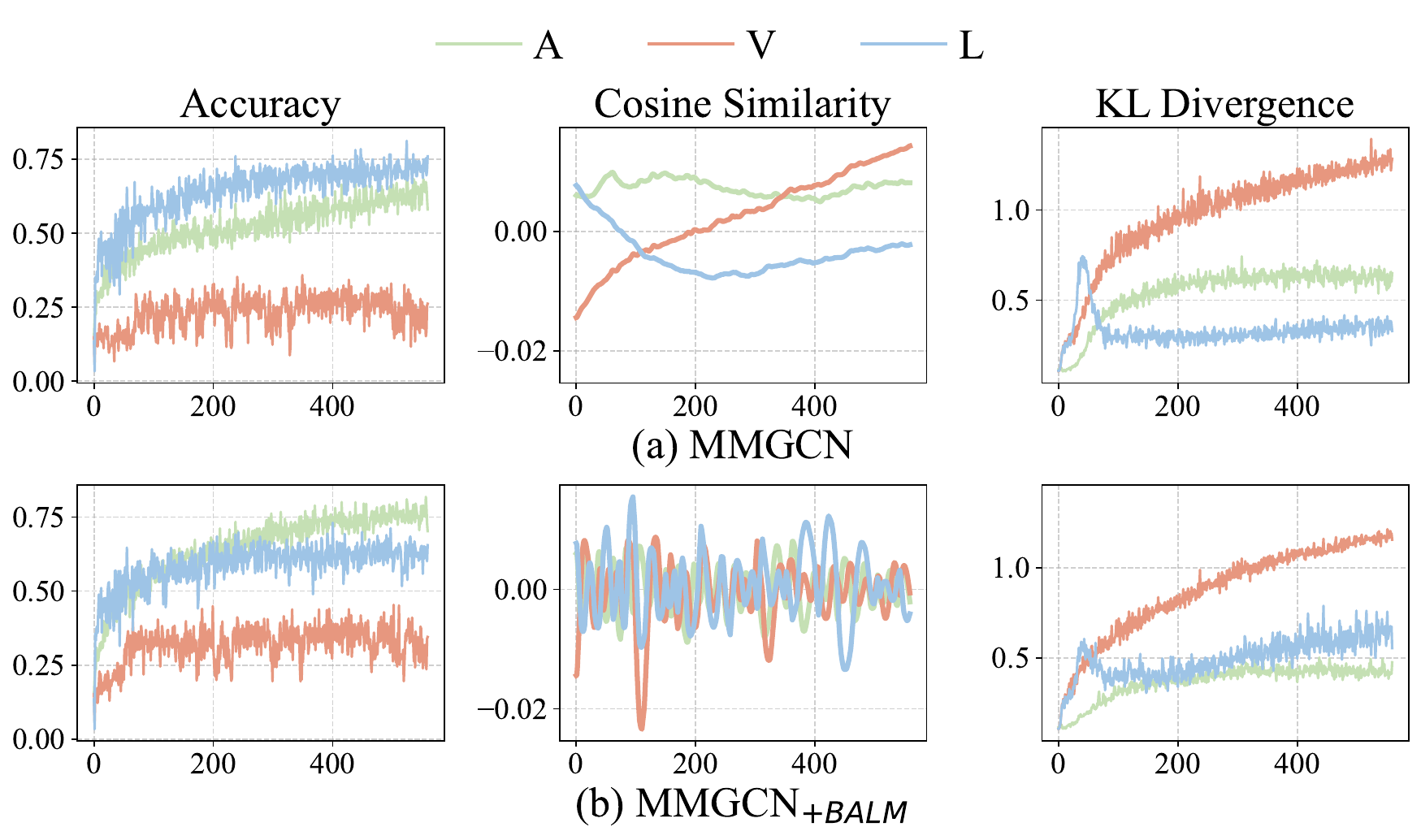}\vspace{-3pt}
    \caption{$(r_A,r_L,r_V)=(0.3, 0.5, 0.7)$}
    \label{fig:modal-dis-begin}
\end{figure}

\begin{figure}[h!]
    \centering
    \includegraphics[width=\linewidth]{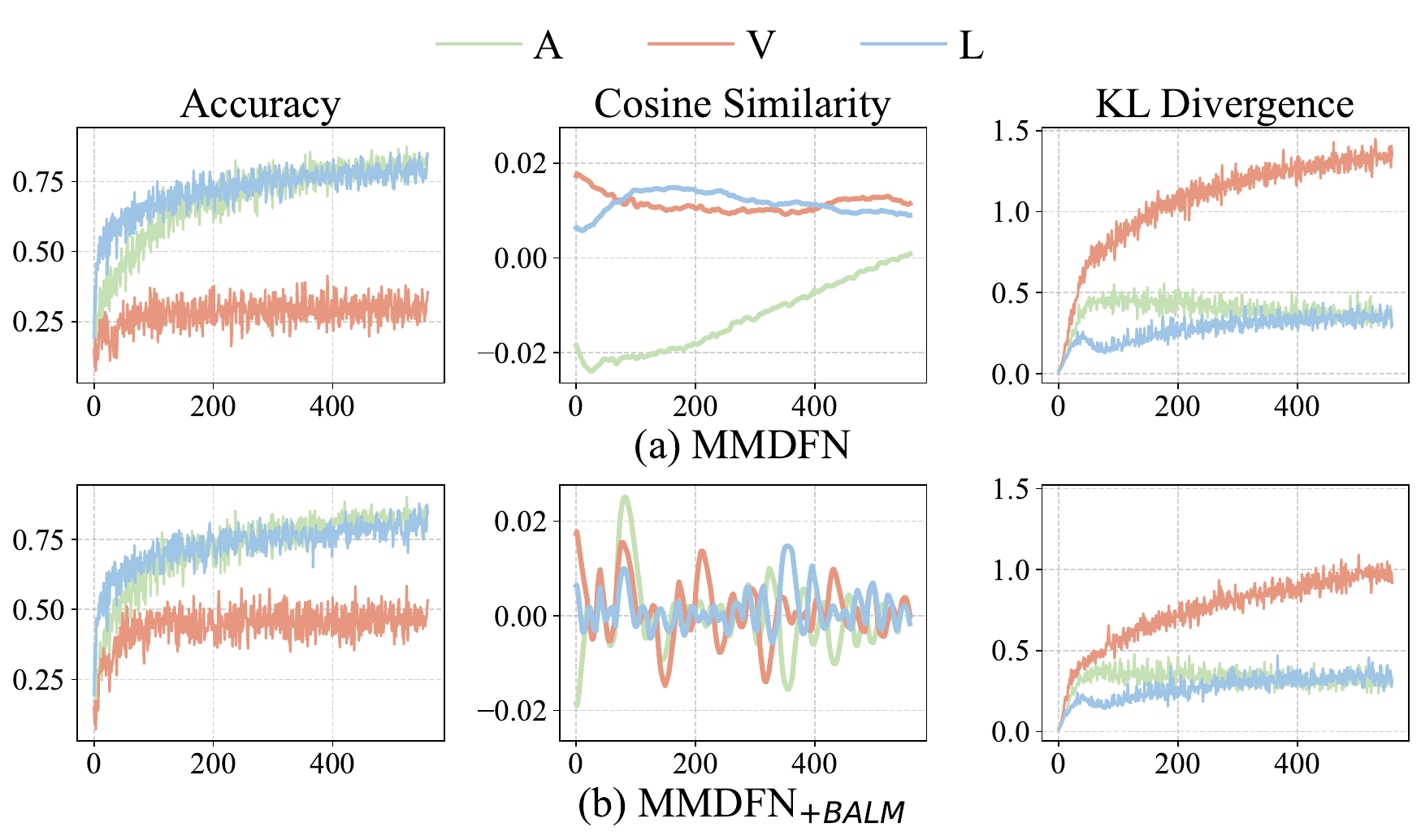}\vspace{-3pt}
    \caption{$(r_A,r_L,r_V)=(0.7, 0.5, 0.3)$}
    \label{fig:modal-dis-end}
\end{figure}

In this section, we provide extended visual comparisons of modality discrepancies for MMGCN vs. MMGCN\textsubscript{+\myName} and MMDFN vs. MMDFN\textsubscript{+\myName} on IEMOCAP under different IMR configurations (from Fig.\ref{fig:modal-dis-begin} to Fig. \ref{fig:modal-dis-end}).

\begin{figure}[h!]
    \centering
    \includegraphics[width=\linewidth]{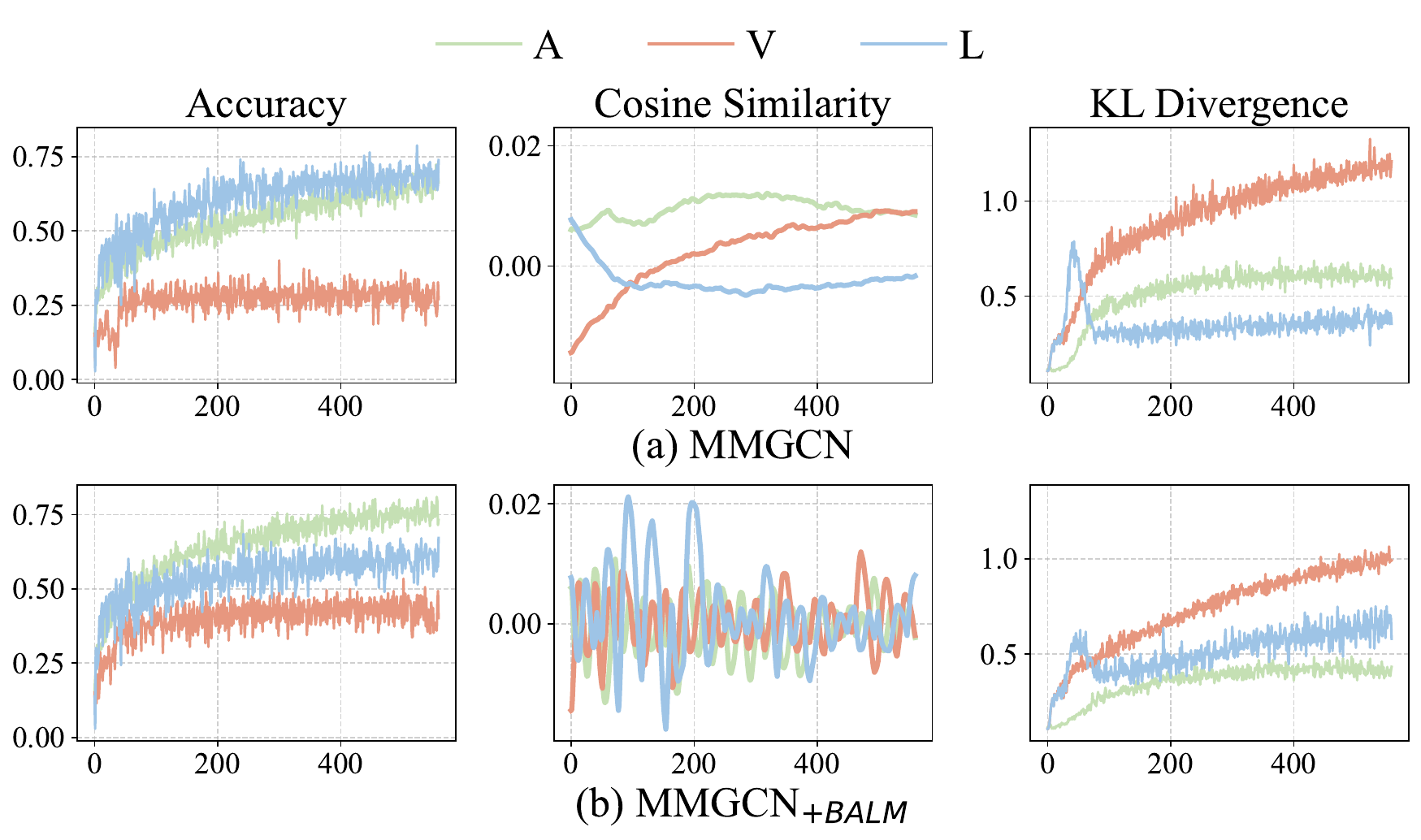}\vspace{-3pt}
    \caption{$(r_A,r_L,r_V)=(0.3, 0.7, 0.5)$}
\end{figure}
\begin{figure}[h!]
    \centering
    \includegraphics[width=\linewidth]{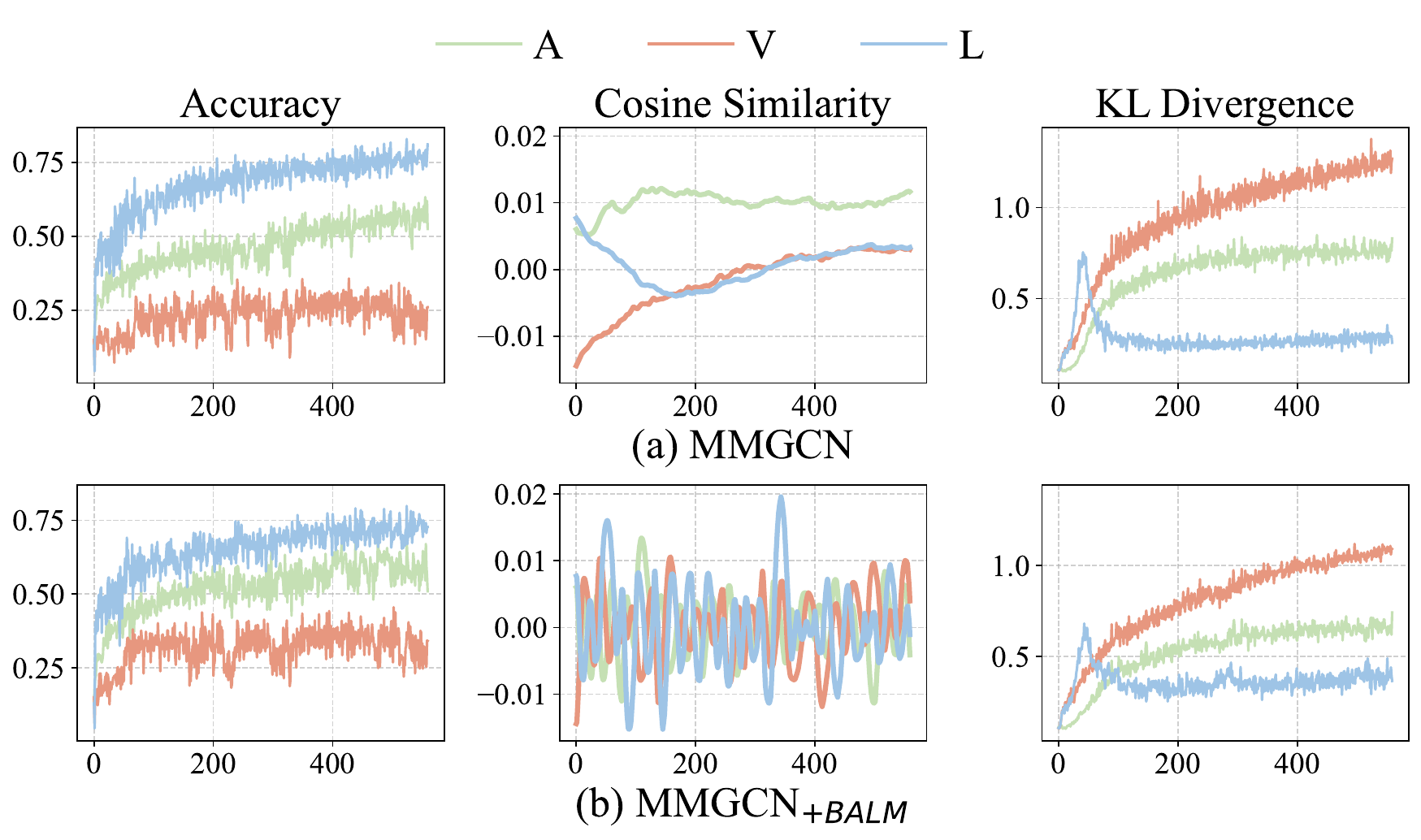}\vspace{-3pt}
    \caption{$(r_A,r_L,r_V)=(0.5, 0.3, 0.7)$}
\end{figure}
\begin{figure}[h!]
    \centering
    \includegraphics[width=\linewidth]{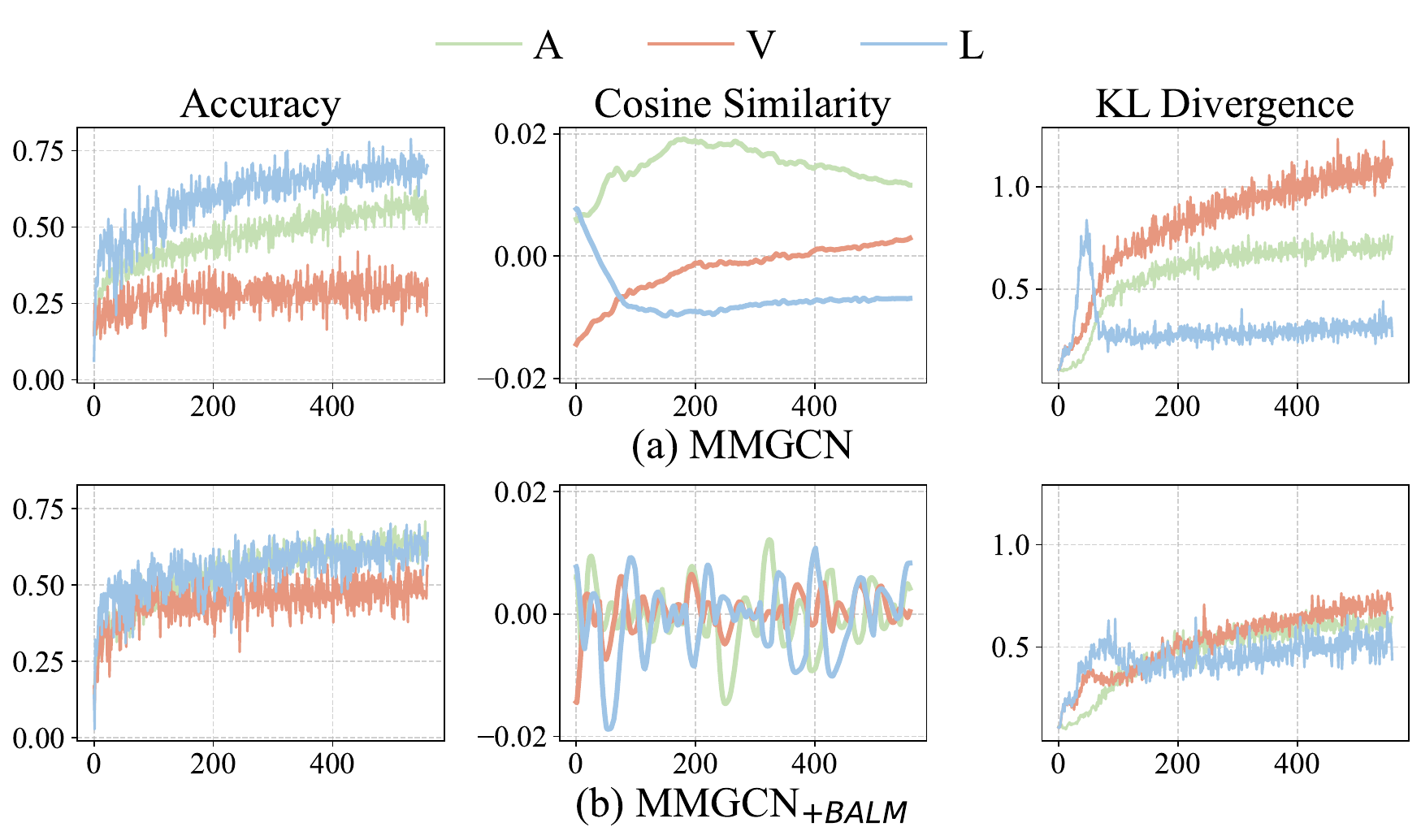}\vspace{-3pt}
    \caption{$(r_A,r_L,r_V)=(0.5, 0.7, 0.3)$}
\end{figure}
\begin{figure}[h!]
    \centering
    \includegraphics[width=\linewidth]{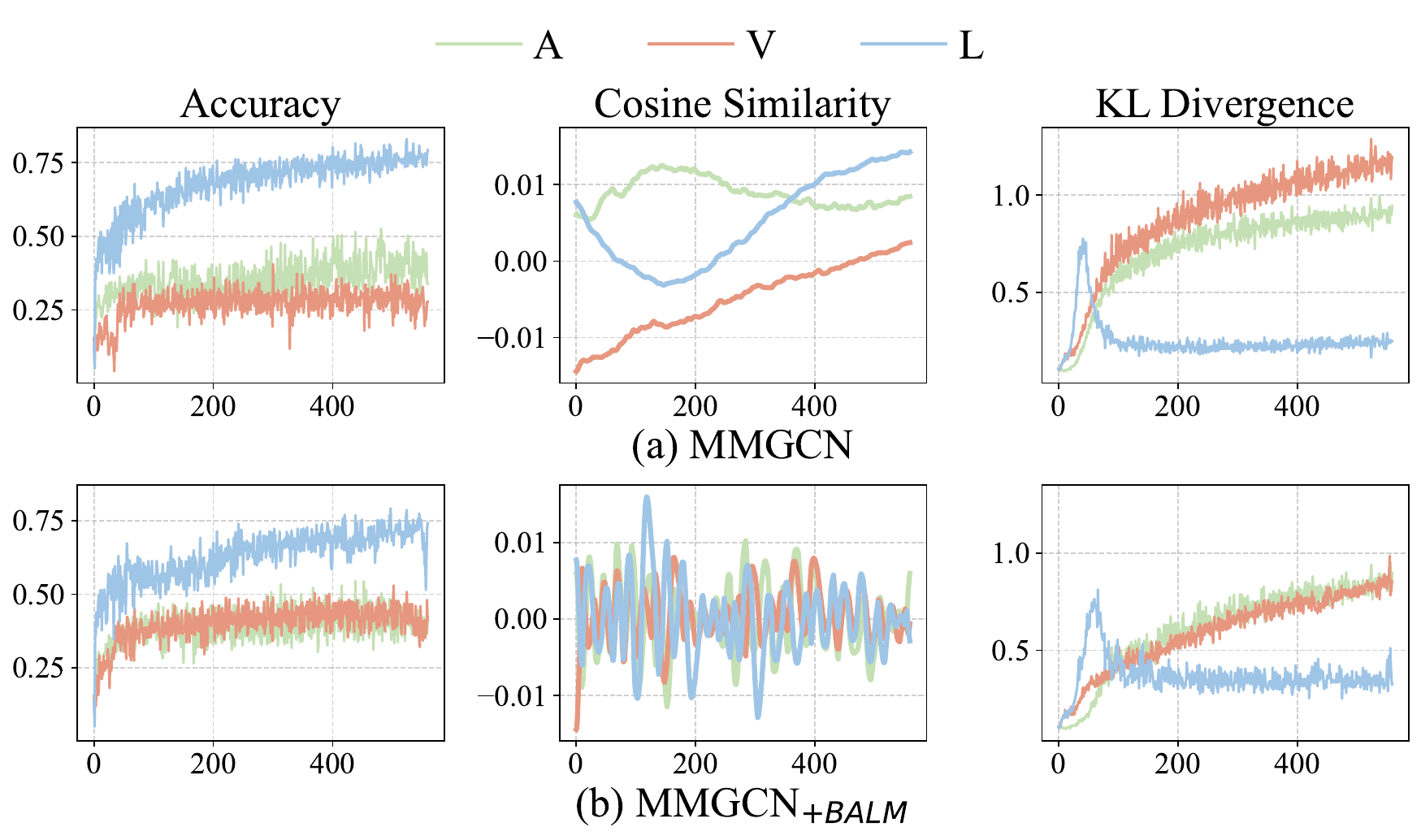}\vspace{-3pt}
    \caption{$(r_A,r_L,r_V)=(0.7, 0.3, 0.5)$}
\end{figure}
\begin{figure}[h!]
    \centering
    \includegraphics[width=\linewidth]{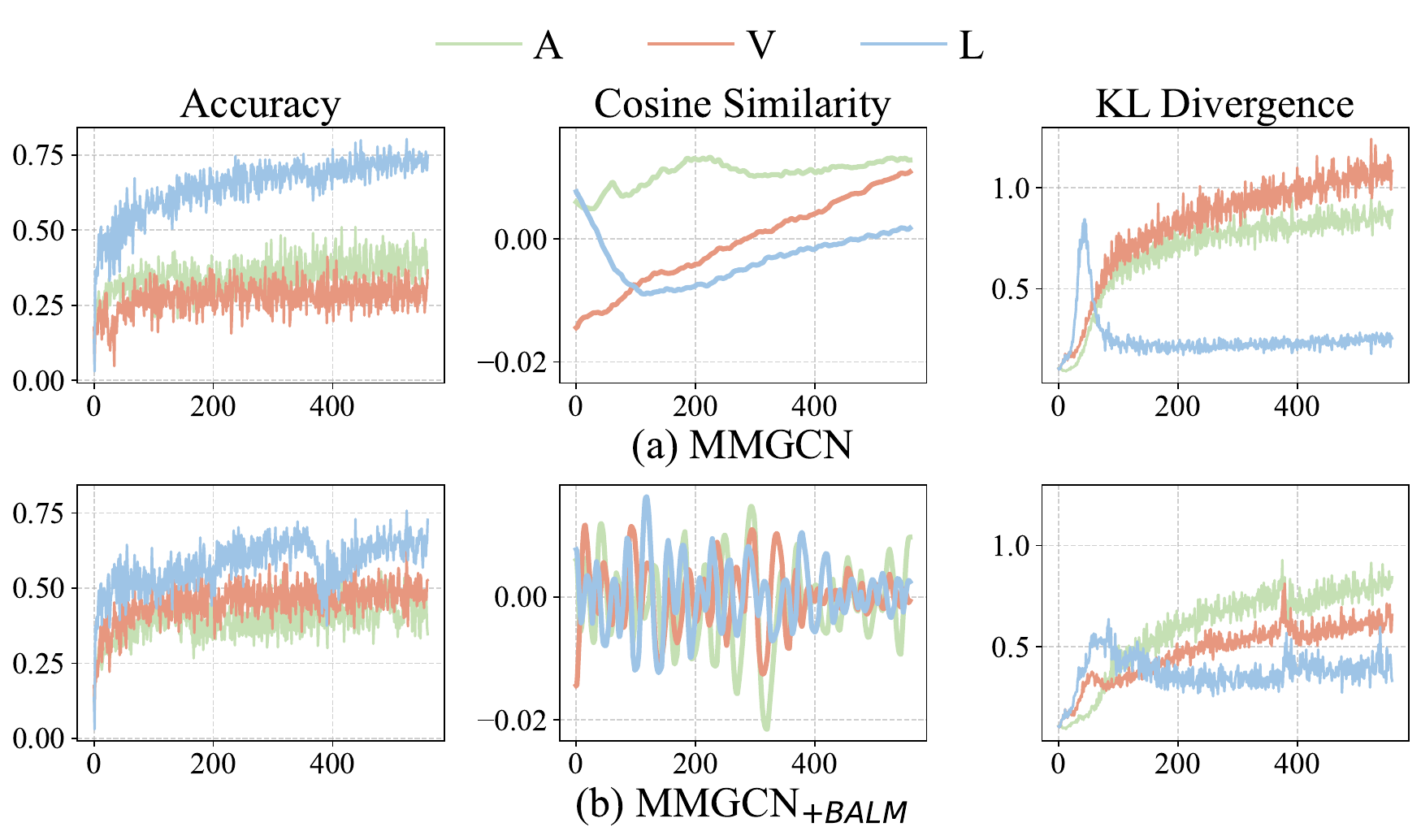}\vspace{-3pt}
    \caption{$(r_A,r_L,r_V)=(0.7, 0.5, 0.3)$}
\end{figure}
\begin{figure}[h!]
    \centering
    \includegraphics[width=\linewidth]{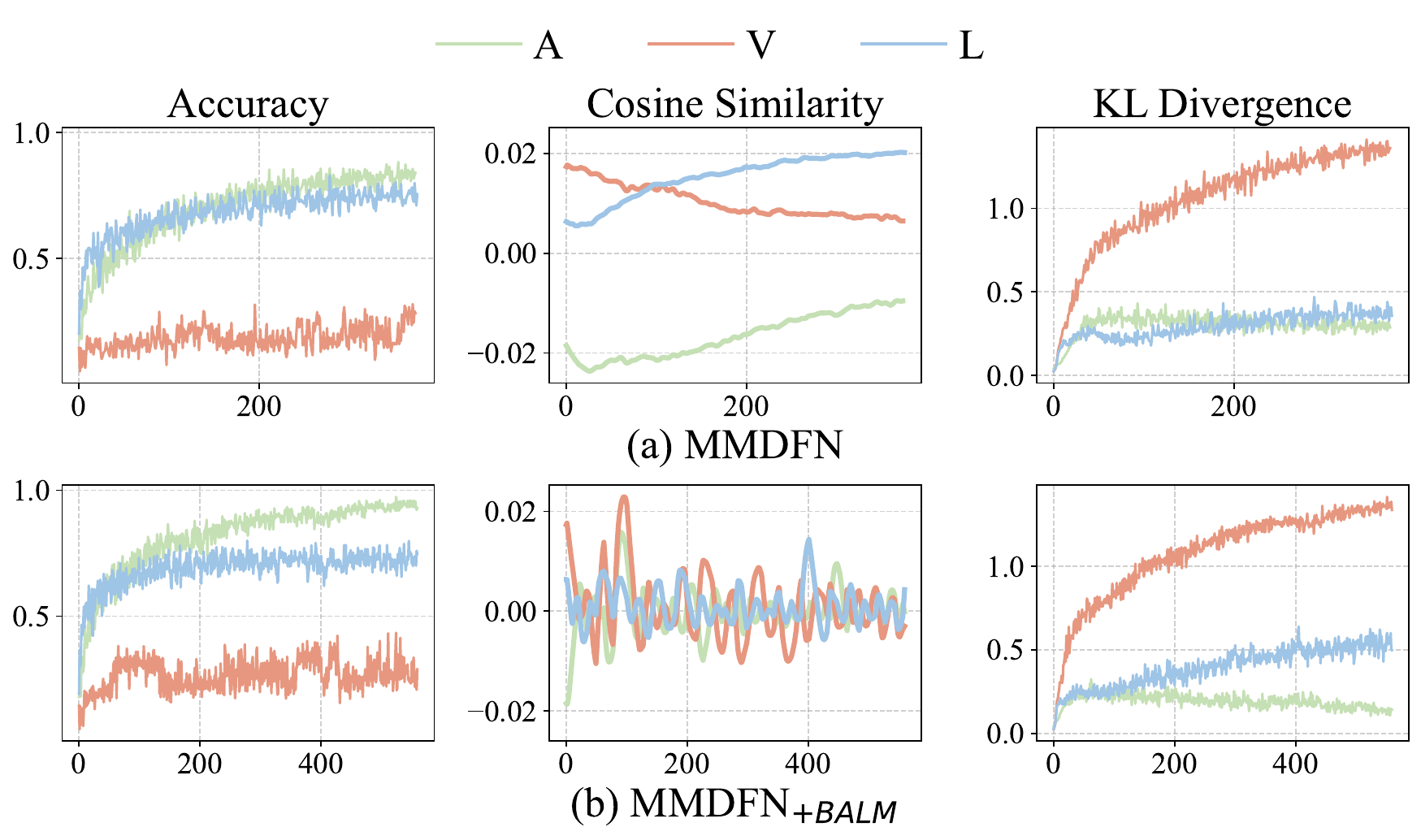}\vspace{-3pt}
    \caption{$(r_A,r_L,r_V)=(0.3, 0.5, 0.7)$}
\end{figure}
\begin{figure}[h!]
    \centering
    \includegraphics[width=\linewidth]{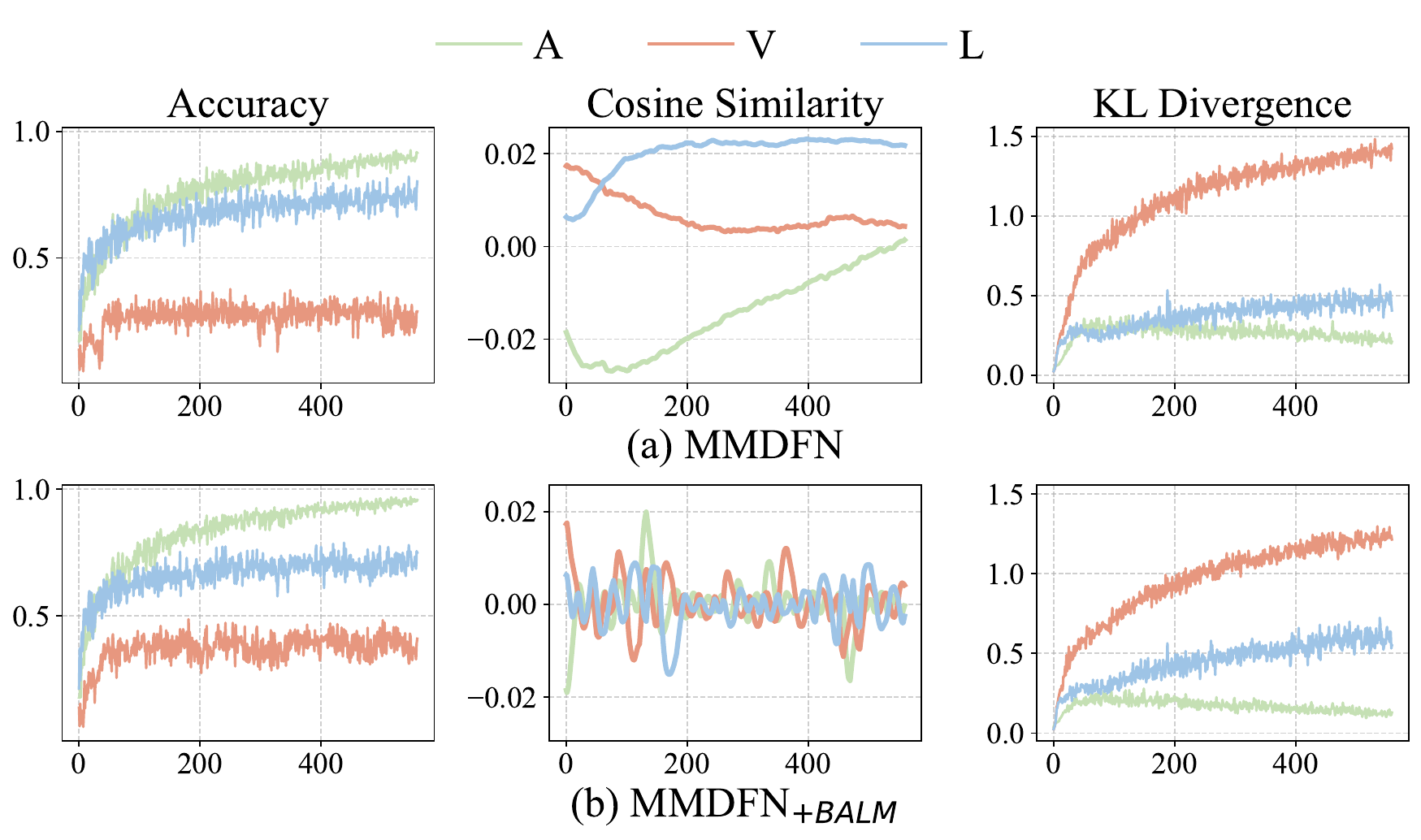}\vspace{-3pt}
    \caption{$(r_A,r_L,r_V)=(0.3, 0.7, 0.5)$}
\end{figure}
\begin{figure}[h!]
    \centering
    \includegraphics[width=\linewidth]{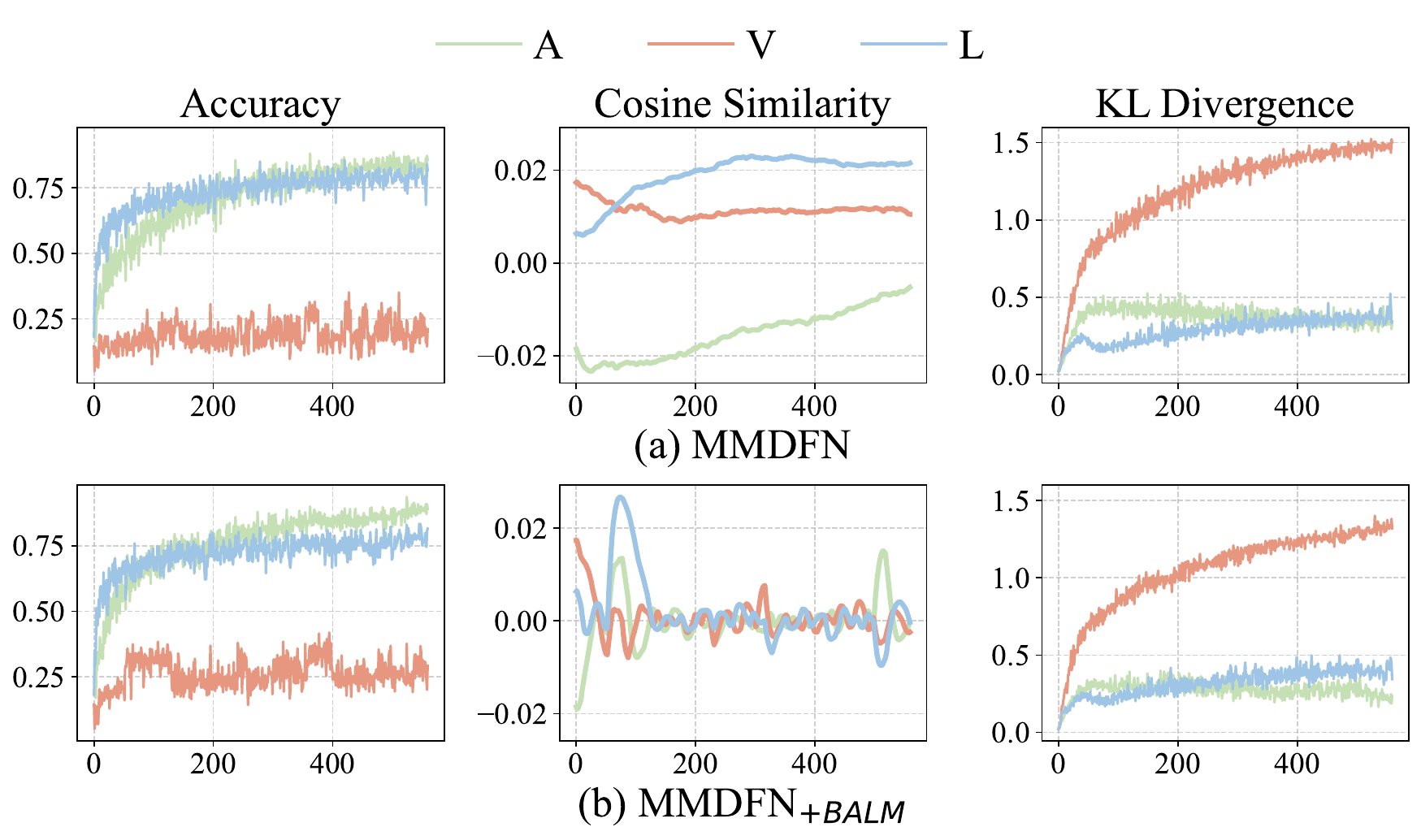}\vspace{-3pt}
    \caption{$(r_A,r_L,r_V)=(0.5, 0.3, 0.7)$}
\end{figure}
\begin{figure}[h!]
    \centering
    \includegraphics[width=\linewidth]{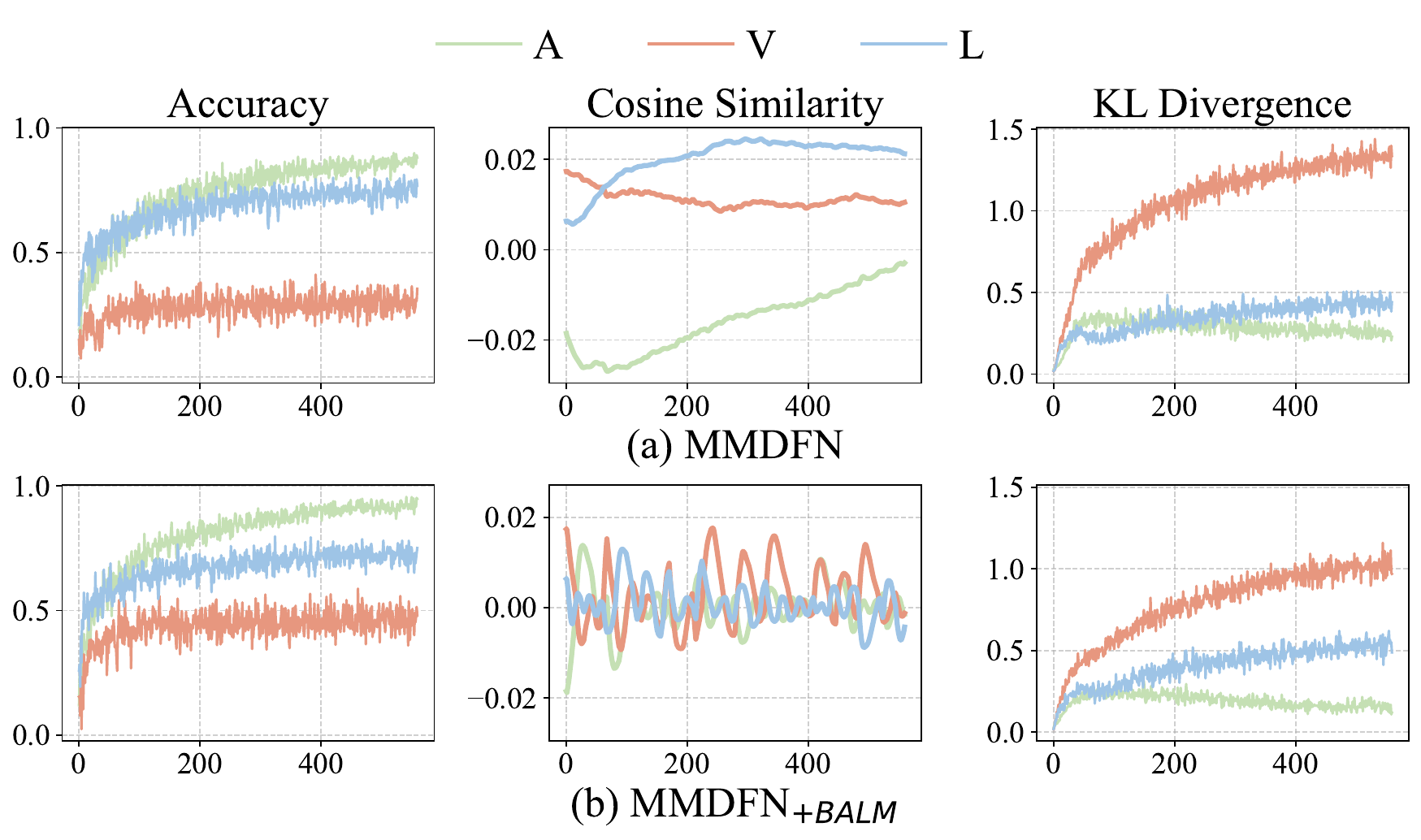}\vspace{-3pt}
    \caption{$(r_A,r_L,r_V)=(0.5, 0.7, 0.3)$}
\end{figure}
\begin{figure}[h!]
    \centering
    \includegraphics[width=\linewidth]{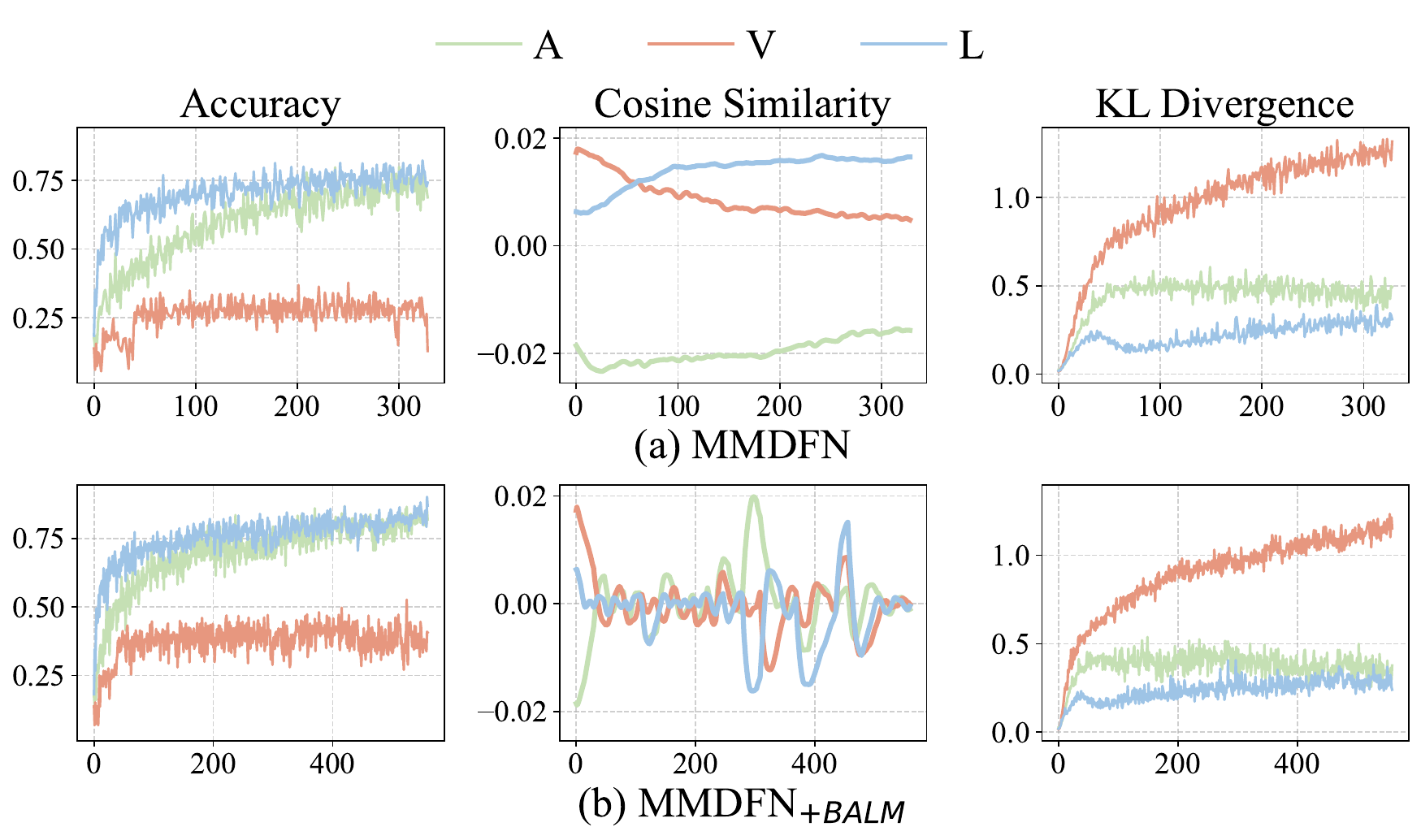}\vspace{-3pt}
    \caption{$(r_A,r_L,r_V)=(0.7, 0.3, 0.5)$}
\end{figure}



%% file: material/dataset_statistics.tex
\resizebox{\linewidth}{!}{%
\begin{tabular}{ccccccc}
\toprule
\multirow{2}{*}{\textbf{Dataset}} & \multicolumn{3}{c}{\textbf{Dialogues}} & \multicolumn{3}{c}{\textbf{Utterances}} \\ \cmidrule(lr){2-4} \cmidrule(lr){5-7}
 & \multicolumn{1}{c}{train} & \multicolumn{1}{c}{valid} & test & \multicolumn{1}{c}{train} & \multicolumn{1}{c}{valid} & test \\ \midrule
\textbf{IEMOCAP} & \multicolumn{2}{c}{120} & 31 & \multicolumn{2}{c}{5810} & 1623 \\ 
\textbf{CMU-MOSEI} & \multicolumn{1}{c}{2249} & \multicolumn{1}{c}{300} & 676 & \multicolumn{1}{c}{16326} & \multicolumn{1}{c}{1871} & 4659 \\ 
\bottomrule
\end{tabular}%
}

%% file: material/train-config.tex
\resizebox{0.7\linewidth}{!}{%
\begin{tabular}{c|cc}
\toprule
   \textbf{Parameters}        & \textbf{IEMOCAP} & \textbf{CMU-MOSEI} \\ \midrule
Audio dim $d_a$         & 1582             & 512                \\
Lexical dim $d_l$         & 1024             & 1024               \\
Visual dim $d_v$         & 342              & 1024               \\ 
\midrule
$d_\text{global}$    & 737              & 640                \\
$\rho$        & \multicolumn{2}{c}{(1.1,1.6]}       \\
$\tau$        & \multicolumn{2}{c}{(0.2,0.8]}       \\ \midrule
batch size & 16               & 32                 \\
epoch      & 80               & 25                 \\ \bottomrule
\end{tabular}%
}

%% file: material/spl.tune-configAB.tex
\resizebox{0.85\linewidth}{!}{%
\begin{tabular}{cc|ccccc|cc|ccccc}
\hline
\multicolumn{2}{c|}{\multirow{2}{*}{\begin{tabular}[c]{@{}c@{}}\textbf{Config A}\\ $(0.5,0.3,0.7)$\end{tabular}}} &
  \multicolumn{5}{c|}{$\rho$} &
  \multicolumn{2}{c|}{\multirow{2}{*}{\begin{tabular}[c]{@{}c@{}}\textbf{Config  B}\\ $(0.5, 0.7,0.3)$\end{tabular}}} &
  \multicolumn{5}{c}{$\rho$} \\ \cline{3-7} \cline{10-14} 
\multicolumn{2}{c|}{}       & 1.0   & 1.2   & 1.4   & 1.6   & 1.8   & \multicolumn{2}{c|}{}       & 1.0   & 1.2   & 1.4   & 1.6   & 1.8   \\ \hline
\multicolumn{1}{c|}{\multirow{5}{*}{$\tau$}} &
  0.1 &
  64.38 &
  64.92 &
  63.71 &
  63.90 &
  65.93 &
  \multicolumn{1}{c|}{\multirow{5}{*}{$\tau$}} &
  0.1 &
  59.49 &
  61.37 &
  \textbf{62.41} &
  59.39 &
  61.26 \\
\multicolumn{1}{c|}{} & 0.4 & 65.50 & 64.71 & 65.53 & 63.80 & 64.31 & \multicolumn{1}{c|}{} & 0.4 & 59.98 & 60.62 & 61.80 & 60.22 & 60.45 \\
\multicolumn{1}{c|}{} & 0.7 & 64.88 & 64.14 & 63.59 & 64.14 & 63.57 & \multicolumn{1}{c|}{} & 0.7 & 60.21 & 59.05 & 60.86 & 61.34 & 59.11 \\
\multicolumn{1}{c|}{} & 1.0 & 65.10 & 63.71 & 64.26 & \textbf{66.30} & 65.67 & \multicolumn{1}{c|}{} & 1.0 & 60.35 & 60.58 & 60.73 & 59.48 & 59.87 \\
\multicolumn{1}{c|}{} & 1.3 & 63.64 & 64.50 & 64.03 & 64.16 & 64.37 & \multicolumn{1}{c|}{} & 1.3 & 61.34 & 60.21 & 59.02 & 58.39 & 59.09 \\ \hline
\end{tabular}%
}

%% file: material/abl-modal-acc.tex
\resizebox{0.55\linewidth}{!}{%
\begin{tabular}{clccccccc}
\toprule
\textbf{MR Setting} & \textbf{Model} & \textbf{A} & \textbf{V} & \textbf{L} & \textbf{AV} & \textbf{AL} & \textbf{LV} & \textbf{ALV} \\ 
\midrule
\multirow{3}{*}{(0.5, 0.5, 0.7)} 
& GCNet& 46.33 & 37.28 & 57.42 & 48.18 & 59.09 & 54.16 & 58.47 \\
& GCNet\textsubscript{\textbf{+\myName}} & 46.95 & 31.44 & 59.09 & 50.89 & 59.64 & 54.41 & 59.33 \\
& MMGCN\textsubscript{\textbf{+\myName}} & 51.45 & 31.98 & 60.63 & 46.77 & 61.18 & 62.60 & 60.44\\
\midrule
\multirow{3}{*}{(0.5, 0.7, 0.5)} 
& GCNet& 49.04 & 31.85 & 56.99 & 48.49 & 56.69 & 53.73 & 57.86 \\
& GCNet\textsubscript{\textbf{+\myName}} & 47.13 & 27.60 & 54.59 & 46.21 & 58.10 & 55.76 & 55.82 \\
& MMGCN\textsubscript{\textbf{+\myName}} & 51.02 & 29.94 & 59.89 & 51.26 & 62.23 & 52.50 & 61.49 \\
\midrule
\multirow{3}{*}{(0.7, 0.5, 0.5)} 
& GCNet& 44.18 & 30.19 & 56.44 & 38.32 & 56.93 & 56.81 & 58.23 \\
& GCNet\textsubscript{\textbf{+\myName}} & 48.37 & 30.75 & 56.90 & 47.57 & 60.01 & 58.60 & 57.86 \\
& MMGCN\textsubscript{\textbf{+\myName}} & 48.92 & 29.94 & 62.97 & 46.03 & 62.54 & 60.07 & 62.85 \\
\midrule
\multirow{3}{*}{(0.3, 0.5, 0.7)} 
& GCNet& 47.87 & 32.72 & 53.54 & 48.55 & 60.20 & 55.51 & 59.46 \\
& GCNet\textsubscript{\textbf{+\myName}} & 51.57 & 32.10 & 54.53 & 53.05 & 60.38 & 54.53 & 61.12 \\
& MMGCN\textsubscript{\textbf{+\myName}} & 52.93 & 28.34 & 60.57 & 52.43 & 62.78 & 60.38 & 62.78 \\
\midrule
\multirow{3}{*}{(0.5, 0.3, 0.7)} 
& GCNet& 45.41 & 32.35 & 59.64 & 45.96 & 59.64 & 57.92 & 58.60 \\
& GCNet\textsubscript{\textbf{+\myName}} & 51.26 & 35.00 & 59.64 & 51.76 & 61.80 & 60.20 & 60.69 \\
& MMGCN\textsubscript{\textbf{+\myName}} & 50.89 & 28.34 & 63.71 & 50.40 & 64.08 & 61.31 & 64.39 \\
\midrule
\multirow{3}{*}{(0.7, 0.5, 0.3)} 
& GCNet& 46.89 & 35.49 & 54.47 & 48.86 & 56.01 & 54.78 & 58.41 \\
& GCNet\textsubscript{\textbf{+\myName}} & 44.55 & 35.43 & 55.51 & 52.00 & 54.10 & 55.88 & 59.33 \\
& MMGCN\textsubscript{\textbf{+\myName}} & 45.96 & 28.90 & 61.61 & 46.95 & 61.86 & 59.09 & 61.74 \\
\midrule
\textbf{Average} 
& GCNet & 46.62 & \textbf{33.31} & 56.42 & 46.39 & 58.09 & 55.49 & 58.51 \\
& GCNet\textsubscript{\textbf{+\myName}} &  48.31 & 32.05 & 56.71 & \textbf{50.25} & 59.01 & 56.56 & 59.02 \\
& MMGCN\textsubscript{\textbf{+\myName}} &  \textbf{50.20} & 29.57 & \textbf{61.56} & 48.97 & \textbf{62.45} & \textbf{59.33} & \textbf{62.28} \\
\bottomrule
\end{tabular}%
}

%% file: material/abl-modal-f1.tex
\resizebox{0.55\linewidth}{!}{%
\begin{tabular}{clccccccc}
\toprule
\textbf{MR Setting} & \textbf{Model} & \textbf{A} & \textbf{V} & \textbf{L} & \textbf{AV} & \textbf{AL} & \textbf{LV} & \textbf{ALV} \\ 
\midrule
\multirow{3}{*}{(0.5, 0.5, 0.7)} 
& GCNet& 45.91 & 34.50 & 57.76 & 48.44 & 59.06 & 54.02 & 58.52 \\
& GCNet\textsubscript{\textbf{+\myName}} & 46.56 & 27.57 & 59.47 & 50.56 & 59.77 & 53.93 & 59.51 \\
& MMGCN\textsubscript{\textbf{+\myName}} & 50.66 & 25.81 & 61.05 & 48.66 & 60.44 & 62.12 & 60.39\\
\midrule
\multirow{3}{*}{(0.5, 0.7, 0.5)} 
& GCNet& 44.14 & 29.65 & 56.58 & 49.40 & 56.71 & 53.81 & 58.15 \\
& GCNet\textsubscript{\textbf{+\myName}} & 48.36 & 26.36 & 54.82 & 46.89 & 58.36 & 56.05 & 58.75 \\
& MMGCN\textsubscript{\textbf{+\myName}} & 50.07 & 28.88 & 59.50 & 51.01 & 61.85 & 51.77 & 61.37 \\
\midrule
\multirow{3}{*}{(0.7, 0.5, 0.5)} 
& GCNet& 42.68 & 25.13 & 56.54 & 35.41 & 57.30 & 57.26 & 58.35 \\
& GCNet\textsubscript{\textbf{+\myName}} & 45.31 & 26.62 & 54.77 & 46.54 & 59.99 & 58.63 & 58.01 \\
& MMGCN\textsubscript{\textbf{+\myName}} & 48.85 & 28.88 & 62.18 & 46.00 & 62.08 & 60.29 & 62.78 \\
\midrule
\multirow{3}{*}{(0.3, 0.5, 0.7)} 
& GCNet& 43.77 & 32.29 & 52.87 & 48.67 & 60.68 & 54.84 & 59.40 \\
& GCNet\textsubscript{\textbf{+\myName}} & 50.54 & 28.40 & 54.75 & 51.26 & 60.57 & 54.52 & 61.40 \\
& MMGCN\textsubscript{\textbf{+\myName}} & 52.51 & 26.86 & 60.95 & 51.91 & 62.73 & 60.61 & 62.17 \\
\midrule
\multirow{3}{*}{(0.5, 0.3, 0.7)} 
& GCNet& 43.47 & 32.25 & 59.85 & 44.77 & 60.09 & 57.73 & 58.88 \\
& GCNet\textsubscript{\textbf{+\myName}} & 49.61 & 34.11 & 59.75 & 50.92 & 62.21 & 60.43 & 60.78 \\
& MMGCN\textsubscript{\textbf{+\myName}} & 50.87 & 26.86 & 63.64 & 50.33 & 63.28 & 61.33 & 63.73 \\
\midrule
\multirow{3}{*}{(0.7, 0.5, 0.3)} 
& GCNet& 46.08 & 33.43 & 54.02 & 49.48 & 56.14 & 54.54 & 58.61 \\
& GCNet\textsubscript{\textbf{+\myName}} & 42.91 & 32.98 & 55.51 & 51.46 & 54.09 & 56.45 & 58.98 \\
& MMGCN\textsubscript{\textbf{+\myName}} & 45.19 & 25.48 & 61.54 & 50.63 & 61.15 & 59.21 & 61.92 \\
\midrule
\textbf{Average} 
& GCNet &  44.34 &  \textbf{31.21} &  56.27 &  46.03 &  58.33 &  55.37 &  58.65 \\
& GCNet\textsubscript{\textbf{+\myName}} &  47.22 &  29.34 &  56.51 &  49.61 &  59.17 &  56.67 &  59.57 \\
& MMGCN\textsubscript{\textbf{+\myName}} &  \textbf{49.69} &  27.13 &  \textbf{61.48} &  \textbf{49.76} &  \textbf{61.92} &  \textbf{59.22} &  \textbf{62.06} \\
\bottomrule
\end{tabular}%
}

%% file: material/spl.abl-FCM-both.tex
\resizebox{\linewidth}{!}{%
\begin{tabular}{lcccccccc}
\toprule
\textbf{Modality} & \multicolumn{4}{c}{\textbf{IEMOCAP}} & \multicolumn{4}{c}{\textbf{CMU-MOSEI}} \\
\cmidrule(lr){2-5} \cmidrule(lr){6-9}
 & \multicolumn{2}{c}{\textbf{MMGCN}} & \multicolumn{2}{c}{\textbf{+FCM}} 
 & \multicolumn{2}{c}{\textbf{MMGCN}} & \multicolumn{2}{c}{\textbf{+FCM}} \\
\cmidrule(lr){2-3}\cmidrule(lr){4-5}\cmidrule(lr){6-7}\cmidrule(lr){8-9}
 & Acc & w-F1 & Acc & w-F1 & Acc & w-F1 & Acc & w-F1 \\
\midrule
A   & 31.05 & 25.97 & 39.99 & 34.75 & 62.85 & 48.51 & 62.85 & 48.51 \\
L   & 61.92 & 61.01 & 62.48 & 61.56 & 83.46 & 83.68 & 85.75 & 85.76 \\
V   & 19.78 & 11.67 & 20.70 & 12.07 & 62.91 & 48.66 & 62.85 & 50.06 \\
AL  & 64.45 & 63.11 & 63.77 & 62.15 & 83.57 & 83.78 & 85.72 & 85.71 \\
AV  & 32.29 & 26.92 & 42.88 & 38.07 & 62.91 & 48.64 & 62.85 & 48.86 \\
LV  & 61.98 & 61.03 & 62.91 & 61.92 & 85.39 & 85.40 & 85.91 & 85.87 \\
ALV & 65.19 & 63.94 & 64.20 & 62.80 & 85.55 & 85.55 & 85.88 & 85.83 \\
\midrule
\textbf{Avg.} & 48.09 & 44.81 & \textbf{50.99} & \textbf{47.62} 
              & 75.23 & 69.17 & \textbf{75.97} & \textbf{70.09} \\
\bottomrule
\end{tabular}%
}

%% file: material/smr-res-iemocap.tex

%
\resizebox{0.9\linewidth}{!}{%
\begin{tabular}{ccccccccccc}
\toprule
 \textbf{SMR}& \multicolumn{2}{c}{\textbf{Mi-CGA}} & \multicolumn{2}{c}{\textbf{SDR-GNN}} & \multicolumn{2}{c}{\textbf{GCNet}} & \multicolumn{2}{c}{\textbf{MMGCN}\textsubscript{+\myName}} & \multicolumn{2}{c}{\textbf{MMDFN}\textsubscript{+\myName}}\\ 
 \cmidrule(lr){2-3}\cmidrule(lr){4-5}\cmidrule(lr){6-7}\cmidrule(lr){8-9}\cmidrule(lr){10-11}
 & Acc & w-F1 & Acc & w-F1 & Acc & w-F1 & Acc & w-F1 & Acc & w-F1\\
\midrule
\multicolumn{1}{c|}{0.0} & 64.45\textsubscript{$\pm0.76$} & 64.49\textsubscript{$\pm0.94$} & 64.86\textsubscript{$\pm0.78$} & 64.65\textsubscript{$\pm0.83$} & 64.90\textsubscript{$\pm0.59$} & 65.01\textsubscript{$\pm0.68$} & 65.63\textsubscript{$\pm1.92$} & 65.49\textsubscript{$\pm1.90$} & 70.02\textsubscript{$\pm0.95$} & 69.89\textsubscript{$\pm0.81$}\\
\multicolumn{1}{c|}{0.1} & 63.43\textsubscript{$\pm1.27$} & 63.42\textsubscript{$\pm1.24$} & 63.43\textsubscript{$\pm1.07$} & 63.35\textsubscript{$\pm0.99$} & 64.10\textsubscript{$\pm0.73$} & 63.99\textsubscript{$\pm0.63$} & 64.28\textsubscript{$\pm0.71$} & 64.05\textsubscript{$\pm0.74$} & 68.90\textsubscript{$\pm0.99$} & 68.78\textsubscript{$\pm1.06$}\\
\multicolumn{1}{c|}{0.2} & 63.71\textsubscript{$\pm1.68$} & 63.76\textsubscript{$\pm1.68$} & 62.28\textsubscript{$\pm1.48$} & 62.00\textsubscript{$\pm1.41$} & 63.18\textsubscript{$\pm1.57$} & 62.83\textsubscript{$\pm1.67$} & 63.73\textsubscript{$\pm1.84$} & 63.54\textsubscript{$\pm1.89$} & 68.56\textsubscript{$\pm0.73$} & 68.45\textsubscript{$\pm0.72$}\\
\multicolumn{1}{c|}{0.3} & 60.46\textsubscript{$\pm1.26$} & 59.93\textsubscript{$\pm1.46$} & 60.54\textsubscript{$\pm3.23$} & 59.92\textsubscript{$\pm2.92$} & 62.50\textsubscript{$\pm0.61$} & 62.14\textsubscript{$\pm0.80$} & 62.76\textsubscript{$\pm1.16$} & 62.67\textsubscript{$\pm1.21$} & 67.43\textsubscript{$\pm1.38$} & 67.44\textsubscript{$\pm1.37$}\\
\multicolumn{1}{c|}{0.4} & 59.99\textsubscript{$\pm1.96$} & 60.18\textsubscript{$\pm2.02$} & 59.16\textsubscript{$\pm1.25$} & 58.85\textsubscript{$\pm1.16$} & 60.31\textsubscript{$\pm1.01$} & 60.01\textsubscript{$\pm1.38$} & 62.12\textsubscript{$\pm0.97$} & 61.90\textsubscript{$\pm0.94$} & 66.21\textsubscript{$\pm0.98$} & 66.15\textsubscript{$\pm0.95$}\\
\multicolumn{1}{c|}{0.5} & 58.07\textsubscript{$\pm1.49$} & 58.09\textsubscript{$\pm1.64$} & 58.79\textsubscript{$\pm2.18$} & 58.20\textsubscript{$\pm1.71$} & 58.68\textsubscript{$\pm0.85$} & 58.54\textsubscript{$\pm0.80$} & 61.77\textsubscript{$\pm1.53$} & 61.82\textsubscript{$\pm1.72$} & 64.93\textsubscript{$\pm1.47$} & 64.78\textsubscript{$\pm1.48$}\\
\multicolumn{1}{c|}{0.6} & 58.43\textsubscript{$\pm1.09$} & 58.42\textsubscript{$\pm1.26$} & 55.71\textsubscript{$\pm2.32$} & 55.27\textsubscript{$\pm2.15$} & 55.00\textsubscript{$\pm2.37$} & 54.83\textsubscript{$\pm2.47$} & 60.75\textsubscript{$\pm1.93$} & 60.71\textsubscript{$\pm1.94$} & 63.64\textsubscript{$\pm2.33$} & 63.27\textsubscript{$\pm2.29$}\\
\multicolumn{1}{c|}{0.7} & 55.79\textsubscript{$\pm0.85$} & 55.40\textsubscript{$\pm1.67$} & 54.41\textsubscript{$\pm2.47$} & 53.65\textsubscript{$\pm2.46$} & 53.91\textsubscript{$\pm0.83$} & 53.52\textsubscript{$\pm1.21$} & 59.68\textsubscript{$\pm2.09$} & 59.59\textsubscript{$\pm2.20$} & 61.63\textsubscript{$\pm2.36$} & 61.80\textsubscript{$\pm2.15$}\\
\bottomrule
\end{tabular}
}

%% file: material/smr-res-mosei.tex

\resizebox{0.9\linewidth}{!}{%
\begin{tabular}{ccccccccccc}
\toprule
 \textbf{SMR}& \multicolumn{2}{c}{\textbf{Mi-CGA}} & \multicolumn{2}{c}{\textbf{SDR-GNN}} & \multicolumn{2}{c}{\textbf{GCNet}} & \multicolumn{2}{c}{\textbf{MMGCN}\textsubscript{+\myName}} & \multicolumn{2}{c}{\textbf{MMDFN}\textsubscript{+\myName}}\\ 
 \cmidrule(lr){2-3}\cmidrule(lr){4-5}\cmidrule(lr){6-7}\cmidrule(lr){8-9}\cmidrule(lr){10-11}
 & Acc & w-F1 & Acc & w-F1 & Acc & w-F1 & Acc & w-F1 & Acc & w-F1 \\
\midrule
\multicolumn{1}{c|}{0.0} & 86.23\textsubscript{$\pm0.44$} & 86.21\textsubscript{$\pm0.35$} & 86.60\textsubscript{$\pm0.56$} & 86.58\textsubscript{$\pm0.51$} & 87.00\textsubscript{$\pm0.32$} & 86.94\textsubscript{$\pm0.39$} & 87.33\textsubscript{$\pm0.16$} & 87.29\textsubscript{$\pm0.19$} & 86.98\textsubscript{$\pm0.38$} & 86.96\textsubscript{$\pm0.36$}\\
\multicolumn{1}{c|}{0.1} & 85.06\textsubscript{$\pm0.43$} & 85.04\textsubscript{$\pm0.50$} & 85.82\textsubscript{$\pm0.30$} & 85.80\textsubscript{$\pm0.31$} & 85.92\textsubscript{$\pm0.29$} & 85.87\textsubscript{$\pm0.23$} & 85.81\textsubscript{$\pm0.22$} & 85.73\textsubscript{$\pm0.16$} & 86.13\textsubscript{$\pm0.53$} & 86.06\textsubscript{$\pm0.50$}\\
\multicolumn{1}{c|}{0.2} & 83.60\textsubscript{$\pm0.68$} & 83.47\textsubscript{$\pm0.56$} & 84.85\textsubscript{$\pm0.36$} & 84.77\textsubscript{$\pm0.34$} & 84.57\textsubscript{$\pm0.83$} & 84.56\textsubscript{$\pm0.71$} & 85.55\textsubscript{$\pm0.19$} & 85.45\textsubscript{$\pm0.18$} & 84.78\textsubscript{$\pm0.36$} & 84.65\textsubscript{$\pm0.34$}\\
\multicolumn{1}{c|}{0.3} & 82.26\textsubscript{$\pm0.50$} & 82.23\textsubscript{$\pm0.47$} & 83.34\textsubscript{$\pm0.39$} & 83.18\textsubscript{$\pm0.26$} & 83.83\textsubscript{$\pm0.30$} & 83.73\textsubscript{$\pm0.29$} & 83.90\textsubscript{$\pm0.22$} & 83.67\textsubscript{$\pm0.34$} & 83.87\textsubscript{$\pm0.13$} & 83.80\textsubscript{$\pm0.18$}\\
\multicolumn{1}{c|}{0.4} & 81.33\textsubscript{$\pm0.38$} & 81.19\textsubscript{$\pm0.42$} & 82.35\textsubscript{$\pm0.56$} & 81.99\textsubscript{$\pm0.72$} & 82.58\textsubscript{$\pm0.53$} & 82.36\textsubscript{$\pm0.50$} & 82.71\textsubscript{$\pm0.50$} & 82.44\textsubscript{$\pm0.48$} & 82.86\textsubscript{$\pm0.55$} & 82.65\textsubscript{$\pm0.50$}\\
\multicolumn{1}{c|}{0.5} & 80.06\textsubscript{$\pm0.74$} & 80.05\textsubscript{$\pm0.71$} & 80.88\textsubscript{$\pm0.72$} & 80.59\textsubscript{$\pm0.83$} & 81.44\textsubscript{$\pm0.64$} & 81.26\textsubscript{$\pm0.63$} & 81.92\textsubscript{$\pm0.22$} & 81.63\textsubscript{$\pm0.21$} & 81.83\textsubscript{$\pm0.32$} & 81.59\textsubscript{$\pm0.41$}\\
\multicolumn{1}{c|}{0.6} & 79.01\textsubscript{$\pm0.61$} & 78.86\textsubscript{$\pm0.53$} & 79.75\textsubscript{$\pm0.60$} & 79.49\textsubscript{$\pm0.56$} & 79.92\textsubscript{$\pm0.79$} & 79.86\textsubscript{$\pm0.68$} & 80.28\textsubscript{$\pm0.19$} & 80.07\textsubscript{$\pm0.21$} & 80.46\textsubscript{$\pm0.28$} & 80.16\textsubscript{$\pm0.18$}\\
\multicolumn{1}{c|}{0.7} & 78.44\textsubscript{$\pm0.31$} & 78.16\textsubscript{$\pm0.35$} & 78.64\textsubscript{$\pm0.75$} & 78.46\textsubscript{$\pm0.64$} & 78.55\textsubscript{$\pm1.16$} & 78.45\textsubscript{$\pm1.01$} & 79.05\textsubscript{$\pm0.23$} & 78.71\textsubscript{$\pm0.23$} & 79.31\textsubscript{$\pm0.35$} & 78.95\textsubscript{$\pm0.24$}\\
\bottomrule
\end{tabular}
}

%% file: material/tab-abl-subgrm.tex
\resizebox{\linewidth}{!}{%
\begin{tabular}{ccccccccc}
\toprule
\textbf{IMR Settings}                        & \multicolumn{4}{c}{\textbf{IEMOCAP}} & \multicolumn{4}{c}{\textbf{CMU-MOSEI}} \\ \cmidrule(lr){2-5} \cmidrule(lr){6-9} 
\textbf{} & \multicolumn{2}{c}{\textbf{+\myName-D}} & \multicolumn{2}{c}{\textbf{+\myName-S}} & \multicolumn{2}{c}{\textbf{+\myName-D}} & \multicolumn{2}{c}{\textbf{+\myName-S}} \\ \cmidrule(lr){2-3} \cmidrule(lr){4-5} \cmidrule(lr){6-7} \cmidrule(lr){8-9} 
                                   & Acc     & w-F1    & Acc     & w-F1   & Acc      & w-F1    & Acc     & w-F1    \\ 
                                \midrule
\multicolumn{1}{c|}{(0.3, 0.5, 0.7)} & 64.51   & 64.02   & 62.29   & 62.47  & 80.79    & 80.69   & 81.04   & 80.39   \\
\multicolumn{1}{c|}{(0.3, 0.7, 0.5)} & 61.49   & 60.80    & 59.15   & 59.51  & 78.01    & 77.82   & 78.32   & 77.77   \\
\multicolumn{1}{c|}{(0.5, 0.3, 0.7)} & 63.83   & 63.50    & 65.13   & 65.18  & 84.04    & 83.76   & 84.07   & 83.86   \\
\multicolumn{1}{c|}{(0.5, 0.7, 0.3)} & 60.63   & 60.53   & 60.87   & 60.90   & 77.99    & 77.66   & 79.25   & 78.94   \\
\multicolumn{1}{c|}{(0.7, 0.3, 0.5)} & 64.76   & 64.65   & 64.26   & 64.49  & 84.56    & 84.31   & 84.31   & 84.22   \\
\multicolumn{1}{c|}{(0.7, 0.5, 0.3)} & 60.57   & 60.60    & 60.44   & 60.63  & 82.55    & 82.43   & 81.65   & 81.53   \\ 
\bottomrule
\end{tabular}%
}

%% file: material/abl-mse-iemocap.tex
\resizebox{\linewidth}{!}{%
\begin{tabular}{ccccccc}
\toprule
\textbf{MR Setting} & \multicolumn{2}{c}{\textbf{GCNet\textsubscript{+\myName-M}}} & \multicolumn{2}{c}{\textbf{MMGCN\textsubscript{+\myName-M}}} & \multicolumn{2}{c}{\textbf{MMDFN\textsubscript{+\myName-M}}} \\ \cmidrule(lr){2-3}\cmidrule(lr){4-5}\cmidrule(lr){6-7} 
                                     & Acc   & w-F1  & Acc   & w-F1  & Acc   & w-F1  \\ \midrule
\multicolumn{1}{c|}{$(0.3, 0.5, 0.7)$} & 59.46\textsubscript{$\downarrow1.85$} & 59.93 & 65.50\textsubscript{$\uparrow1.11$} & 65.26 & 69.01\textsubscript{$\uparrow1.67$} & 68.79 \\
\multicolumn{1}{c|}{$(0.3, 0.7, 0.5)$} & 58.23\textsubscript{$\uparrow0.87$} & 58.47 & 60.57\textsubscript{$\uparrow0.50$} & 60.69 & 64.08\textsubscript{$\uparrow0.68$} & 63.54 \\
\multicolumn{1}{c|}{$(0.5, 0.3, 0.7)$} & 61.43\textsubscript{$\uparrow0.12$} & 61.57 & 65.13\textsubscript{$\downarrow0.24$} & 65.19 & 69.32\textsubscript{$\uparrow0.56$} & 69.06 \\
\multicolumn{1}{c|}{$(0.5, 0.7, 0.3)$} & 56.87\textsubscript{$\downarrow0.77$} & 57.12 & 60.63\textsubscript{$\downarrow1.66$} & 60.68 & 64.39\textsubscript{$\uparrow0.37$} & 64.10 \\
\multicolumn{1}{c|}{$(0.7, 0.3, 0.5$)} & 58.16\textsubscript{$\downarrow1.17$} & 58.31 & 63.89\textsubscript{$\downarrow1.11$} & 63.90 & 68.45\textsubscript{$\uparrow1.23$} & 68.29 \\
\multicolumn{1}{c|}{$(0.7, 0.5, 0.3)$} & 54.71\textsubscript{$\downarrow5.55$} & 54.97 & 61.18\textsubscript{$\downarrow0.56$} & 61.25 & 65.13\textsubscript{$\downarrow1.37$} & 64.70 \\ \bottomrule
\end{tabular}%
}

%% file: material/abl-mse-mosei.tex
\resizebox{\linewidth}{!}{%
\begin{tabular}{ccccccc}
\toprule
\textbf{MR Setting} & \multicolumn{2}{c}{\textbf{GCNet\textsubscript{+\myName-M}}} & \multicolumn{2}{c}{\textbf{MMGCN\textsubscript{+\myName-M}}} & \multicolumn{2}{c}{\textbf{MMDFN\textsubscript{+\myName-M}}} \\ \cmidrule(lr){2-3}\cmidrule(lr){4-5}\cmidrule(lr){6-7} 
                                     & Acc   & w-F1  & Acc   & w-F1  & Acc   & w-F1  \\ \midrule
\multicolumn{1}{c|}{$(0.3, 0.5, 0.7)$} & 81.70\textsubscript{$\uparrow0.16$} & 81.44 & 81.29\textsubscript{$\downarrow0.74$} & 80.69 & 81.40\textsubscript{$\downarrow0.66$} & 80.96\\
\multicolumn{1}{c|}{$(0.3, 0.7, 0.5)$} & 78.59\textsubscript{$\uparrow0.08$} & 78.10 & 78.40\textsubscript{$\downarrow0.33$} & 78.10 & 77.44\textsubscript{$\downarrow1.37$} & 77.34\\
\multicolumn{1}{c|}{$(0.5, 0.3, 0.7)$} & 82.83\textsubscript{$\downarrow0.44$} & 82.66 & 82.97\textsubscript{$\downarrow1.57$} & 82.97 & 83.65\textsubscript{$\downarrow0.69$} & 83.55\\
\multicolumn{1}{c|}{$(0.5, 0.7, 0.3)$} & 77.55\textsubscript{$\downarrow1.95$} & 77.54 & 77.96\textsubscript{$\downarrow1.87$} & 78.00 & 78.70\textsubscript{$\downarrow0.66$} & 78.50\\
\multicolumn{1}{c|}{$(0.7, 0.3, 0.5)$} & 82.97\textsubscript{$\downarrow0.66$} & 82.96 & 84.26\textsubscript{$\downarrow0.58$} & 84.07 & 84.48\textsubscript{$\downarrow0.30$} & 84.36\\
\multicolumn{1}{c|}{$(0.7, 0.5, 0.3)$} & 81.15\textsubscript{$\downarrow0.94$} & 81.11 & 82.72\textsubscript{$\uparrow0.22$} & 82.29 & 82.31\textsubscript{$\downarrow0.22$} & 82.16\\ \bottomrule
\end{tabular}%
}